\documentclass[11pt]{article}
\usepackage[margin=1in]{geometry}

\usepackage{natbib}
\bibpunct[, ]{(}{)}{,}{a}{}{,}%
\def\newblock{\ }%

\usepackage[utf8]{inputenc}
\usepackage{amssymb}
\usepackage{enumerate}
\usepackage[OT1]{fontenc}
\usepackage{bm}
\usepackage{multirow}
\usepackage{diagbox}
\usepackage{amsmath,amssymb,mathrsfs}
\usepackage{natbib}
\usepackage{xcolor}
\usepackage{bbm, dsfont}
\usepackage{makecell}
\usepackage{upgreek}
\usepackage{mathtools}
\usepackage{algorithm}
\usepackage{algorithmic}
\usepackage{graphicx}

\usepackage{dsfont,makecell,chngcntr}
\usepackage{multirow,stfloats,booktabs}

\usepackage[english]{babel}
\usepackage{graphicx} 
\usepackage{subcaption}
\usepackage{makecell}
\usepackage{array}
\usepackage[hypertexnames=false,
colorlinks,
linkcolor=blue,
anchorcolor=blue,
citecolor=blue
]{hyperref}

\def\tr{\mathop{\text{tr}}\kern.2ex}

\def\E{{\mathbb E}}

\def\cI{\mathcal{I}}
\def\xb{\mathbf{x}}
\def \bSigma {{\mathbf\Sigma}}
\def \bx {\boldsymbol{x}}

\def\supp{\mathop{\text{supp}}}

\long\def\comment#1{}

\def\tr{\mathop{\text{Tr}}}

\def\cS{{\mathcal{S}}}

\def\cA{\mathcal{A}}
\def\cB{\mathcal{B}}
\def\cK{\mathcal{K}}

\newcommand{\bel}{\begin{eqnarray}\label}
	\newcommand{\eel}{\end{eqnarray}}
\newcommand{\bes}{\begin{eqnarray*}}
	\newcommand{\ees}{\end{eqnarray*}}

\def\Wb{\mathbf{W}}

\definecolor{brandeisblue}{rgb}{0.0, 0.44, 1.0}

\let\emptyset\varnothing
\let\hat\widehat
\let\tilde\widetilde
\def\mid{\,|\,}

\def\wb{\mathbf{w}}
\def\EE{{\mathbb E}}

\def\II{\mathbb{I}}

\def\supp{\mathop{\text{supp}\kern.2ex}}

\def\tr{{\rm{Tr}}}

\def\supp{\mathop{\text{supp}}}

\def\tr{\mathrm{Tr}}

\newcommand{\sharpbound}{0}
\newcommand{\mcmodel}{0}
\newcommand{\theorysection}{0}
\usepackage{mathrsfs}

\usepackage{enumitem}

\def \bSigma {\mbox{\boldmath $\Sigma$}}
\usepackage{tabularx,array}
\newcolumntype{M}{>{\centering\arraybackslash}m{0.17\textwidth}}
\newcommand{\btheta}{\bm{\theta}}
\def\bu{\boldsymbol{u}}
\usepackage{hyperref}

\def\Sb{{\mathbf{S}}}
\def\bcS{{\boldsymbol{\mathcal{S}}}}
\def\Ab{{\mathbf{A}}}
\def\bcA{{\boldsymbol{\mathcal{A}}}}
\def\gb{{\mathbf{g}}}
\def\Gb{\mathbf{G}}
\def\rb{{\mathbf{r}}}
\def\Pb{{\mathbf{P}}}

\def\ub{{\mathbf{u}}}
\def\cN{{\mathcal{N}}}
\def\thetab{\boldsymbol{\theta}}
\def\vb{\mathbf{v}}
\def\eb{\boldsymbol{e}}
\def\cL{\mathcal{L}}
\def\EE{\mathbb{E}}
\def\cE{\mathcal{E}}
\def\PP{\mathbb{P}}
\def\Ib{\mathbf{I}}
\def\Bb{\mathbf{B}}
\def\Cb{\mathbf{C}}
\def\lambdab{\boldsymbol{\lambda}}

\def\bTheta{\mathbf{\Theta}}
\def\RR{\mathbb{R}}
\def\bcE{\boldsymbol{\cE}}
\def\ponep{\Pb_{\mathbf{1}}^{\perp}}
\def\Mb{\mathbf{M}}
\def\Lb{\mathbf{L}}
\def\Db{\mathbf{D}}
\def\Rb{\mathbf{R}}
\newcommand{\bz}{\bm{z}}

\newcommand{\bv}{\bm{v}}

\def\diag{\operatorname{diag}}

\usepackage{amsthm}
\theoremstyle{plain}
\newtheorem{theorem}{Theorem}
\newtheorem{lemma}{Lemma}
\newtheorem{corollary}{Corollary}
\newtheorem{proposition}{Proposition}
\theoremstyle{definition}
\newtheorem{definition}{Definition}
\newtheorem{example}{Example}
\newtheorem{assumption}{Assumption}
\theoremstyle{remark}
\newtheorem{remark}{Remark}
\newcommand{\Halmos}{\ensuremath{\square}}
\renewcommand{\proof}[1]{\par\medskip\noindent{\itshape #1}\ \ignorespaces}
\renewcommand{\endproof}{\hfill\Halmos\par\medskip}
\newcommand{\ECswitch}{%
  \setcounter{section}{0}\renewcommand{\thesection}{EC.\arabic{section}}%
  \setcounter{equation}{0}\renewcommand{\theequation}{EC.\arabic{equation}}%
  \setcounter{figure}{0}\renewcommand{\thefigure}{EC.\arabic{figure}}%
  \setcounter{table}{0}\renewcommand{\thetable}{EC.\arabic{table}}%
  \setcounter{theorem}{0}\renewcommand{\thetheorem}{EC.\arabic{theorem}}%
  \setcounter{lemma}{0}\renewcommand{\thelemma}{EC.\arabic{lemma}}%
  \setcounter{corollary}{0}\renewcommand{\thecorollary}{EC.\arabic{corollary}}%
  \setcounter{proposition}{0}\renewcommand{\theproposition}{EC.\arabic{proposition}}%
  \setcounter{definition}{0}\renewcommand{\thedefinition}{EC.\arabic{definition}}%
  \setcounter{example}{0}\renewcommand{\theexample}{EC.\arabic{example}}%
  \setcounter{assumption}{0}\renewcommand{\theassumption}{EC.\arabic{assumption}}%
  \setcounter{remark}{0}\renewcommand{\theremark}{EC.\arabic{remark}}%
}

\begin{document}

\title{Combinatorial Inference on the Optimal Assortment in Multinomial Logit Models}
\author{Shuting Shen\thanks{Department of Statistics and Data Science, National University of Singapore, \textit{shuting\_shen@nus.edu.sg}}
  \and Xi Chen\thanks{Department of Technology, Operations, and Statistics, Leonard N. Stern School of Business, New York University, \textit{xc13@stern.nyu.edu}}
  \and Ethan X. Fang\thanks{Department of Biostatistics \& Bioinformatics and Fuqua School of Business, Duke University, \textit{ethan.fang@duke.edu}}
  \and Junwei Lu\thanks{Department of Biostatistics, Harvard T.H. Chan School of Public Health, \textit{junweilu@hsph.harvard.edu}}}
\date{\today}
\maketitle

\begin{abstract}
Assortment optimization has been extensively explored in recent decades due to its practical significance. Despite the substantial literature on optimization algorithms and latent score estimation, there remains a need to explore uncertainty quantification for the optimal assortment. Instead of estimating and recovering the complete optimal offer set, decision-makers may be interested in testing whether a given property holds  for the optimal assortment. For instance, they may wish to determine whether certain products of interest should be included in the optimal set or ascertain how many categories of products the optimal set should contain. This paper proposes a novel inferential framework for testing such properties within the context of the widely adopted multinomial logit (MNL) model. In this model, each customer is assumed to purchase an item within the offered products with a probability proportional to the underlying preference score associated with the product. We reduce inferring a general optimal assortment property to quantifying the uncertainty associated with the sign change point detection of the marginal revenue gaps. We establish the asymptotic normality of the marginal revenue gap estimator and construct a maximum statistic using the gap estimators to detect the sign change point. By approximating the distribution of the maximum statistic through multiplier bootstrap techniques, we propose a valid testing procedure. Numerical experiments are also conducted to assess the performance of our method.
\end{abstract}

\section{Introduction}\label{sec: intro}

Assortment optimization has generated extensive research interest due to its important implications in revenue management. Essentially, assortment optimization aims to study the balance between customer demand and product revenues, where the total expected revenues depend on the profits of individual products as well as customers' preferences over the available products. Researchers have proposed various merchandising strategies to align the offered assortment with the customer decision-making process, aiming to maximize expected revenues. A multitude of applications can be found in the fields of economics \citep{cachon2005retail, COMEZDOLGAN2022108550}, marketing \citep{mantrala2009assortment, kok2015retail}, and operations \citep{blanchet2016mcpaper, aouad2018approximability}.

To model customer choice behavior for assortment planning, many parametric models have been proposed. Among them, the multinomial logit (MNL) model \citep{McFa73} is one of the most widely used because of its tractability and its efficient algorithmic solutions to assortment-optimization problems. For a full assortment of $n$ products labeled as $\{1,\ldots,n\}$, the MNL model associates each product $i$ with a preference score $u_i^*>0$. A customer chooses from an offered assortment $\cS\subseteq\{1,\ldots,n\}$, with a nonzero probability of making no purchase. The probability that the customer chooses product $j$ from the offer set $\cS$ is proportional to the underlying preference score, i.e.,
\[
    \PP(j\mid \cS\cup\{0\})=
    \frac{u_j^*}{1+\sum_{i\in\cS}u_i^*},\qquad j\in\cS,
\]
where $0$ denotes the no-purchase alternative. The objective of assortment optimization is to determine an optimal assortment $\cS^*$ that maximizes the expected revenue,
\[
    \cS^*\in\arg\max_{\cS\subseteq[n]} r(\cS)
    =\arg\max_{\cS\subseteq[n]}
    \sum_{i\in\cS}\frac{u_i^*r_i}{1+\sum_{j\in\cS}u_j^*},
\]
where $r_i>0$ is the revenue associated with product $i$, and $r(\cS)=\sum_{i\in\cS}u_i^*r_i/(1+\sum_{j\in\cS}u_j^*)$ is the total expected revenue from assortment $\cS$. The primary goal of this paper is to provide uncertainty quantification for structural properties of the optimal assortment $\cS^*$.

Previous efforts on assortment optimization have predominantly focused on the algorithmic determination of the optimal assortment \citep{talluri2004revenue,gallego2004managing,li2018integrating} or its point estimate from observed customer choice data \citep{agrawal2019mnl, chen2020dynamicopt}. However, inferential analysis on the optimal assortment remains underexplored, despite the practical importance of accounting for statistical uncertainty in data-driven decision-making \citep{conejo2010decision, lam2013multiple, jaillet2016routing}. Customer choice data are random, and a point estimate may fail to recover the true optimal assortment with nonnegligible probability \citep{agrawal2019mnl}. Relying exclusively on point estimates can therefore create risks for downstream decisions. First, point estimates may not be reproducible: even under the same MNL model and the same underlying preference scores, different realizations of customer choices can lead to different estimated optimal assortments. Second, point estimates do not quantify confidence. In contrast, uncertainty quantification provides confidence sets or hypothesis tests that explicitly account for sampling variability and allow decision makers to assess which structural conclusions about the optimal assortment are statistically supported.

As mentioned previously, prior research has made substantial advancements in various aspects of choice models, including the MNL model, with particular emphasis on optimization procedures for obtaining the optimal assortment. The tractability of the MNL model has enabled efficient assortment-optimization algorithms \citep{talluri2004revenue,gallego2004managing}. \citet{li2018integrating} considered assortment optimization for the two-stage MNL model using an empirical estimation method based on aggregated data. Other works have generalized the MNL model to accommodate richer choice behavior, including the nested logit model \citep{WilliamsHCWL1977OtFo}, the generalized attraction model \citep{GallegoGuillermo2015AGAM}, the marginal distribution model \citep{Natarajan2009persistency}, and the group marginal distribution model \citep{ruan2022limit}. However, optimization under more general choice models often becomes computationally difficult. For instance, \citet{rusmevichientong2010assortment} and \citet{davis2014assortment} showed that assortment optimization is NP-hard under a mixture of MNL models and under the nested logit model, respectively, and \citet{aouad2018approximability} characterized hardness of approximation for general choice models by reducing the optimization problem to a computational problem of detecting large independent sets in graphs.

The preceding literature focuses primarily on solving the assortment-optimization problem when the choice model is known. In practice, however, the preference parameters governing customer choice behavior are typically unknown and must be learned from data. This has motivated a large stream of work on learning unknown choice parameters. In particular, online learning methods provide dynamic estimates to balance exploration and exploitation. \citet{agrawal2019mnl} proposed an efficient algorithm for the capacitated MNL model that achieved a near-optimal regret bound, and \citet{chen2020dynamicopt} studied a contextual MNL model with time-varying features, upon which they designed a dynamic policy to simultaneously learn the unknown features while making adaptive decisions for the offered assortment. These methods provide point estimates or policies with regret guarantees. However, uncertainty quantification for the resulting optimal assortment remains largely unexplored. 

Robust optimization provides another approach to uncertainty quantification in customer choice behavior. For example, \citet{RUSMEVICHIENTONGPaat2012RAOi} brought the robustness against perturbation into the scope of assortment optimization under the MNL model by considering the robust optimization approach \citep{BertsimasDimitris2004TPoR,ahipacsaouglu2019distributionally,chen2021robust,chen2022distributionally,perakis2022robust,zhu2022joint}. Robust optimization, however, is tailored to worst-case performance and can be conservative: it solves
\[
\cS^{*}_{\mathrm{rob}} \in \arg\max_{\cS \subseteq [n]}\; \min_{\ub \in \cB(\ub^*)}  \sum_{i \in \cS} \frac{u_i r_i}{1 + \sum_{j \in \cS} u_j},
\]
where $\cB(\ub^*)$ is an uncertainty set containing the true utility $\ub^* = (u^*_1, \ldots, u^*_n)$. When $\cB(\ub^*)$ is large, $\cS^{*}_\mathrm{rob}$ for the worst-case scenario could be significantly more conservative than the optimal assortment $\cS^{*}$. In contrast, our goal is not to optimize against a worst-case utility vector, but to infer structural properties of the true MNL optimum $\cS^*$ with probabilistic error control. \citet{blanchet2016mcpaper} proposed an alternative Markov-chain-based approximation to random-utility choice models and developed an efficient algorithm. However, this line of work focuses on approximation and optimization, and does not provide uncertainty quantification for the optimal assortment.

{
A concrete setting motivating the proposed inference problem is a digital retail or marketplace platform that selects products for a search-results module, recommendation carousel, or promotional collection from a prescreened set of candidate items. Such display and recommendation settings have been studied in recent assortment-optimization models under MNL-type choice behavior \citep{feldman2023display, ali2025clickbasedmnl}. In such settings, an MNL-type model provides a tractable approximation for substitution among displayed products, and the no-purchase alternative corresponds to a customer leaving the module without purchasing. After products that are unavailable, contractually infeasible, or operationally irrelevant have been removed from the candidate pool, the remaining assortment decision can be treated as unconstrained for the purpose of the MNL optimization problem. Nevertheless, the decision maker may still need to answer structural questions before committing to a final assortment: whether a strategic product should be included, whether a focal category constitutes a sufficiently large share of the revenue-maximizing assortment, or whether one supplier's products constitute the largest share of the assortment. Such category-, supplier-, and diversity-related considerations are common in assortment planning \citep{flamand2018integrated, elhousni2024assortment}. These are inferential questions about the structure of the unconstrained MNL optimum, not additional constraints imposed on the optimization problem.}

{
This setting also motivates the observational-data framework studied in this paper. Historical impressions from recommendation modules, category pages, or search-result displays generate choice observations from past assortments. Active experimentation is possible in principle, but it can be costly because displaying inferior assortments to live users may reduce revenue, degrade the customer experience, or affect supplier relationships. The weighted comparison-graph condition introduced in Section~\ref{sec: mtd} captures whether historical displays provide sufficiently rich comparisons across products to estimate relative preferences at the scale needed for inference. Thus, our procedure is aimed at regimes in which historical assortments are informative enough for estimation, but exact recovery of the optimal cutoff $K^*$ remains statistically uncertain.
}

{
The above setting clarifies the practical role of the hypothesis tests considered in this paper. Many assortment decisions are governed by structural assessments rather than by the immediate need to identify a single exact assortment. For example, a merchandiser or platform operator may need to determine whether certain products or product categories should be included in the revenue-maximizing assortment, whether a product category is sufficiently represented, or whether a supplier's products dominate the optimal collection. Such structural assessments arise in assortment and retail-space planning, where decisions are often coupled with shelf-space, inventory, supplier, or category-balance considerations \citep{flamand2018integrated, Hubner2024Decision, elhousni2024assortment}. 
}

We next present three illustrative inferential questions. They should be interpreted as tests of structural properties of the unconstrained MNL optimum. The decision on whether to offer a given product of interest is a popular topic in marketing and management \citep{varley2014retail}. Denoting by $[n] = \{1,\ldots,n\}$ the full assortment of products, and by $\mathcal{S}^*$ the optimal offered assortment, we summarize the single product inclusion problem in the following example. 

$\bullet$ \textbf{Example 1.} For a product $i$ of interest, we test whether $i$ belongs to the optimal assortment:
\begin{align*}
    &\mathrm{H}_0: \text{Product }i\text{ is not in the optimal assortment }\cS^*,\\
    &\mathrm{H}_1: \text{Product }i\text{ is in the optimal assortment }\cS^*.
\end{align*}

Apart from inferring the inclusion of a single product, decision makers may also be interested in the category composition of the optimal assortment. Such questions are related to category-level assortment and shelf-space considerations in retail planning \citep{flamand2023store}. {Note that this is not a constrained-assortment problem with a category-quota constraint. Rather, it asks whether the unconstrained revenue-maximizing assortment implied by the MNL model has a particular category-composition property.} We formulate this application as the following hypothesis testing problem.

$\bullet$ \textbf{Example 2.} For a category of products $A=\{i_1,\ldots,i_K\}$ and a given ratio $q\%$, we test whether the category constitutes more than $q\%$ of the optimal assortment:
\begin{align*}
    &\mathrm{H}_0: \text{More than } q\% \text{ of }\cS^*\text{ are from category }A,\\
    &\mathrm{H}_1: \text{No more than } q\% \text{ of }\cS^*\text{ are from category }A.
\end{align*}

Decision makers may also be interested in whether products from a focal supplier or brand dominate the optimal assortment. Such supplier-management considerations are important in marketing and retail operations \citep{eda2009supplier}. We summarize this question in the example below.

$\bullet$ \textbf{Example 3.} For a partition $\{A_1,\ldots,A_m\}$ of the products, we test whether products in $A_1$ constitute the largest share of the optimal assortment:
\begin{align*}
    &\mathrm{H}_0: \text{$A_1$ contains the most products in }\cS^*\text{ compared with any other }A_j,\\
    &\mathrm{H}_1: \text{At least one other }A_j\text{ contains more products in }\cS^*\text{ than }A_1.
\end{align*}

{
These property questions do not require the decision maker to know the entire optimal assortment. Answering them by first recovering the full assortment can be inefficient, because controlling uncertainty for the full combinatorial object is typically more demanding than controlling uncertainty for the lower-dimensional feature that defines the property of interest. Even under the unconstrained MNL model, where exact recovery of $\cS^*$ is equivalent to identifying the lowest-revenue product in the optimal assortment $K^*$ (see Theorem~\ref{thm: alg}), this equivalence holds only at the level of point estimation. The estimator $\hat K$ depends on estimation errors across all items, and when the signal strength is insufficient, $\hat K$ may differ from $K^*$ with nonnegligible probability.
}

{
For instance, if the decision maker only needs to determine whether a few top-revenue products should be included, then collecting enough information to identify the lowest-revenue product in the full optimal assortment may be unnecessary. At the same time, relying solely on a point estimate can be misleading: if $\hat K$ excludes some of these products due to random estimation error, the resulting decision may prematurely discard high-revenue candidates. This creates a signal-strength regime in which exact recovery of $K^*$ is infeasible, yet valid inference on a partial property of $\cS^*$ remains possible. Intuitively, exact recovery requires uniform control over all items up to the threshold $K^*$, whereas a property test only needs to control the items or categories involved in the null hypothesis. The proposed confidence interval for $K^*$ is designed for this regime: it quantifies uncertainty around the revenue threshold and provides a statistically plausible set of assortment sizes that can be carried forward to downstream operational evaluation.
}

{
In Corollary~\ref{col: compare signal} (Section~\ref{sec: test proc}), we formalize this gap theoretically. We show that when $K^*$ is much larger than the cardinalities of the null sets under comparison, exact identification of the lowest-revenue product in the optimal assortment is substantially more demanding than inferring null properties of $\cS^*$ that involve far fewer items. Under the signal--noise regimes we analyze, exact identification of $K^*$ can be unattainable with high confidence, and point estimation of $K^*$ may err with nonnegligible probability. By contrast, valid inference on low-dimensional properties remains feasible under the same regimes by explicitly quantifying the uncertainty in $\hat K$, without requiring exact recovery of $K^*$.
}

{This perspective connects the present problem to broader questions in combinatorial inference, where the target is a discrete structural quantity and property testing can be feasible under weaker conditions than exact recovery. Inference on an integer-valued target, such as $K^*$ in our setting, is challenging because the discreteness of the parameter can make traditional smooth inferential techniques inapplicable. Related examples include testing the rank of a matrix \citep{chen2019improved} and testing the number of communities in stochastic block models \citep{lei2016goodness, wang2017likelihood}. In our setting, the discrete target is the revenue threshold $K^*$, and the proposed procedure provides uncertainty quantification for structural properties of the optimal assortment rather than only a point estimate of the assortment itself.}

In summary, the aforementioned hypotheses aim to test whether the optimal assortment satisfies a given property of interest. More generally, let $\boldsymbol{\cS}$ be the set of all nonempty offer sets and let $\boldsymbol{\cS}_0\subseteq\boldsymbol{\cS}$ be a subset of offer sets satisfying a property of interest. We study the hypothesis testing problem
\begin{equation}\label{eq: hypo test S S0}
    \mathrm{H}_0: \cS^*\in\bcS_0
    \qquad\text{versus}\qquad
    \mathrm{H}_1: \cS^*\notin\bcS_0.
\end{equation}
Examples 1 to 3 are concrete examples of the general hypothesis testing problem \eqref{eq: hypo test S S0}. In Section~\ref{sec: intro inf on s}, we will provide an equivalent representation of \eqref{eq: hypo test S S0} that will ease the computational difficulties caused by the combinatorial nature of assortment optimization and enable efficient inference.

\subsection{Major Contributions}

To the best of our knowledge, this paper provides the first inferential framework for performing general tests on structural properties of the optimal assortment under the MNL model. The proposed method can test a broad class of properties of the optimal assortment, and we provide theoretical guarantees for the validity of the resulting tests. We summarize our main contributions below.

First, we establish statistical rates for estimating the latent preference scores under the MNL model from general historical assortment data. Previous studies of MNL assortment optimization have largely focused on deterministic optimization or online learning policies, whereas our analysis characterizes estimation error from random choice data. Our convergence rate is consistent with previous results under related choice models \citep{gao2021uncertainty,chen2019topk}. Because regularized estimators generally suffer from bias \citep{zhang2013confint}, we also develop a debiasing procedure tailored to the MNL model, thereby enabling valid inferential analysis and serving as the foundation for our subsequent inference procedures.

Second, by reducing property testing for the optimal assortment to sign-change-point inference for marginal revenue gaps \citep{changepoint2022burg}, we provide a general framework for testing structural properties of the MNL optimum. Specifically, we develop a multiplier-bootstrap procedure that constructs a confidence interval for the optimal cutoff $K^*$ and then uses this interval to test whether the optimal assortment satisfies a given property. We also discuss concrete examples under the general framework and show how case-specific structure can simplify the inferential procedure while maintaining validity.

Third, we provide theoretical guarantees for the validity of the proposed inference procedure. The optimal-assortment inference problem differs from ranking problems studied in recent literature \citep{chen2019topk, gao2021uncertainty, liu2021lagrangian}. Ranking problems focus on latent preference scores, whereas assortment inference involves the interaction between preferences and revenues and concerns the output of a discrete optimization procedure. This interaction requires new uncertainty-quantification theory beyond existing ranking-inference tools.

\subsection{Literature Review}

The study of choice models has a long history. Regarding the optimization problem under the MNL model, apart from classic static assortment optimization \citep{ryzin1999relationship, mahajan2001stocking}, many variants have been proposed to incorporate additional information and make the model more realistic in practical scenarios. These include dynamic assortment optimization with unknown customer choice behavior \citep{CaroFelipe2007DAwD, RusmevichientongPaat2010DAOw, SaureDenis2013ODAP,AgrawalShipra2017TSft,ChenXi2018Anoa,WangYining2018NPfD,agrawal2019mnl,chen2020dynamicopt}, personalized assortment optimization integrating customer data \citep{golrezaei2014real,cheung2017thompson,chen2020assortment,chen2022statistical, ali2025clickbasedmnl}, robust assortment optimization allowing for model misspecification \citep{besbes2015surprising,chen2019robust}, multi-purchase MNL models \citep{bai2024multi}, MNL models with rank cutoffs \citep{bai2024assortment}, display optimization with sequential nonoverlapping offerings \citep{feldman2023display}, and models in which customers view only a subset of the offered assortment \citep{wang2018impact, gallego2020approximation,aouad2021display}. For example, \citet{cheung2017thompson} studied a personalized MNL model with known time-varying customer features, and \citet{ali2025clickbasedmnl} introduced a click-based MNL model in which customers consider only items they have clicked. To allow for outlier behavior, \citet{chen2019robust} adopted an $\varepsilon$-contamination model, in which a small proportion of the observations may be generated from an arbitrary contaminating distribution. 

Pricing optimization provides another perspective on revenue maximization and has generated a substantial literature \citep{li2020convex, yan2022representative, liu2022product}. For example, \citet{yan2022representative} developed a data-driven framework for multi-product pricing based on a separable representative consumer model, and \citet{liu2022product} studied a perturbed utility model for customer choice over primary products and ancillary services.

Since optimization problems in choice models usually require knowledge of latent parameters, recent work has developed estimation methods for choice data. In terms of preference-score estimation, the Bradley--Terry--Luce model \citep{chen2019topk} is a widely studied special case of the MNL model with pairwise comparisons and no no-purchase option. Many works focus on estimating preference scores and recovering rankings. For instance, \citet{chen2019topk} established optimality of the MLE and spectral method up to constants for exact top-$k$ recovery, and \citet{chen2022partial} refined the leading constants and established minimax partial-recovery rates. Other works generalize estimation to different choice models; for example, \citet{teo2012cmm} proposed a parsimonious discrete choice model and estimated choice probabilities through semidefinite optimization.

Apart from parameter estimation, uncertainty quantification has recently received attention in ranking problems. \citet{gao2021uncertainty} characterized asymptotic normality of the MLE and spectral estimator under the sparse BTL model, and \citet{liu2021lagrangian} applied a Lagrangian debiasing correction to perform inference on ranking properties. Compared with the rich literature on score estimation, inferential analysis remains limited in choice-model studies, especially for assortment optimization. This paper aims to bridge this gap by developing a general inferential procedure for structural properties of the optimal assortment.

\noindent\textbf{Paper Organization.} The rest of the paper is organized as follows. In Section~\ref{sec: prob setup}, we provide the preliminary setup for optimal assortment inference under the MNL model and introduce inference-related concepts. In Section~\ref{sec: mtd}, we propose our inferential procedure for testing combinatorial properties of the optimal assortment based on a debiased likelihood estimator. We also establish the convergence rate of the estimator and the validity of the inferential procedure. In Section~\ref{sec: numerical results}, we evaluate the performance of the proposed method in numerical experiments, followed by concluding remarks in Section~\ref{sec: conclusion}.

\noindent \textbf{Notations.} We denote by $|\cdot|$ the cardinality of a set. For two positive sequences $ x_n $ and $ y_n $, we denote $x_n\lesssim y_n$ or $x_n = O(y_n)$ if there exists a positive constant $C > 0$ independent of $n$ such that $x_n \le C y_n$ for all $n$ sufficiently large. We say $x_n\asymp y_n$ if $x_n \lesssim y_n$ and $ y_n \lesssim x_n $. If $\lim_{n \rightarrow \infty}x_n/y_n = 0$, then we say $x_n=o(y_n)$. For two integers $ j > i \ge 1 $, we denote by  $[i]$  the set $\{1,2,\ldots, i\}$,  by $i : j$ the set $\{ i, i+1, \ldots, j \}$, and by  $:$  the full index set. 
For a vector $\vb = (v_1, \ldots, v_d)^{\top}$, we use $\|\vb\|_q := \left(\sum_{i = 1}^d |v_i|^q \right)^{1/q}$ to denote the vector $\ell_q$-norm for an integer $q \ge 1$, and $\|\vb\|_{\infty} := \lim_{q \rightarrow \infty} \|\vb\|_q = \max_{i} |v_i|$ to denote the vector $\ell_{\infty}$-norm. For a matrix $\Ab = [A_{ij}]$, we use $\|\Ab\|_2$ to denote the matrix spectral norm, $\|\Ab\|_{\infty} := \sup_{\|\xb\|_{\infty} = 1} \|\Ab \xb\|_{\infty} = \max_{i} \sum_j |A_{ij}|$ to denote the matrix $\ell_{\infty}$-norm, and $\|\Ab\|_{2,\infty} := \sup_{\|\xb\|_{2} = 1} \|\Ab \xb\|_{\infty} = \max_{i} \|\Ab_i\|_2$ to denote the 2-to-$\infty$ norm. We denote by $\mathbf{1}_d$ a $d$-dimensional vector with all entries equal to 1, and we omit the subscript when the dimension is clear from the context.  We let $\{\eb_i\}_{i=1}^d$ be the canonical basis for $\RR^{d}$, where the dimension $d$ might change from place to place. Throughout the paper, we use $c, C$ to represent generic constants, whose values might change in different contexts.

\section{Problem Setup}\label{sec: prob setup}
We provide the problem formulation by briefly reviewing the assortment optimization problem under the multinomial logit model, followed by inferential analysis of the optimal assortment. Finally, we define the property sets corresponding to the optimal offer set to facilitate our discussions. 
\subsection{Multinomial Logit Choice Model}\label{sec: intro mc model}
We consider the multinomial logit (MNL) model, where the marginal probability of selecting a product from the offered assortment is proportional to the underlying preference scores associated with each product. More specifically, index the $n$ products by $[n] = \{1, \ldots, n\}$ and the no-purchase alternative by 0. Define $[n]_+ = [n] \cup \{0\}$. To each item $i \in [n]_+$, we assign a preference score $u_i^* > 0$ and denote by $\bu^* = (u_0^*, u_1^*, \ldots, u_n^*)^{\top}$ the preference score vector.
We let $\thetab^* = (\theta_0^*, \theta_1^*, \ldots, \theta_n^*)^{\top}$ be the log-transformation of $\bu^*$, where $\theta_i^* = \log (u_i^*)$ for $i \in [n]_+$. We set $\theta_0^* = 0$ to ensure identifiability, since the model is invariant to shifting all $\theta_i^*$'s by a constant. 
We let the condition number of the preference scores be 
$$\kappa_{\thetab} = \frac{\max_{i \in [n]_+} u_i^*}{\min_{i \in [n]_+} u_i^*}= \frac{\max_{i \in [n]_+} e^{\theta_i^*}}{\min_{i \in [n]_+} e^{\theta_i^*}} .$$
We denote by $\mathcal{S} \subseteq [n]$ the offered assortment, i.e., the available set of products, and define $\mathcal{S}_+ = \mathcal{S} \cup \{0\}$. Under the MNL model, recall that the probability of choosing item $j \in \mathcal{S}_+$ is 
\begin{equation}\label{eq:mnl_choice_prob}
	\PP(j\mid \mathcal{S}_+) = \frac{u_j^*}{1 +  \sum_{i \in \mathcal{S}} u_i^* } = \frac{\exp(\theta_j^*)}{1 +  \sum_{i \in \mathcal{S}} \exp(\theta_i^*) }, \quad \text{where } \theta_0^* = 0.
\end{equation}
{Throughout the paper, we assume $\kappa_{\thetab}=O(1)$. This boundedness condition is commonly imposed in the ranking-inference literature \citep{chen2015spectral, chen2019topk, gao2021uncertainty}. It excludes regimes in which some products are overwhelmingly more attractive than others, and ensures that historical comparisons among products remain informative. Indeed, if $\kappa_{\thetab}$ is unbounded, then there exists a pair $(i,j)$ such that $u_j^*\gg u_i^*$. For any offer set $\cS$ containing both $i$ and $j$, the MNL choice probabilities imply
\[
\PP\{\text{choose } i \mid \text{choose either } i \text{ or } j \text{ from } \cS_+\}
=
\frac{u_i^*}{u_i^*+u_j^*}
=o(1).
\]
Thus, item $i$ is almost never selected when it is compared with item $j$, and the historical choice data contain little effective information for distinguishing its preference score from those of other low-probability alternatives. From a practical perspective, such a regime corresponds to assortments with nearly degenerate relative demand, where comparisons across all products are not meaningful at the same statistical scale.}

Each item $i \in [n]_+$ is associated with a corresponding revenue parameter $r_i$, where $r_0 = 0$ for the no-purchase alternative. Without loss of generality, assume that we label the products such that $r_1 \ge r_2 \ge \ldots \ge r_n$.
We introduce the condition number for the revenues as $$\kappa_r = r_1 / r_n.$$
{A larger value of $\kappa_r$ indicates greater dispersion between the highest- and lowest-revenue products.} Given an offered assortment $\mathcal{S} \subseteq [n]$, recall the total expected revenue is defined as 
\begin{equation}\label{eq: def total exp rev}
	r(\mathcal{S}) = \sum_{i \in \mathcal{S}} \frac{u_i^* r_i}{1 + \sum_{j \in \mathcal{S}} u_j^*}.
\end{equation}
We are interested in finding an optimal assortment that maximizes the total expected revenue, formulated as:
\begin{equation}\label{eq: max rev}
	\mathcal{S}' \in \arg\max_{\mathcal{S} \in \bcS} r(\mathcal{S}),
\end{equation}
{ where $\bcS$ is the set of all non-empty subsets of $[n]$.}
Since there may not be a unique optimal assortment, we focus on the smallest optimal assortment, i.e., the optimal assortment with the smallest cardinality, which we denote by $\mathcal{S}^*$. 

Under the MNL model, the following theorem shows that the smallest optimal assortment $\mathcal{S}^*$ can be efficiently recovered at a low computational cost. 
{
	\begin{theorem}\label{thm: alg}[\cite{talluri2004revenue}]
		Under the MNL model, define $\Delta_k = \sum_{i = 1}^k r_i u_i^* - (\sum_{i=0}^k u_i^*)r_k $ for $k = 1,\ldots, n$. Then, $\Delta_k$ is non-decreasing with respect to $k \in [n]$, and the smallest optimal assortment $\mathcal{S}^*$ is of the form $\mathcal{S}^* = \{1,2,\ldots, K^*\}$, where $K^* = \max\{k \in [n]: \Delta_k < 0\}$. 
\end{theorem}}
{ The intuition of Theorem~\ref{thm: alg} is that we include item $k$ in the optimal assortment if its revenue $r_k$ exceeds the weighted average revenue of the items up to $k$, $(\sum_{i = 1}^k r_i u_i^*)/(\sum_{i=0}^k u_i^*)$. The choice of $\mathcal{S}^*$ as the optimal assortment with the smallest cardinality ensures the uniqueness of the assortment of interest and facilitates notational clarity. In practice, retailers often focus on assortments with smaller cardinalities to reduce inventory costs. However, if retailers are more interested in offering the largest optimal assortment to provide more options to customers, we redefine $K^* = \max\{k \in [n]: \Delta_k \le 0\}$.  Our subsequent theoretical results still hold.} { Note that the recovery of $\cS^*$ in Theorem~\ref{thm: alg} requires knowledge of $u_i^*$'s, which are not known to merchandisers in practice.  We will discuss in Section~\ref{sec: mtd} how to obtain $\cS^*$ when $u_i^*$'s are unknown.} With the concrete form of $\mathcal{S}^*$ in place, we are now ready to introduce the inference problem on $\mathcal{S}^*$.
\if1\mcmodel{
	We consider the Markov-chain-based choice model for the purchase behavior. In the Markov chain based model, each state corresponds to an item (a product or the no-purchase alternative). The customer would initially arrives at the state of a most preferred item proportionally to its preference score, and if the item is unavailable, the customer will transition to another item following the Markov chain transition probabilities. The customer will keep transitioning until arriving at an available item or the no-purchase alternative. More specifically, a customer would initially choose each item $i \in [n]_+$ with arrival probability $\lambda_i = u_i^*/ \sum_{j =0}^n u_j^*$ ($i=0$ suggests that the customer does not purchase any product). If $i \in \mathcal{S} \cup \{0\}$, the customer will stay with item $i$. If $i \in [n] \backslash \mathcal{S}$, the customer will transition from item $i$ to item $j \in [n]_+ \backslash \{i\}$ with transition probability $\rho_{ij}(\mathcal{S})$ and repeat the substitution process until the customer chooses a product that is in the set $\mathcal{S}_+ := \mathcal{S} \cup \{0\}$, i.e., either available products or the no-purchase alternative. Under the MNL model, the transition probabilities are defined as following 
	\begin{align*}
		\rho_{i j}(\mathcal{S})=\left\{\begin{array}{ll}
			0 & \text { if } i \in \mathcal{S}_+, j \neq i \\
			1 & \text { if } i \in \mathcal{S}_+, j=i \\
			\frac{u_j^*}{\sum_{k \neq i}u_k^*} & \text { otherwise. }
		\end{array}\right.
	\end{align*}
	We can see that each item $i \in \mathcal{S}_+$ is an absorbing state. After proper permutations of the rows and columns, the transition matrix can be written in the following block matrix form: 
	\[ \Pb (\mathcal{S}) = \left\{\rho_{i j}(\mathcal{S})\right\} = 
	\begin{pmatrix}
		\Ib & \mathbf{0}\\
		\Bb & \Cb
	\end{pmatrix}.
	\]
	We denote by $\pi(j, \mathcal{S})$ the probability that the customer finally arrives at absorbing state $j \in \mathcal{S}_+$. According to Theorem 3.1 in \cite{blanchet2016mcpaper}, we have
	$$\pi(j, \mathcal{S}) = \lim_{q \rightarrow \infty} \lambdab^{\top}\Pb(\mathcal{S})^q \eb_j = u_j^*/ (\sum_{i \in \mathcal{S}_+} u_i^*),\quad j \in \mathcal{S}_+,$$
	where $\lambdab = (\lambda_0, \lambda_1,  \ldots, \lambda_n)^{\top}$ is the vector of initial arrival probabilities.
	For each item $i \in [n]_{+}$, let $r_i$ denote the associated revenue ($r_0 = 0$), and the total expected revenue for the offer set $\mathcal{S}$ can be calculated as 
	$$r(\mathcal{S}) = \sum_{j \in \mathcal{S}} r_j \cdot \pi(j, \mathcal{S}) = \lim_{q \rightarrow \infty} \lambdab^{\top} \Pb(\mathcal{S})^q \rb, \quad \rb = (r_0, r_1, \ldots, r_n).$$
	We are interested in the optimal offer sets that maximize the total expected revenue. To study the optimal sets, same as in \cite{blanchet2016mcpaper}, we introduce $g_i(\mathcal{S}) := \lim_{q \rightarrow \infty} \eb_i^{\top} \Pb(\mathcal{S})^q \rb$, which is the expected revenue when the customer initially arrives at item $i \in [n]_+$. By the structure of $\Pb(\mathcal{S})$, it is not hard to see that 
	\begin{align*}
		g_i(\mathcal{S})=\left\{\begin{array}{ll}
			r_i & \text { if } i \in \mathcal{S}_+ \\
			\sum_{j \in [n]} \rho_{ij}(\mathcal{S})g_j(\mathcal{S}) & \text { otherwise, }
		\end{array}\right.
	\end{align*}
	and $r(\mathcal{S}) = \lambdab^{\top} \gb(\mathcal{S})$, where $\gb(\mathcal{S}) := (g_0(\mathcal{S}), g_1(\mathcal{S}), \ldots, g_n(\mathcal{S}))^{\top}$. For $i \in [n]$, we define $ g_i = \max_{\mathcal{S} \subseteq [n]} g_i(\mathcal{S})$, and denote $\Pb_0 = (\mathbf{1}^{\top} \ub^*)^{-1} \mathbf{1} \ub^{*\top} \in \RR^{n \times n}$ with $\ub^* = (u_1^*, \ldots, u_n^*)^{\top}$. Then from Theorem 3.1 and Lemma 5.2 in \cite{blanchet2016mcpaper} we know that under the condition that $ \sqrt{n}(\mathbf{1}^{\top}\ub^*)^{-1} \|\ub^*\| < 1$, $\gb = (g_1, \ldots, g_n)^{\top}$ is the unique solution to the following equation
	$$
	\gb = \max(\rb, \Pb_0 \gb).
	$$
	Now define $\mathcal{S}^* = \{i \in [n] : g_i = r_i \}$, then according to Theorem 5.1 in \cite{blanchet2016mcpaper}, $g_i = g_i(\mathcal{S}^*)$ for $i \in [n]$ and $\mathcal{S}^* \in \operatorname{argmax} r(\mathcal{S})$. It is not hard to see that for $i \in [n]\backslash \mathcal{S}^*$, $g_i > r_i$, and thus for any $\mathcal{S} \subseteq [n]$ such that $\mathcal{S} \cap [n] \backslash \mathcal{S}^* \neq \emptyset$, we have that $g_i(\mathcal{S}) = r_i < g_i, \forall i \in \mathcal{S} \cap [n] \backslash \mathcal{S}^*$, and $r(\mathcal{S}) < r(\mathcal{S}^*)$. Thus we can see that $\mathcal{S}^* = \cup_{\mathcal{S} \in \operatorname{argmax}r(\mathcal{S})} \mathcal{S}$. In other words, $\mathcal{S}^*$ is the largest optimal set. }\fi
\subsection{Inference on the Optimal Offer Set}\label{sec: intro inf on s}

Recall that we aim to conduct inference on the smallest optimal offer set $\mathcal{S}^*$. In particular, we seek to conduct a general hypothesis test as proposed in \eqref{eq: hypo test S S0}. We aim to propose a general inferential procedure that controls the asymptotic Type I error under any given level $\alpha \in (0, 1)$ with diminishing Type II error. {By Theorem~\ref{thm: alg}, under the MNL model the smallest optimal assortment is revenue-ordered: it contains exactly the products whose revenues exceed the relevant weighted average revenue threshold. Hence, the optimal assortment can be represented by a single cutoff index \(K^*\), with \(\cS^*=[K^*]\).} Since knowing $\mathcal{S}^*$ and knowing $K^*$ are equivalent, we can see that the hypothesis testing in \eqref{eq: hypo test S S0} essentially translates into testing whether $K^*$ satisfies certain properties, e.g., Example 3 in Section~\ref{sec: intro} tests whether all the first $K^*$ products belong to a certain set $A$. Then we can reformulate the testing problem in \eqref{eq: hypo test S S0} as follows: for a given set $\mathcal{K}_0 \subseteq [n]$, we test whether $K^*$ belongs to $\mathcal{K}_0$,
\begin{equation}\label{eq: hypo test K S0}
	\mathrm{H}_0: K^* \in \mathcal{K}_0 \text{~versus~}
	\mathrm{H}_1: K^* \notin \mathcal{K}_0.
\end{equation}
We call $\mathcal{K}_0$ the property set of $K^*$. We formally introduce below the applications provided in Section~\ref{sec: intro} along with other examples and define the corresponding property set $\mathcal{K}_0$ as concrete formulations of \eqref{eq: hypo test K S0}.
\begin{example}\label{exm 1}
	For a given product $i \in [n]$, we test if $i$ is in the optimal assortment $\mathcal{S}^*$, i.e.,
	$$
	\mathrm{H}_0: i \not\in \mathcal{S}^* \text{~versus~} \mathrm{H}_1:  i \in  \mathcal{S}^*.
	$$
	{ The corresponding property set is  $\mathcal{K}_0 = \{k \in [n] :  k \le i-1\}$.}
\end{example}
\begin{example}\label{exm 2}
	For a given set $A \subseteq [n]$, we test if $A$ is a subset of the optimal assortment $\mathcal{S}^*$, i.e.,
	$$
	\mathrm{H}_0: A \nsubseteq \mathcal{S}^* \text{~versus~} \mathrm{H}_1:  A \subseteq  \mathcal{S}^*.
	$$
	The corresponding property set is $\mathcal{K}_0 = \{k \in [n] : k \le (\max_{i \in A} i) - 1 \}$.
\end{example}
\begin{example}\label{exm 3}
	For a given set $A \subseteq [n]$, we test if the optimal assortment $\mathcal{S}^*$ { is a subset of} $A$, i.e.,
	$$
	\mathrm{H}_0: \mathcal{S}^* \subseteq A \text{~versus~} \mathrm{H}_1:  \mathcal{S}^* \nsubseteq  A.
	$$
{The corresponding property set is $\mathcal{K}_0 = \{k \in [n] : [k] \subseteq A\}$.}
\end{example}
\begin{example}\label{exm 4}
	For a given set $A \subseteq [n]$ and a given ratio $q \% $, we test if the proportion of $\mathcal{S}^*$ contained in $A$ is larger than $q \%$, i.e.,
	$$
	\mathrm{H}_0: \lvert A \cap \mathcal{S}^*\rvert/\lvert\mathcal{S}^*\rvert > q \% \text{~versus~} \mathrm{H}_1:  \lvert A \cap \mathcal{S}^*\rvert/\lvert\mathcal{S}^*\rvert \le q \%.
	$$
	The corresponding property set is $\mathcal{K}_0 = \{k \in [n] : |[k]\cap A|/k > q \%\}$.
\end{example}
\begin{example}\label{exm 5}
	For a partition $\{A_1, \ldots, A_m\}$ of the products, i.e., $\bigcup_{j=1}^m A_j = [n]$ and $A_j \cap A_k = \emptyset$ for $j \neq k$, { we test whether the products in $A_1$ constitute the largest proportion of the optimal assortment $\mathcal{S}^*$ compared with all the other $A_j$'s}, i.e.,
	$$
	\mathrm{H}_0: \lvert \mathcal{S}^* \cap A_1 \rvert = \max_j \lvert \mathcal{S}^* \cap A_j \rvert \text{~versus~}
	\mathrm{H}_1: \lvert \mathcal{S}^* \cap A_1 \rvert < \max_j \lvert \mathcal{S}^* \cap A_j \rvert.
	$$
	The corresponding property set is $\mathcal{K}_0 = \{k \in [n] : |[k]\cap A_1| = \max_j |[k]\cap A_j|\}$.
\end{example}
\begin{example}\label{exm 6}
	For a partition $\{A_1, \ldots, A_m\}$ of the products, i.e., $\bigcup_{j=1}^m A_j = [n]$ and $A_j \cap A_k = \emptyset$ for $j \neq k$, we test if no less than $n_0$ out of the $m$ subsets $A_j$'s contain products that constitute the optimal assortment $\mathcal{S}^*$, where $n_0$ is a given count, i.e.,
	\begin{align*}
		&\mathrm{H}_0: \lvert\{ j \in [m] : A_j \cap \mathcal{S}^* \neq \emptyset\}\rvert \ge n_0,\\
		&\mathrm{H}_1: \lvert\{ j \in [m] : A_j \cap \mathcal{S}^* \neq \emptyset\}\rvert < n_0.
	\end{align*}
	The corresponding property set is $\mathcal{K}_0 = \{k \in [n] : |\{j \in [m] { :} \min_{i \in A_j} i \le k\}|\ge n_0\}$.
\end{example}
We can see that hypothesis testing on $\mathcal{S}^*$ in essence tests whether $K^*$ satisfies the properties of interest. In practice, the parameter $\thetab^*$ is unknown, and we will approximate $\thetab^*$ by its estimator $\hat\thetab$ obtained from the observed data and estimate $\bu^*$ by the plug-in $\hat\bu = \exp(\hat\thetab)$. Then we can estimate $\mathcal{S}^*$ by replacing $\Delta_k$ with $\hat\Delta_k = \sum_{i = 1}^k r_i \hat{u}_i - (\sum_{i=0}^k \hat{u}_i)r_k $ in the algorithm proposed in Theorem \ref{thm: alg}. Furthermore, we can
construct a confidence interval $ [\hat{K}_L, \hat{K}_U]$ of confidence level $1-\alpha$ for $K^*$ by quantifying the uncertainty of $\{\hat\Delta_k\}_{k \in [n]}$. Subsequently, we will reject ${\rm H}_0$ if $[\hat{K}_L, \hat{K}_U] \cap \mathcal{K}_0 = \emptyset$. 

\section{Method}\label{sec: mtd}
In this section, we propose an algorithm for conducting general inference on the optimal assortment $\cS^*$. We begin by introducing how to estimate the latent preference scores from the observed data via penalized optimization, and then provide a debiasing approach for the resulting estimator. Then we propose the inferential procedure based on multiplier bootstrap to test arbitrary properties of the optimal offer set $\cS^*$. 
\subsection{Observed Data and Latent Score Estimation}
{ 
	We adopt an observational study approach to establish our framework, utilizing available historical data on customer choice outcomes for estimation and subsequent inference. Observational studies are commonly used in ranking problems and other operations research studies due to their ability to leverage historical data when the active collection of real-time samples is costly or infeasible \citep{Nikolaev2013, jena2019partiallyranked, chen2019topk, qiu2021multi, gao2021uncertainty, flamand2023store}. Compared to online learning approaches, where data is actively collected based on revenue outcomes or real-time analysis \citep{agrawal2019mnl, chen2020dynamicopt, miao2022online}, observational studies are less adaptive but offer distinct advantages. These include cost-effectiveness, as they use existing data without incurring the expenses of new data collection, and the ability to analyze large datasets accumulated over historical periods.
	
	We first discuss the assumptions on the observed historical data. Since there is an exponential number of possible assortments in total, in reality, merchandisers usually observe choice results only for a subset of offer sets among all possible assortments.} The goal is to estimate the latent preference scores from the historical data and apply the estimators to the subsequent inference on $\cS^*$. 


More specifically, { recall that $\bcS = \{\cS \subseteq [n]: \cS \ne \emptyset\}$ is the set of all offer sets.} { We assume that the retailers only observe a subset of assortments in $\bcS$.}
Specifically, for each $\cS \in \bcS$, we define $\cE_{\cS} = 1$ if the offer set $\cS$ is observed in the historical data
and $\cE_{\cS} = 0$ otherwise, 
and we denote by $\boldsymbol{\cE} = \{\cS \in \bcS  : \cE_{\cS} = 1\}$ the set of observed offer sets.  
{ For each $\cS \in \bcE$, we observe $L_{\cS}$ independent choice outcomes among the items in $\cS_+$, and let $L := \sum_{\cS \in \bcE} L_{\cS}$ be the total sample size for all observed assortments.} 

{
To ensure that the observed data are informative about the latent preference scores, we impose a connectivity assumption on the observed offer sets $\bcE$, i.e., no item is isolated or excluded from comparisons, and the induced comparison hypergraph is connected. This assumption is natural and necessary for valid estimation and inference. Indeed, if purchase outcomes are observed only for a strict subset $\cI \subsetneq [n]$, then the latent scores $\theta_i^*$ for $i \in [n]\setminus \cI$ are not identifiable. Likewise, if the comparison graph is disconnected, latent score estimation is impossible because the lack of comparisons across components also renders the model unidentifiable.
To study the connectivity structure, we introduce a weighted Laplacian as the matrix representation of the comparison graph induced by the observed offer sets. In graph theory, the Laplacian is a matrix representation of a graph. In the pairwise-comparison case, let $\Ab=\{A_{ij}\}_{i<j}$ be the symmetric adjacency matrix, {where $A_{ij}=1$ if items $i$ and $j$ are jointly observed in at least one historical assortment, and $A_{ij}=0$ otherwise.} The corresponding graph Laplacian is $\Lb=\Db-\Ab$, where $\Db$ is diagonal with its $i$th entry equal to the degree of node $i$ \citep{Chung1997}. In our setting, comparisons arise from multi-item offer sets, and the resulting structure is an Erd\H{o}s–Rényi–like hypergraph with products as nodes and each observed offer set as a hyperedge connecting multiple nodes. We characterize connectivity through the following weighted Laplacian constructed by a clique expansion within each observed offer set.
\begin{definition}\label{def: weighted graph lap}
    For the observed assortment set $\bcE$, let $L = \sum_{\cS \in \bcE} L_{\cS}$ be the total sample size. We define the weighted comparison-graph Laplacian as follows:
\begin{equation}\label{eq: con graph mat}
    \Lb_{\bcE} = \sum_{\cS \in \bcE} \bigg\{\frac{L_{\cS}}{L \cdot |\cS_+|^2} \sum_{i < j,\ i , j \in \cS_+} (\eb_i - \eb_j)(\eb_i - \eb_j)^{\top}\bigg\},
\end{equation}
where $\eb_i$ is the $i$th standard basis vector and $|\cS_+|$ denotes the cardinality of $\cS_+$.
\end{definition}
\begin{remark}
    The matrix $\Lb_{\bcE}$ is essentially the weighted Laplacian of the complete graph on each $\cS \in \bcE$, weighted by $L_{\cS}/(L \cdot |\cS_+|^2)$. Intuitively, the number of induced pairwise comparisons within $\cS_+$ scales with $|\cS_+|^2$, so the effective sample size per comparison is proportional to $L_{\cS}/|\cS_+|^2$. Hence, the larger $L_{\cS}/|\cS_+|^2$ is, the more information $\cS$ provides, and therefore it should be assigned more weight.
\end{remark}
We impose the following assumption on the graph Laplacian to guarantee comparison-graph connectivity.
\begin{assumption}\label{asp: connectivity graph}
 Let
$\bar\lambda:=\lambda_{\max}(\Lb_{\bcE})$, $\underline\lambda:=\lambda_{n}(\Lb_{\bcE})$, and $\kappa_{\bcE}:=\bar\lambda/\underline\lambda$. Assume that $\bar\lambda \ge \underline{\lambda} > 0$.
\end{assumption}
\begin{remark}
    It is immediate that $\lambda_{n+1}(\Lb_{\bcE})=0$ because $\Lb_{\bcE}\,\mathbf{1}_{n+1}=\mathbf{0}$, and Assumption~\ref{asp: connectivity graph} guarantees that the comparison graph is connected by imposing that the second-smallest eigenvalue of $\Lb_{\bcE}$ is positive. This is analogous to the pairwise-comparison graph Laplacian, for which it is well known that the rank of the Laplacian equals the number of vertices minus the number of connected components \citep{Chung1997}, and a positive second-smallest eigenvalue guarantees that the entire graph is connected. {
In the present MNL setting, the no-purchase option belongs to every observed offer set. Therefore, if every product is offered at least once, the induced comparison structure is connected in a binary sense through the no-purchase option. Assumption~\ref{asp: connectivity graph}, however, is stronger than this binary connectivity requirement: the estimation rate depends on the magnitude of the weighted connectivity, as captured by the second-smallest eigenvalue $\underline{\lambda}$ of $\Lb_{\bcE}$. Products that are rarely offered, or offered only in assortments with small weights $L_{\cS}/(L|\cS_+|^2)$, can lead to a small $\underline{\lambda}$ even when the induced graph is connected in the unweighted sense. Thus, Assumption~\ref{asp: connectivity graph} should be interpreted as an informativeness condition on the frequency and balance of historical comparisons, rather than merely as a requirement that the binary comparison graph be connected.
}
\end{remark}
While Assumption~\ref{asp: connectivity graph} provides a general condition guaranteeing connectivity of the comparison graph, we include the following lemma to provide a concrete example in which Assumption~\ref{asp: connectivity graph} is satisfied. Because we study the MNL model in which the optimal assortments are revenue-ordered, we focus on the case where the historical data consist only of revenue-ordered assortments.
\begin{definition}[Revenue-ordered assortments] We say that the {historical data consist only of revenue-ordered assortments} if \(\bcE \subseteq \{[k]: k\in[n]\}\).
\end{definition}

\begin{lemma}\label{prop: rev order assort}
Assume that the historical data consist only of revenue‑ordered assortments. For each $k \in [n]$, denote $m_k = L_{[k]} / L$, where $L = \sum_{k=1}^n L_{[k]}$ is the total sample size of all observed revenue‑ordered assortments. Then we have $\bar\lambda \asymp \sum_{k=1}^n m_k/k$ and $\underline{\lambda} \asymp m_n / n$. Moreover, if $\sum_{k=1}^{n-1} m_k / k \lesssim m_n/ n$, then we have $\bar\lambda \asymp \underline\lambda \asymp m_n / n$ and $\kappa_{\bcE} = O(1)$.
\end{lemma}
\begin{remark}
    The proof is deferred to Supplementary Materials~\ref{sec: proof prop rev order assort}. Lemma~\ref{prop: rev order assort} states that when all observed assortments in $\bcE$ are revenue-ordered, the bottleneck of the comparison graph is the effective sample size of the assortment $[n]$, which is the only assortment that includes comparisons involving item $n$. The simplest $\bcE$ satisfying Assumption~\ref{asp: connectivity graph} is the singleton set $\bcE=\{[n]\}$, containing only the revenue-ordered assortment $\cS=[n]$. This is a mild condition on the observed data, because the retailer can gather historical purchasing outcomes on $\cS=[n]$ to draw inference on properties of the true optimal assortment $\cS^*$. Moreover, choosing $\cS=[n]$ is natural and without loss of generality: if prior knowledge implies that $|\cS^*| \le n' < n$, we can simply exclude items $i \in [n]\setminus [n']$ as irrelevant to $\cS^*$ and reset $n$ to $n'$.
\end{remark}
Lemma~\ref{prop: rev order assort} shows that Assumption~\ref{asp: connectivity graph} is satisfied when the observed assortments contain only revenue-ordered offer sets, provided that $[n]$ is included in $\bcE$ with sufficient sample size. In the BTL literature \citep{chen2019topk}, pairwise comparisons are often assumed to be independently sampled across pairs. Although this assumption on the assortment-sampling scheme is unnecessary for our theoretical guarantees, we also provide an auxiliary lemma in Supplementary Materials~\ref{sec: lm eigen of L cE} that presents an analogous result when \(\cS \in \bcE\) is sampled independently from \(\bcS\) and Assumption~\ref{asp: connectivity graph} holds.
}

{ The remainder of the technical results are established for the observed set of assortments $\bcE$. Denote $x_{\cS}^{(i, \ell)} = \II\{o_{\cS}^{(\ell)} = i\}$, where $o_{\cS}^{(\ell)} \in \cS_+$ is the choice outcome of the $\ell$-th customer from offer set $\cS$. Under the MNL model, $o_{\cS}^{(\ell)}$ follows the multinomial distribution satisfying
					
					\begin{equation}
						o_{\cS}^{(\ell)} = i \text { with probability } 
						\frac{u_i^*}{1 + \sum_{j \in \cS} u_j^*}= \frac{e^{\theta_i^*}}{ 1+ \sum_{j \in \cS} e^{\theta_j^*}}
						,\quad 
						\text{ for $i \in \cS_+$ and $\cS \in \bcS$},
					\end{equation}
					where 
					we let $\theta_0^* = 0$ for identifiability.
}

 Then the negative log-likelihood function can be written as 
\begin{equation}\label{eq: log-likelihood}
	\cL(\thetab; \bx) = - \sum_{\cS \in \bcE} \left\{ \sum_{i \in \cS_+} x_{\cS}^{(i)}\theta_i - \frac{L_{\cS}}{L}\log \left(1+\sum_{i \in \cS}e^{\theta_i}\right) \right\}, \quad \theta_0 = 0,
\end{equation}
where $x_{\cS}^{(i)} = \frac{1}{L}\sum_{\ell = 1}^{L_{\cS}} x_{\cS}^{(i, \ell)}$. Let $\bx = \{x_{\cS}^{(i,\ell)}\}$. We denote by $\hat\thetab$ the penalized maximum likelihood estimate (MLE), which solves the following convex problem
\begin{equation}\label{eq: convex prob 0}
	\min_{\thetab \in \RR^{n+1}, \theta_0 = 0} \cL_{\tau}(\thetab; \bx) := \cL(\thetab; \bx) + \frac{\tau}{2}\sum_{i \in [n]_+} (\theta_i - \bar{\theta})^2,
\end{equation}
where $\bar\theta = \sum_{i \in [n]_+} \theta_i / (n+1)$, and {$\tau>0$} is a tuning parameter. Here, { we apply regularization to the negative log-likelihood function because, due to the identifiability constraint, the Hessian matrix of the negative log-likelihood function is singular. We need regularization to enforce convexity of the objective function and to guarantee that there is a unique maximizer. Moreover,} we regularize over the sample variance of $\theta_i$'s
rather than the $\ell_2$-norm of $\thetab$ because the regularization over the sample variance of $\theta_i$'s will prevent $\theta_i$'s from getting too far away from their mean and ensure the exchangeability of all $\theta_i$'s, which is consistent with the MNL model.
We characterize the convergence rate of $\hat\thetab$ in the following theorem.

\begin{theorem}\label{thm: inf norm MLE}
	Recall that $\hat\thetab$ is the regularized MLE. { Suppose that Assumption~\ref{asp: connectivity graph} holds.} Under the conditions that { $\tau \asymp  \sqrt{{\bar\lambda \log n}/{L}}$ and $\sqrt{{n \kappa_{\bcE} \log n}/{(\underline{\lambda}L)}} \le c$ }for some small enough constant $c > 0$
	, with probability at least $1-O(n^{-10})$ we have that
	\begin{equation}\label{eq: inf norm MLE}
		\|\hat\thetab - \thetab^*\|_{2} \lesssim  { \sqrt{\frac{n \kappa_{\bcE} \log n}{\underline{\lambda}L}} .}
	\end{equation}
	
\end{theorem}
We refer interested readers to Supplementary Materials \ref{sec: proof thm inf norm mle} for the proof of Theorem~\ref{thm: inf norm MLE}.
\begin{remark}\label{rmk: connectivity p}
{The purpose of Theorem~\ref{thm: inf norm MLE} is to provide an explicit error rate for the likelihood-based estimator under general historical assortment data. The regularization handles the nonidentifiability of the MNL likelihood up to additive shifts and yields a well-conditioned optimization problem on the centered parameter space; when the tuning parameter is chosen at the order specified in Theorem~\ref{thm: inf norm MLE}, the resulting estimator is expected to be statistically comparable to the ordinary MLE in identifiable formulations. The rate in~\eqref{eq: inf norm MLE} is governed by the effective information level \(L\underline{\lambda}\), where \(\underline{\lambda}\) is the second-smallest eigenvalue of the weighted comparison-graph Laplacian \(\Lb_{\bcE}\). A larger value of \(L\underline{\lambda}\) corresponds to richer and more balanced historical comparisons across products, whereas a small \(\underline{\lambda}\) reflects weak overlap or highly uneven exposure in the observed assortments. Thus, the weighted comparison-graph condition summarizes the informativeness of the historical assortment variation for estimating relative preference scores.
}
To provide a concrete understanding of \eqref{eq: inf norm MLE}, we consider the scenarios in Lemma~\ref{prop: rev order assort} and Lemma~\ref{lm: eigen of L cE} under which Assumption~\ref{asp: connectivity graph} holds. When the observed assortments consist solely of the revenue-ordered offer set \([n]\), i.e., \(\bcE=\{[n]\}\) and \(L=L_{[n]}\), Lemma~\ref{prop: rev order assort} implies \(\bar\lambda \asymp \underline{\lambda} \asymp 1/n\) and \(\kappa_{\bcE}=O(1)\), so the rate in \eqref{eq: inf norm MLE} simplifies to \(n\sqrt{\log n / L}\).
When \(\cS \in \bcE\) is sampled independently, for simplicity take \(p_{\cS}=p\) for all \(\cS \in \bcS\) and assume the conditions in Lemma~\ref{lm: eigen of L cE} hold; then the rate in \eqref{eq: inf norm MLE} becomes \(n\sqrt{\log n / (2^{n} p L)}\), which is analogous to results in the BTL literature \citep{chen2019topk}.
\end{remark}

\subsection{Newton Debiasing with Centralization}\label{sec: lagrangian debias}
Recall that $\hat\thetab$ is the regularized MLE solving the problem \eqref{eq: convex prob 0}, and we have the gradient and Hessian of $\cL(\thetab; \bx)$ as 
{ 
\begin{equation}\label{eq: gradient likelihood}
	\nabla \cL(\thetab; \bx) = - \sum_{\cS \in \bcE} \left\{ \sum_{i \in \cS_+} \left(x_{\cS}^{(i)} - \frac{L_{\cS}}{L} \cdot \frac{e^{\theta_i}}{\sum_{j \in \cS_+} e^{\theta_j}}\right) \eb_i \right\};
\end{equation}
\begin{equation}\label{eq: hessian likelihood}
	\nabla^2 \cL(\thetab; \bx) = \sum_{\cS \in \bcE} \frac{L_{\cS}}{L}\left\{ \sum_{i \in \cS_+} \frac{e^{\theta_i}(\sum_{j \in \cS_+, j \neq i} e^{\theta_j})}{\left(\sum_{j \in \cS_+} e^{\theta_j}\right)^2}\eb_i\eb_i^{\top} - \sum_{\substack{i,s\in \cS_+\\i\neq s}} \frac{e^{\theta_i}e^{\theta_s}}{\left(\sum_{j \in \cS_+} e^{\theta_j}\right)^2}\eb_i \eb_s^{\top} \right\}.
\end{equation}}

{ As our goal is to perform inference on $\mathcal{S}^*$ based on $\hat{\boldsymbol{\theta}}$ and provide uncertainty quantification beyond estimation, despite the consistency of $\hat{\boldsymbol{\theta}}$ as demonstrated in Theorem~\ref{thm: inf norm MLE}, we need to apply extra debiasing and uncertainty quantification to $\hat{\boldsymbol{\theta}}$ to ensure its validity for subsequent inference. In particular, as shown in previous literature, the regularized estimation will induce bias in the estimator \citep{zhang2013confint}. Therefore, to conduct inferential analysis based on $\hat{\boldsymbol{\theta}}$, we need to correct for this bias first.} We debias the regularized MLE $\hat\thetab$ by a one-step Newton correction with centralization, expressed as 
\begin{equation}\label{eq: debias mle}
	\hat\thetab^d  = \hat\thetab' - \nabla^2 \cL(\hat\thetab; \bx)^{\dagger}  \nabla \cL(\hat\thetab; \bx) ,
\end{equation}
where $\nabla^2 \cL(\hat\thetab; \bx)^{\dagger}$ is the Moore-Penrose inverse of the Hessian evaluated at $\hat\thetab$ and $\hat\thetab'$ is the centralized MLE. Namely, $\hat\theta_i' = \hat\theta_i - (n+1)^{-1} \sum_{j \in [n]_+} \hat\theta_j$ for $i \in [n]_+$. 

Here we apply centralization to $\hat\thetab$ because by \eqref{eq: hessian likelihood}, we have $\nabla^2 \cL(\thetab; \bx)^{\dagger} \mathbf{1}_{n+1} = \mathbf{0}$ for any $\thetab \in \RR^{n+1}$, which indicates that due to the singularity of the Hessian matrix, the Newton correction corrects bias only along the direction perpendicular to $\mathbf{1}$. Hence to eliminate the bias in the direction of $\mathbf{1}$, we will first project $\hat\thetab$ onto the perpendicular space of $\mathbf{1}$, i.e., apply centralization to $\hat\thetab$.  Besides, from \eqref{eq: gradient likelihood} and \eqref{eq: hessian likelihood} we can observe that $\nabla \cL(\thetab; \bx)$ and $\nabla^2 \cL(\thetab; \bx)$ are invariant after adding a constant to $\thetab$, and hence the projection of $\hat\thetab$ does not change the likelihood evaluation.



With the debiased estimator in place, we are now ready to present the inferential procedures. Recall from Section~\ref{sec: intro inf on s} that under the MNL model, the inference on the optimal offer set $\cS^*$ is equivalent to the inference on its cardinality $K^*$. Furthermore, a product $k \in [n]$ is in $\cS^*$ if and only if $\Delta_k < 0$, where $\Delta_k$ is defined in Theorem~\ref{thm: alg}. As the mapping from $\cS^*$ to $\thetab^*$ is combinatorial, inferring $\cS^*$ directly from the estimators of $\thetab^*$ is challenging without the aid of $\Delta_k$'s. Thus, $\Delta_k$'s serve as pivotal intermediate quantities in the inference of $\cS^*$, and to provide uncertainty quantification for $\cS^*$, we first provide uncertainty quantification for $\Delta_k$'s.

More specifically, for $k \in [n]$, we define $\hat\Delta_k = \sum_{i = 1}^k r_i \exp(\hat\theta^d_i) - \left(\sum_{i = 0}^k \exp(\hat\theta^d_i)\right)r_k$, which is the estimator for $\Delta_k$ by plugging in the debiased estimator $\hat\thetab^d$. { Note that the centralization of $\hat\thetab$ when calculating $\hat\thetab^d$ will scale the plug-in estimators $\hat\Delta_k $'s by a positive factor.  However, since we are only interested in the signs of $\hat\Delta_k $'s, the scaling of  $\hat\Delta_k $'s bears no impact on the inference.} { Hence with a slight abuse of notation, we redefine
	$$
	\Delta_k =  \sum_{i = 1}^k r_i \exp(\theta^*_i - \bar\theta^*) - \left(\sum_{i = 0}^k \exp(\theta^*_i - \bar\theta^*)\right)r_k = \exp(-\bar\theta^*)\left( \sum_{i = 1}^k r_i u_i^* - (\sum_{i=0}^k u_i^*)r_k\right) ,  \quad k \in [n],
	$$
	where  $\bar\theta^* = (n+1)^{-1}\sum_{j \in [n]_+} \theta^*_j $.} The following theorem depicts the asymptotic distribution of $\hat\Delta_k$'s.
\begin{theorem}\label{thm: asymp normal}
	Under the same conditions as in Theorem~\ref{thm: inf norm MLE}, suppose that $\kappa_r=r_1/r_n \lesssim \sqrt{n}$ and { ${\kappa_{\bcE}^3 n \log n}{\sqrt{\kappa_{\bcE} \cdot n/(\underline{\lambda} L})}= o(1)$}. Then, for any $k \in [n]$, we have that 
	\begin{equation}
		\sqrt{\frac{L}{\vb_k^{\top} \nabla^2 \cL(\thetab^*; \bx)^{\dagger} \vb_k}} (\hat\Delta_k - \Delta_k)\overset{d}{\longrightarrow} N(0,1),
	\end{equation}
	where $\vb_k = { \exp(-\bar\theta^*)}\cdot \big({(0-r_k) u_0^*, (r_1 - r_k)u_1^*, (r_2 - r_k)u_2^*, \ldots, (r_{k-1}- r_k)u_{k-1}^*}, 0, \ldots, 0\big)^{\top}$.
\end{theorem}
The proof of Theorem \ref{thm: asymp normal} is deferred to Supplementary Materials \ref{sec: proof thm asymp normal}. Compared with Theorem~\ref{thm: inf norm MLE}, we have an extra scaling condition on the condition number $\kappa_r$ to ensure that the revenue parameters are not ill-conditioned. Besides, we have a stronger scaling condition on the sample size $L$ to guarantee the distributional convergence. Since $\thetab^*$ is unknown in practice, we estimate the asymptotic variance by plugging in the regularized MLE $\hat\thetab$. The following corollary of Theorem \ref{thm: asymp normal} shows the validity of the plug-in estimate.  
\begin{corollary}\label{col: var est}
	Under the same conditions as Theorem \ref{thm: asymp normal}, we have 
	\begin{equation}
		\sqrt{\frac{L}{\hat\vb_k^{\top}  \nabla^2 \cL(\hat\thetab; \bx)^{\dagger}\hat\vb_k}} (\hat\Delta_k - \Delta_k)\overset{d}{\longrightarrow} N(0,1),
	\end{equation}
	where  $\hat\vb_k = \Big({(0-r_k) \hat{u}_0 , (r_1 - r_k)\hat{u}_1, (r_2 - r_k)\hat{u}_2, \ldots, (r_{k-1}- r_k)\hat{u}_{k-1}} , 0, \ldots, 0\Big)^{\top}$ with $\hat{u}_i = \exp(\hat\theta_i')$.
\end{corollary}
See Supplementary Materials \ref{sec: proof col var est} for the proof of Corollary \ref{col: var est}. Based on the asymptotic normality of $\Delta_k$'s, we are ready to formally introduce the inferential procedure in the next section.
\subsection{Hypothesis Testing for \texorpdfstring{$\cS^*$}{cS star}}\label{sec: test proc}
In this section, we will perform a general inference on $\cS^*$ to test whether it satisfies certain properties of interest. As established in Section~\ref{sec: intro inf on s}, the general hypothesis testing for $\cS^*$ defined in \eqref{eq: hypo test S S0} is equivalent to the following test on $K^*$,
$$
\mathrm{H}_0: K^* \in \mathcal{K}_0 \text{~versus~}
\mathrm{H}_1: K^* \notin \mathcal{K}_0,
$$
where $\cK_0$ is the property set of $K^*$ satisfying certain properties. Then to perform the inference on~$K^*$, for a given level $\alpha \in (0,1)$, we utilize the equivalence between the events $\{k \in \cS^*\}$ and $\{\Delta_k < 0 \}$, and construct a confidence interval 
of confidence level $1-\alpha$ for $K^*$. Specifically, we define the maximal statistic
\begin{equation}\label{eq: T}
	T = \max_{k \in [n]} \big({\hat\Delta_k - \Delta_k}\big)\Big/{\sqrt{\hat\vb_k^{\top} \nabla^2 \cL(\hat\thetab; \bx)^{\dagger}\hat\vb_k/L}},
\end{equation}
and we construct the confidence interval of $K^*$ by estimating the quantile of $T$. We consider the Gaussian multiplier bootstrap proposed in \citet{cck2013aos} for estimating the quantile of $T$, which approximates the maximum of a sum of random vectors by the empirical quantiles of Gaussian maximum. In particular, we define the following statistic as the approximation of $T$,
{
\begin{equation}\label{eq: W}
	W =\max _{k \in [n]}  \left\{\frac{\hat\vb_k^{\top} \nabla^2 \cL(\hat\thetab; \bx)^{\dagger}}{\sqrt{L \cdot \hat\vb_k^{\top} \nabla^2 \cL(\hat\thetab; \bx)^{\dagger}\hat\vb_k}}\sum_{\cS \in \bcE} \sum_{\ell=1}^{L_{\cS}} \left\{ \sum_{i \in \cS_+} \left(x_{\cS}^{(i, \ell)} - \frac{e^{\hat\theta_i}}{\sum_{j \in \cS_+} e^{\hat\theta_j}}\right) \eb_i \right\} z_{\cS, \ell}\right\},
\end{equation}}
where  
$\{\{z_{\cS, \ell}\}_{\ell=1}^{L_{\cS}}\}_{\cS \in \bcE}$ are i.i.d. standard Gaussian.
For a given level $\alpha \in (0,1)$, we define the quantile of $W$ conditional on the data $\bx$ and the observed sets $\bcE$ as
\begin{equation}\label{eq: c alpha}
	c_{W}(\alpha, \bcE)=\inf \{t \in \mathbb{R}: \mathbb{P}(W>t \mid \boldsymbol{x}, \bcE) \leq \alpha\}.   
\end{equation}
We will show that $W$ approximates $T$, and $c_{W}(\alpha, \bcE)$ approximates $T$'s $\alpha$-quantile under some mild conditions.
{ We can then establish a \((1-\alpha)\)-level confidence interval for \(K^*\), denoted as \([\hat{K}_{L}, \hat{K}_{U}]\), as follows:
	{\small $$\hat{K}_{L} \!=\! \max \!\Bigg\{\!k: \!\hat\Delta_k \!<\! -c_{W}\big(\frac{\alpha}{2}, \bcE\big)\!\sqrt{\!\frac{\hat\vb_k^{\top} \nabla^2 \cL(\hat\thetab; \bx)^{\dagger} \hat\vb_k}{L} } \!\Bigg\}, \quad \hat{K}_{U} \!= \!\min\!\Bigg\{\!k\!:\! \hat\Delta_k \!>\! c_{W}\big(\frac{\alpha}{2}, \bcE\big)\!\sqrt{\!\frac{\hat\vb_k^{\top} \nabla^2 \cL(\hat\thetab; \bx)^{\dagger} \hat\vb_k }{L}} \!\Bigg\} -1,$$  
	} where, for \(\hat{K}_{U}\), if the minimum is taken over an empty set, we set \(\hat{K}_{U} = n\).}
We summarize the framework of constructing the confidence interval in Algorithm \ref{alg: compute ci}. 

Define the parameter space of $\thetab^*$ as $\bTheta = \{\theta: \kappa_{\theta} \le C_{\kappa}\}$, where $C_{\kappa}$ is a constant independent of~$n$. 
The following theorem shows the validity of the testing procedure based on the confidence interval for $K^*$. 
\begin{theorem}[Validity of Test]\label{thm: legit ci for K}
	Under the same conditions as Theorem \ref{thm: inf norm MLE}, suppose  that $ \kappa_r \lesssim \sqrt{n}$ and { $\kappa_{\bcE}^7 \cdot {n}^3 \log^{3} n / (\underline{\lambda} L) = o(1)$}. Then, 
	for any level $0 < \alpha <1$, we have that 
	\begin{equation}\label{eq: legit CI}
		\liminf_{n, L \rightarrow \infty} \inf_{\thetab^* \in \bTheta} \PP_{\thetab^*}(K^* \in [\hat{K}_{L}, \hat{K}_{U}]) \ge 1- \alpha . 
	\end{equation}
	Moreover, 
	\begin{equation}\label{eq: valid test}
		\limsup_{n, L \rightarrow \infty} \sup_{\thetab^*: \cS^* \in \bcS_0} \PP_{\thetab^*}(\text{\rm Reject } {\rm H}_0) \le \alpha.
	\end{equation}
\end{theorem}

{ Note that the theoretical validity does not necessitate any assumptions on the structure of $\mathcal{K}_0$, rendering our framework highly general. We give below a proof sketch of Theorem~\ref{thm: legit ci for K} to facilitate understanding. The detailed proof is provided in Section~\ref{sec: proof thm legit CI}.
	\proof{Proof Sketch.}
	The proof contains three steps. First,  we reduce quantifying the uncertainty of $\cS^*$ to quantifying the uncertainty of the sign change point detection based on the maximal scaled estimation error $T$. Second, we show that the multiplier bootstrap quantile $c_{W}(\alpha, \bcE)$ is valid and in turn the confidence interval is valid. Third, based on the equivalence between the two hypothesis tests \eqref{eq: hypo test S S0} and \eqref{eq: hypo test K S0}, we establish the asymptotic validity of our inferential procedure.
	
	
	\textbf{Step 1.} By previous arguments, we establish the equivalence of the following events:
	$$
	\{k \in \cS^*\} = \{k \le K^*\} = \{\Delta_k < 0\},
	$$
	which indicates that making inference on $\cS^*$ is equivalent to 
	inferring the signs of $\Delta_k$'s. Moreover, we have the following lower bound for the coverage probability
	\begin{align*}
		\PP(K^* \in [\hat{K}_L, \hat{K}_U]) \ge 1\! -\! \PP(\hat{K}_L \!>\! K^*) \!-\! \PP(\hat{K}_U\! +\! 1\! \le \!K^*) 
		\ge  1 \!-\! \PP(\Delta_{\hat{K}_L}\! \ge\!0)  - \PP\left(\Delta_{\hat{K}_U + 1} \!<\! 0\right) .
	\end{align*}
	By our definition of $\hat{K}_L$ and $\hat{K}_U$, the estimators of $\Delta_{\hat{K}_L}$ and $\Delta_{\hat{K}_U + 1}$ failing to predict the signs means that the estimation errors of both $\hat\Delta_{\hat{K}_L}$ and $\hat\Delta_{\hat{K}_U + 1}$ exceed the cut-off threshold, i.e.,
	$$
	\hat\Delta_{\hat{K}_L}\!\! -\! \Delta_{\hat{K}_L} \! \! < -c_{W}\big(\frac{\alpha}{2}, \bcE\big)\sqrt{\frac{\hat\vb_{\hat{K}_L}^{\top}\!\! \nabla^2 \cL(\hat\thetab; \bx)^{\dagger} \hat\vb_{\hat{K}_L}}{L} },\, \hat\Delta_{\hat{K}_U+1}\! -\! \Delta_{\hat{K}_U+1} \! \! > c_{W}\big(\frac{\alpha}{2}, \bcE\big)\sqrt{\frac{\hat\vb_{\hat{K}_U+1}^{\top} \!\nabla^2 \cL(\hat\thetab; \bx)^{\dagger} \hat\vb_{\hat{K}_U+1}}{L} }.
	$$
	Since the statistic $T$ controls the difference $(\hat\Delta_k - \Delta_k)$'s uniformly, we can further lower bound the coverage probability by 
	$$\PP(K^* \in [\hat{K}_L, \hat{K}_U]) \ge 1 - \PP\big(-T < -c_{W}({\alpha}/{2}, \bcE)\big) - \PP\big(T > c_{W}({\alpha}/{2}, \bcE)\big).$$
	Namely, so long as we can accurately estimate the quantile of $T$, we will have a valid confidence interval for $K^*$.
	
	\textbf{Step 2.} The following lemma shows that the quantile $c_{W}(\alpha, \bcE)$ obtained from the multiplier bootstrap in Algorithm~\ref{alg: compute ci} serves as a valid quantile of $T$ for any $\thetab^* \in \bTheta$.
	\begin{lemma}\label{lm: cw valid quantile cond on bcE}
		Under the same conditions as Theorem \ref{thm: legit ci for K}, we have that 
        \begin{equation}\label{eq: lem valid test}
            \sup_{\thetab^* \in \bTheta}\sup_{\alpha \in(0,1)}\left|\mathbb{P}_{\thetab^*}\left(T>c_{W}(\alpha, \bcE)  \right)-\alpha\right|=o(1) .
        \end{equation}
	\end{lemma}
	The proof of Lemma~\ref{lm: cw valid quantile cond on bcE} is deferred to Supplementary Materials~\ref{sec: proof thm legit ci k}.  Then from \textbf{Step 1}, we have that 
	$$
	\sup_{\thetab^* \in \bTheta}\PP_{\thetab^*}(\hat{K}_{L} \le K^* \le \hat{K}_U) \ge 1 - \sup_{\thetab^* \in \bTheta}\PP_{\thetab^*}(-T < -c_{W}(\alpha/2, \bcE) ) - \sup_{\thetab^* \in \bTheta}\PP_{\thetab^*}(T \ge c_{W}(\alpha/2, \bcE) ) \rightarrow 1- \alpha.
	$$
	Hence, \eqref{eq: legit CI} holds, and $[\hat{K}_L, \hat{K}_U]$ is a valid confidence interval.
	
	\textbf{Step 3.} Recall that we reject the null if and only if $[\hat{K}_L, \hat{K}_U]$ and $\cK_0$ have no intersection. Then by the equivalence between testing $\cS^* \in \bcS_0$ and testing $K^* \in \cK_0$, for large enough $n$ and $L$ we have that 
	$$
	\sup_{ \thetab^*: \cS^* \in \bcS_0}\PP_{\thetab^*}(\text{Reject } H_0)  = \sup_{\thetab^*: K^* \in \cK_0}\PP_{ \thetab^*}( [\hat{K}_L, \hat{K}_U] \cap \cK_0 = \emptyset)  \le 1 - \inf_{\thetab^* \in \bTheta}\PP_{\thetab^*}(K^* \in [\hat{K}_L, \hat{K}_U]) \le \alpha,
	$$
	suggesting that our test is asymptotically valid. \Halmos

	\endproof
}

Thus, for the hypothesis test on the smallest optimal set $\mathcal{S}^*$, we first construct the property set $\mathcal{K}_0$ of $K^*$ corresponding to the null hypothesis ${\rm H}_0$ as described in Section \ref{sec: intro inf on s}. Then, for a given level $\alpha \in (0,1)$, we construct the confidence interval { $ [\hat{K}_L, \hat{K}_U]$}
by Algorithm \ref{alg: compute ci} and reject ${\rm H}_0$ if and only if $ [\hat{K}_L, \hat{K}_U] \cap \mathcal{K}_0 = \emptyset$. Theorem~\ref{thm: legit ci for K} shows that the testing procedure is asymptotically valid. 

{ In the following part, we show the power of the proposed test. Recall that \(\bcS_0\) and \(\cK_0\) are the null sets of \(\cS^*\) and \(K^*\), corresponding to the null hypothesis \({\rm H}_0\). For any \(\thetab^* \in \bTheta\) such that \(\cS^* \notin \bcS_0\), i.e., the alternative hypothesis \({\rm H}_1\) holds at \(\thetab^*\), denote $\cK_0^{\rm L} = \max\{k \in \cK_0 \cup \{-1\}: k < K^* \}$ and $\cK_0^{\rm R} = \min\{k \in \cK_0 \cup \{n + 1\}: k > K^* \}$. 
	We then define the signal strength for $\thetab^*$ as 
	\begin{equation}\label{eq: signal strength}
		\delta(\thetab^*, \cK_0) = \min_{k \in \{ \cK_0^{\rm L} + 1, \cK_0^{\rm R} \} } \left[
		\sum_{j = 0}^{k-1}(r_j - r_k)^2\right]^{-1/2}\left|\Delta_k \right|,
	\end{equation}
	where $\Delta_0 = - \infty$ and $\Delta_{n+1} = + \infty$ for notational convenience.
	Additionally, let $\hat\cS$ be the maximizer in $\arg\max_{\cS \in \bcS}\sum_{i \in \mathcal{S}} { r_i \exp(\hat{\theta}_i^d)}/{\big[1 + \sum_{j \in \mathcal{S}} \exp(\hat{\theta}_j^d)\big]}$ with the smallest cardinality.  The following theorem shows that our procedure is asymptotically powerful under an appropriate signal strength condition. We also compare this signal strength condition to that required for $\hat \cS$ being a consistent estimator of  \(\cS^*\).
	\begin{theorem}[Power of Test]\label{thm: power and compare with recovery}
		Define the rate { $\delta_n = \sqrt{ {\log n}/{(\underline{\lambda} L )}}$}. Under the same conditions as Theorem~\ref{thm: legit ci for K}, we have the following results.
		\begin{enumerate}
			\item  There exists a large enough constant $C > 0$ such that
			$$
			\lim_{n, L \rightarrow \infty}  \inf_{\thetab^*: \cS^* \notin \bcS_0, \delta(\thetab^*, \cK_0) \ge C \delta_n} \PP_{\thetab^*} (\text{\rm Reject } {\rm H}_0 ) = 1.
			$$
			\item Under the condition that
			\begin{equation}\label{eq: signal strength perfect recovery}
				\delta(\thetab^*, \{K^*\}^{\rm c}) =\min_{k \in \{K^*, K^*+1\}} \left[\sum_{j=0}^{k-1} (r_j - r_k)^2\right]^{-1/2} |\Delta_k| \ge C \delta_n,
			\end{equation}
			for some large enough $C > 0$, we have $\lim_{n, L \rightarrow \infty} \PP(\hat\cS = \cS^*) = 1$. On the other hand, if there exists a small enough constant $c > 0$ such that
			$$
			\delta(\thetab^*, \{K^*\}^{\rm c}) \le c ({ \kappa_{\bcE}}\log n/[1-(\underline{K} + 2)/(n+1)])^{-1/2}\delta_n,
			$$ 
			where $\underline{K} = |\{j \in [n]: r_j > r_{K^*}\}|$, we have $\liminf_{n, L \rightarrow \infty} \PP(\hat\cS \ne \cS^*) \ge 1/3$.
		\end{enumerate}
		
	\end{theorem}
	We provide the proof of Theorem~\ref{thm: power and compare with recovery} in Supplementary Materials~\ref{sec: proof of thm power and compare with recovery}. 
	
	Note that  the signal strength for the perfect recovery $\delta(\thetab^*, \{K^*\}^{\rm c})$ in \eqref{eq: signal strength perfect recovery} is a special case of the signal strength for the powerful property test $\delta(\thetab^*, \cK_0)$ in \eqref{eq: signal strength}. The difference between these two quantities shows that the hardness of property test and perfect recovery might be different.
	{ The following corollary of Theorem~\ref{thm: power and compare with recovery} shows that we could still have a powerful test of some properties of $\cS^*$ even if $\hat \cS$ could not perfectly recover $\cS^*$.
	\begin{corollary}\label{col: compare signal}
		Under the same conditions as Theorem~\ref{thm: legit ci for K}, for \(\thetab^* \in \bTheta\) such that \(\cS^* \notin \bcS_0\), if \(\max\cK_0 < K^* \leq c n\), for some constant \(c \in (0,1)\), and
		\begin{equation}\label{eq:cor2}
			\sum_{j=0}^{K^*-1} (r_j - r_{K^*})^2 \gg {{ \kappa_{\bcE}}\log n} \cdot  \sum_{j=0}^{\cK_0^{\rm L}} (r_j - r_{\cK_0^{\rm L} + 1})^2,
		\end{equation}
		then $\delta(\thetab^*, \{K^*\}^{\rm c})\sqrt{{ \kappa_{\bcE}}\log n} \ll \delta(\thetab^*, \cK_0)$. Furthermore, for some large enough constants $C_1, C_2 > 0$ that
				\begin{equation}\label{eq:cor2-interval}
					{ \underline{\lambda} L} \in [ C_1 \delta^{-2} 
					(\thetab^*, \cK_0 ) \log n  , \  C_2 { \kappa_{\bcE}^{-1}}\delta^{-2} 
					(\thetab^*, \{K^*\}^{\rm c}) ],
				\end{equation}
				 we have \(\liminf_{n, L \rightarrow \infty} \PP(\hat{\cS} \neq \cS^*) \ge 1/3\), and \(\lim_{n, L \rightarrow \infty} \PP_{\thetab^*} (\text{\rm Reject } \mathrm{H}_0) = 1\).
	\end{corollary}
	\begin{remark}
		The condition in \eqref{eq:cor2} suggests that if \(K^* \gg { \kappa_{\bcE} \log n} \cdot K_0^{\rm L} = { \kappa_{\bcE} \log n} \cdot \max \cK_0\), i.e., the cardinality of \(\cS^*\) is much larger than that of assortments in the null set, perfect recovery of the entire \(\cS^*\) becomes significantly more challenging than testing the properties of \(\cS^*\). The intuition is that perfect recovery requires controlling the uncertainty for recovering \(K^*\) items, while property testing only needs to manage \(\max \cK_0\) items. The condition in \eqref{eq:cor2-interval} further characterizes the specific regime of the total sample size as reflected in { $\underline{\lambda} L$} in which a powerful property test is achievable, but perfect recovery is not. This regime becomes infinitely wide as \(\delta(\thetab^*, \{K^*\}^{\rm c})\sqrt{{ \kappa_{\bcE} }\cdot\log n} \ll \delta(\thetab^*, \cK_0)\).
		
	\end{remark}}
}

\begin{algorithm}[ht]
	\caption{Construction of confidence interval for $K^*$}\label{alg: compute ci}
	\begin{algorithmic}
		\STATE \textbf{Input:} { Observed} sets $\bcE \subseteq \bcS$, customer choice data $\{x_{\cS}^{(i,\ell)}\}_{S \in \bcE, \ell \in [L_{\cS}]}$, regularization tuning parameter $\tau$, revenue parameters $\{r_i\}_{i \in [n]}$, given level $\alpha$.
		\STATE \textbf{Output: } Regularized MLE $\hat\thetab$,  $(1-\alpha)$-level $ [\hat{K}_L,\hat{K}_U]$.
	\end{algorithmic}
	\begin{algorithmic}[1]
		\STATE Compute the regularized MLE $\hat\thetab \gets \operatorname{argmin}_{\thetab \in \RR^{n+1}, \theta_0 = 0} \cL(\thetab; \bx) + \frac{\tau}{2}\sum_{i \in [n]_+} (\theta_i - \bar{\theta})^2$;
		\STATE Obtain the debiased MLE $\hat\thetab^d$ by  \eqref{eq: debias mle};
		\STATE Generate $z_{\cS,\ell} \overset{\text{i.i.d.}}{\sim} N(0,1)$ for $\cS \in \bcE$, $\ell \in [L_{\cS}]$, 
		\FOR{$k \in [n]$}
		\STATE { $w_k \gets \left\{\frac{\hat\vb_k^{\top} \nabla^2 \cL(\hat\thetab; \bx)^{\dagger}}{\sqrt{L \hat\vb_k^{\top} \nabla^2 \cL(\hat\thetab; \bx)^{\dagger}\hat\vb_k}}\sum_{\cS \in \bcE} \sum_{\ell=1}^{L_{\cS}} \left\{ \sum_{i \in \cS_+} \left(x_{\cS}^{(i, \ell)} - \frac{e^{\hat\theta_i}}{\sum_{j \in \cS_+} e^{\hat\theta_j}}\right) \eb_i \right\} z_{\cS, \ell}\right\}$;}
		\ENDFOR
		\STATE $W \gets \max_{k \in [n]} w_k$, compute the $1-\alpha/2$ quantile of $W$: $c_W(\alpha/2, \bcE)$;
		\STATE Compute the confidence interval of level $1-\alpha$, $[\hat{K}_L, \hat{K}_U]$: 
		$$\hat{K}_{L} = \max\left\{k: \hat\Delta_k < -c_{W}(\alpha/2, \bcE)\sqrt{{\hat\vb_k^{\top} \nabla^2 \cL(\hat\thetab; \bx)^{\dagger} \hat\vb_k}/{L} } \right\}, $$
		{ $$\hat{K}_{U} = \min\left\{k: \hat\Delta_k > c_{W}(\alpha/2, \bcE)\sqrt{{\hat\vb_k^{\top} \nabla^2 \cL(\hat\thetab; \bx)^{\dagger} \hat\vb_k }/{L}} \right\} -1 .$$}
	\end{algorithmic}
\end{algorithm}
{
	
	Note that given the specific hypothesis, the testing procedure can be simplified by taking advantage of the structure of the property set $\mathcal{K}_0$. For instance, Example~\ref{exm 1} essentially boils down to testing ${\rm H}_0: \Delta_{i} \ge 0$. As shown by Corollary \ref{col: var est} in Section \ref{sec: lagrangian debias}, under certain  conditions, for any $i \in [n]$,  $\big(\hat\vb_i^{\top}  \nabla^2 \cL(\hat\thetab; \bx)^{\dagger} \hat\vb_i/L\big)^{-1/2}(\hat\Delta_i - \Delta_i)$ converges  in distribution to the standard Gaussian. Thus for a given level $\alpha \in (0,1)$, we can construct the confidence interval for $\Delta_i$ as 
	$$
	\left[ \,\, \hat\Delta_i - \Phi^{-1}(1-\frac{\alpha}{2})\sqrt{\frac{\hat\vb_i^{\top}  \nabla^2 \cL(\hat\thetab; \bx)^{\dagger}\hat\vb_i}{L}},\, \hat\Delta_i + \Phi^{-1}(1-\frac{\alpha}{2})\sqrt{\frac{\hat\vb_i^{\top}  \nabla^2 \cL(\hat\thetab; \bx)^{\dagger}\hat\vb_i}{L}} \,\, \right], \quad i \in [n],
	$$
	where $\Phi(\cdot)$ denotes the cumulative distribution function of the standard normal distribution.
	Then for Example~\ref{exm 1}, we will reject the null hypothesis if and only if $$\hat\Delta_{i} + \Phi^{-1}(1-\frac{\alpha}{2})\sqrt{{\hat\vb_{i}^{\top}  \nabla^2 \cL(\hat\thetab; \bx)^{\dagger}\hat\vb_{i}}/{L}} < 0, $$
	which is more efficient than Algorithm~\ref{alg: compute ci}.
}
\if1\theorysection{
	\section{Theoretical Analysis}\label{sec: theory anal}
	We denote by $\delta_r = r_1 - r_2$ the difference between the highest and the second highest revenue. Then we introduce the condition number for the revenue as
	$$
	\kappa_r := \max \left( \frac{r_1}{\delta_r}, \frac{r_1}{r_n}\right).
	$$
	Intuitively, the condition number for the revenue characterizes how much the largest revenue stands out from all the products. A larger condition number indicates a less uniform distribution of the revenues.
	\begin{theorem}\label{thm: inf norm MLE}
		Recall that $\hat\thetab$ is the regularized MLE. Under the conditions that $\lambda \asymp  \sqrt{\frac{2^n p \log n}{nL}}$, $2^n p \ge C n\log n$ for some large enough constant $C >0$ and $n \sqrt{\log n / (2^n p L)} \le c$ for some small enough constant $c > 0$
		, we have that the event  
		\begin{equation}\label{eq: inf norm MLE}
			\cA_{\hat\thetab} := \left\{\|\hat\thetab - \thetab^*\|_{2} \lesssim  n\sqrt{\frac{\log n}{2^n p L}} \right\}
		\end{equation}
		occurs with probability at least $1-O(n^{-10})$.
	\end{theorem}
	Please refer to Supplementary Materials \ref{sec: proof thm inf norm mle} for the proof of Theorem \ref{thm: inf norm MLE}.
	\begin{theorem}\label{thm: asymp normal}
		Under the same conditions as Theorem \ref{thm: inf norm MLE} along with the conditions that $\kappa_r \lesssim \sqrt{n}$ and ${n^2 \log n}/{\sqrt{2^n p L}}= o(1)$, for all $k \in [n]$ we have that 
		\begin{equation}
			\sqrt{\frac{L}{\vb_k^{\top} \nabla^2 \cL(\thetab^*; \bx)^{\dagger}  \vb_k}} (\hat\Delta_k - \Delta_k)\overset{d}{\rightarrow} N(0,1),
		\end{equation}
		where $\vb_k = \Big(\underbrace{(0-r_k) u_0^*, (r_1 - r_k)u_1^*, (r_2 - r_k)u_2^*, \ldots, (r_{k-1}- r_k)u_{k-1}^*}_{\text{first $k$ entries}} , 0, \ldots, 0\Big)^{\top}$.
	\end{theorem}
	The proof of Theorem \ref{thm: asymp normal} is deferred to Supplementary Materials \ref{sec: proof thm asymp normal}.
	\begin{corollary}\label{col: var est}
		Under the same conditions as Theorem \ref{thm: asymp normal}, we have 
		\begin{equation}
			\sqrt{\frac{L}{\hat\vb_k^{\top}  \nabla^2 \cL(\hat\thetab; \bx)^{\dagger}^* \hat\vb_k}} (\hat\Delta_k - \Delta_k)\overset{d}{\rightarrow} N(0,1),
		\end{equation}
		where $\hat\vb_k = \vb_k(\hat\thetab)$.
	\end{corollary}
	See Supplementary Materials \ref{sec: proof col var est} for the proof of Corollary \ref{col: var est}.
	\begin{theorem}\label{thm: legit ci for K}
		Under the same conditions as Theorem \ref{thm: inf norm MLE} along with the conditions that $ \kappa_r \lesssim \sqrt{n}$, $n\left(\log(nL)\right)^7 / L = o(1)$ and $\sqrt{\log n}\max(\frac{n^2 \log n}{\sqrt{2^n p} L}, \frac{n^4 \log n}{2^n p L})  = o(1)$ we have that 
		\begin{equation}
			\liminf_{n \rightarrow \infty} \PP(K^* \in [\hat{K}_{L}, \hat{K}_{U}]) \ge 1- \alpha .
		\end{equation}
	\end{theorem}
	
	The proof of Theorem \ref{thm: legit ci for K} is deferred to Supplementary Materials \ref{sec: proof thm legit ci k}. 
}\fi

\section{Numerical Results}\label{sec: numerical results}
In this section, we conduct simulation studies to evaluate the empirical performance of the proposed method (Alg.~\ref{alg: compute ci}). { We primarily consider the setting of Lemma~\ref{lm: eigen of L cE}, where historical assortments are generated independently, and in Section~\ref{sec: simu rev ordered} we further evaluate the method when the historical assortments are revenue-ordered, aligning with the setup of Lemma~\ref{prop: rev order assort}. {Additional numerical results evaluating $\kappa_{\bcE}$ under different data-generating regimes, the robustness of the proposed method, and the empirical values of $\underline{\lambda}$ and $\kappa_{\bcE}$ on publicly available data sets are deferred to Supplementary Materials~\ref{sec: add simu}.}
}

{ For the independent generation of the observed offer sets $\bcE$, we set the observation probability to $p_{\cS} = p = n\log n/2^n$ and the sample size to $L_{\cS} = L_0$ for all $\cS \in \bcS$, and we generate $\bcE$ through the following procedure:
	\begin{enumerate}
		\item Generate the total {number} of observed offer sets from the binomial distribution: $M \sim {\rm Binomial} (2^n - 1, p)$;
		\item For $j = 1, \ldots, M$, generate the random vector $\bv_j \in \{0,1\}^n$, where $\bv_{ji} \sim{\rm Bernoulli (1/2)}$ i.i.d. for $i = 1, \ldots, n$. Then we define the $j$-th observed offer set as $\cS_j = \{i \in [n]: \bv_{ji} = 1\}$. If $\cS_j = \cS_{j'}$ for some $j' < j$ or $\cS_j = \emptyset$, repeat the generation of $\bv_j$ until we successfully generate a non-empty $\cS_j$ without replacement.
		\item Output the observed offer sets as $\bcE = \{\cS_j\}_{j=1}^M$.
	\end{enumerate}
	It can be checked that following the above procedure, each offer set $\cS \in \bcS$ will be included in $\bcE$ uniformly with probability $p$, and the total generation complexity is $O(n^2 \log n)$ (we refer interested readers to Proposition~\ref{prop: sample complexity cE} in Supplementary Materials~\ref{sec: sample complexity cE} for details).
}

With $\bcE$ in place, we begin by evaluating the validity of the constructed confidence intervals through the empirical coverage probabilities, i.e., the empirical evaluation of $\PP(\hat{K}_L \le K^* \le \hat{K}_U)$. We then apply Algorithm~\ref{alg: compute ci} to Example~\ref{exm 2} and Example~\ref{exm 5} discussed in Section~\ref{sec: intro inf on s} as representative illustrations of the method's performance. For each example, we provide the empirical evaluation of the Type I error, i.e., the probability of rejecting the null when it is true, and the power, i.e., the probability of rejecting the null when it is false. 
Recall from Theorem~\ref{thm: alg} that the optimal assortment is of the form $\mathcal{S}^* = [K^*]$, and we aim to conduct general inference on the properties of $\mathcal{S}^*$, which is equivalent to testing the general null hypothesis ${\rm H}_0 : K^* \in \mathcal{K}_0$  in \eqref{eq: hypo test K S0}. Then the Type I error and power are defined by $\PP_{K^* \in \mathcal{K}_0}([\hat{K}_L,\hat{K}_U] \cap \mathcal{K}_0 = \emptyset)$ and $\PP_{K^* \notin \mathcal{K}_0}([\hat{K}_L,\hat{K}_U] \cap \mathcal{K}_0 = \emptyset)$, respectively.

Note that for the multiplier bootstrap in Algorithm~\ref{alg: compute ci}, we generate 200 independent Gaussian samples to evaluate the quantile of $W$ in \eqref{eq: W}. We generate the $\log$ preference score parameter~$\thetab^*\in\RR^n$ by $\theta_i^* \overset{\text{i.i.d.}}{\sim} N(0, \sigma_{\theta}^2)$ for $i \in [n]$ and $\theta_0^* = 0$. For each given $K^*$, we set $\Delta_k = -0.001$ if $k\le K^*$, $\Delta_k = 0$ if $k = K^* + 1$ and $\Delta_k = 0.001$ if $k > K^* +1$, then we solve for the revenue parameters from $\Delta_k$'s through the relationship $\Delta_k = \exp(-\bar\theta^*)\left(\sum_{i = 1}^k r_i u_i^* - (\sum_{i=0}^k u_i^*)r_k \right)$, $k \in [n]$. Note that since 
$r_k - r_{k+1} = \exp(\bar\theta^*) (\sum_{i=0}^k u_i^*)^{-1}(\Delta_{k+1} - \Delta_k)$, the corresponding revenue parameters satisfy that $r_1 \ge r_2 \ge \ldots \ge r_n$.
We set the penalty parameter $\tau = c\sqrt{2^np\log n/(nL_0)}$ according to Theorem \ref{thm: inf norm MLE}, where $c$ is a tuning constant.

For all settings, we fix $n = 30$ and $\sigma^2_{\theta} = 3$. 
We set the customer sample size { $L_0$} at different values to evaluate how the performance of Algorithm \ref{alg: compute ci}  changes over different sample sizes. We summarize the results of 300 Monte Carlo simulations in the following sections. {The average value of $\kappa_{\bcE}$ over all Monte Carlo replications is $2.63$.}

\subsection{Asymptotic Normality of Debiased Estimators}\label{sec: simu ind dat ci}
\begin{figure}[htbp]
	\centering
	\begin{tabular}{cc}
		\quad (a) Q-Q plot for $\hat\theta_1^d$ & \quad (b) Q-Q plot for $\hat\Delta_1$ \\
		\includegraphics[height = 0.37\textwidth]{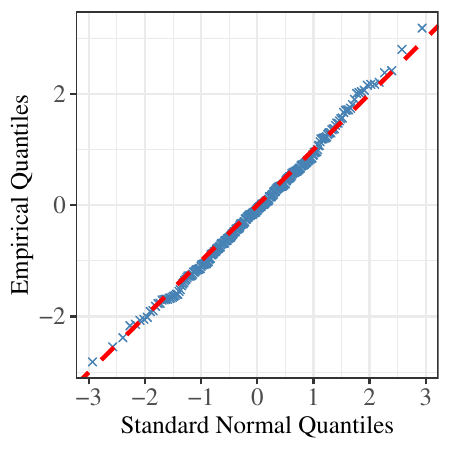}& \includegraphics[height = 0.37\textwidth]{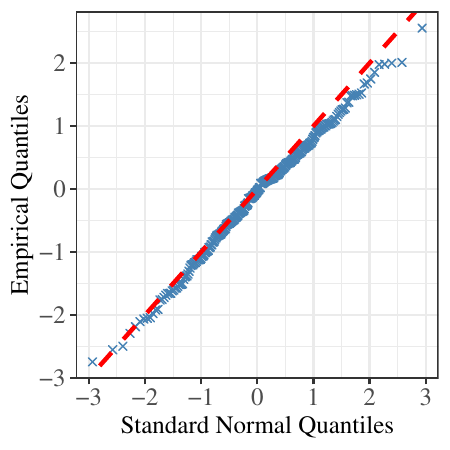}
	\end{tabular}
	\caption{Q-Q plots for $\hat\theta_1^d$ and $\hat\Delta_1$ versus the standard normal distribution.}
	\label{fig:qq plot}
\end{figure}

Before evaluating the performance of Algorithm~\ref{alg: compute ci}, we demonstrate in this section the asymptotic normality of the debiased MLE $\hat{\btheta}^d$ and the plug-in estimator $\hat\Delta_k$. We select the tuning constant $c = 0.25$ for the penalty parameter $\tau$ via cross-validation, and set 
$K^* = 7$. We set the customer sample size at $L_0 = 1000$. Figure~\ref{fig:qq plot} provides the Q-Q plots of $\hat\theta_1^d$ and $\hat\Delta_1$. We center $\hat\theta_1^d$ by $\theta^*_1 - \bar\theta^*$ and scale $\hat\theta_1^d$ by $\sqrt{\eb_2^{\top} \nabla^2 \cL(\hat\thetab; \bx)^{\dagger}\eb_2/L_0}$, and we center $\hat\Delta_1$ by $\Delta_1$ and scale $\hat\Delta_1$ by $\sqrt{\hat\vb_1^{\top} \nabla^2 \cL(\hat\thetab; \bx)^{\dagger}\hat\vb_1/L_0}$ according to Corollary~\ref{col: var est}. We  see that both $\hat\theta_1^d$ and $\hat\Delta_1$ are  approximately normally distributed. 
\subsection{Performance of Algorithm~\ref{alg: compute ci} on Hypothesis Testing}\label{sec: ind dat inf}
\begin{figure}
	\centering
	\includegraphics[width = 0.75\textwidth]{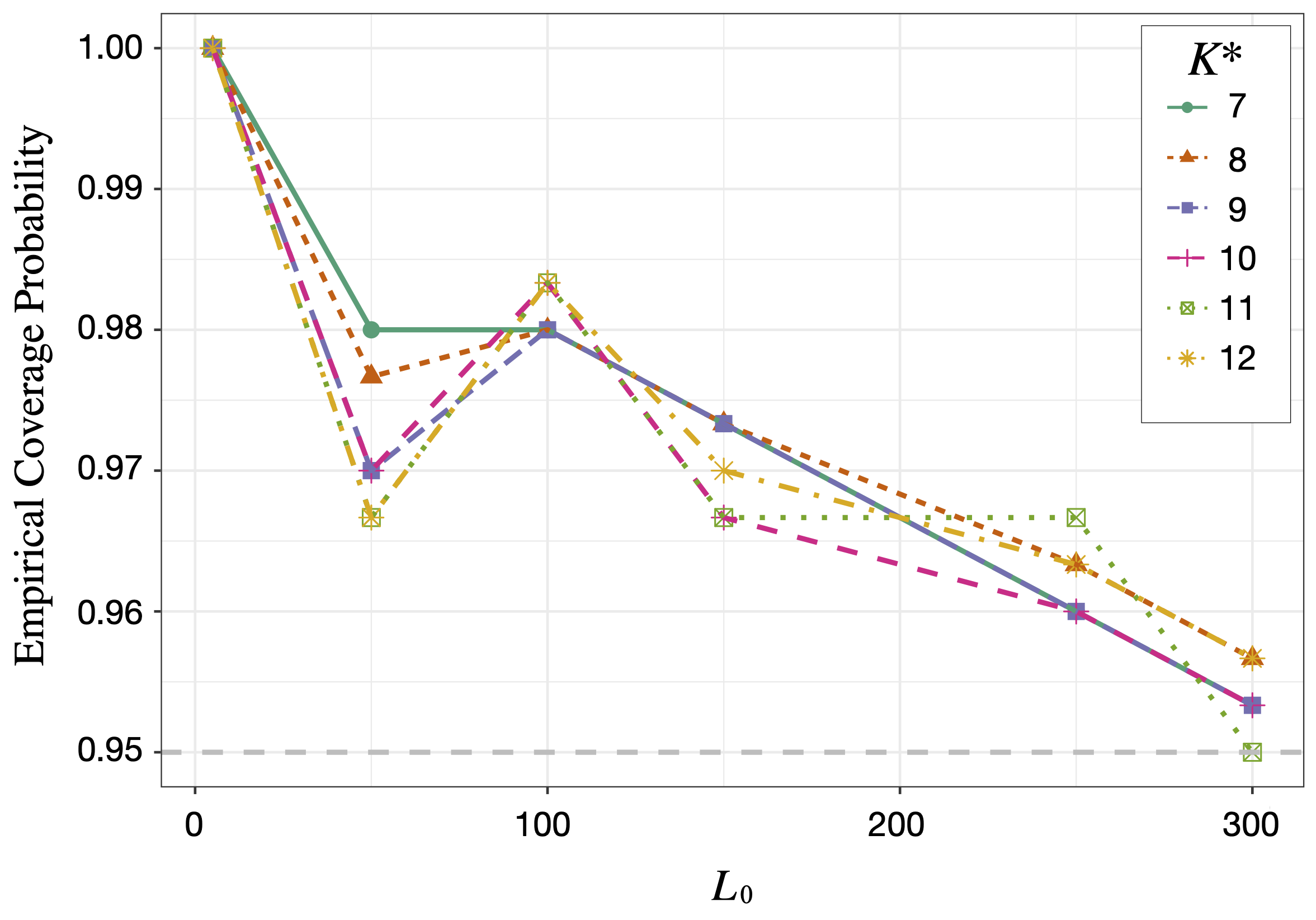}
	\caption{The empirical coverage probability of $[\hat{K}_L, \hat{K}_U]$ constructed by Algorithm~\ref{alg: compute ci} at $L_0 \in \{5,50,100,150,250,300\}$ when $K^* \in \{7,8,9,10,11,12\}$. The grey dashed line represents the nominal coverage probability equal to 0.95. }
	\label{fig: emp cov prob}
\end{figure}

We apply Algorithm~\ref{alg: compute ci} to concrete examples provided in Section~\ref{sec: intro inf on s}. We select the tuning constant of $\tau$ through cross-validation by letting $c = 1$ for constructing the confidence intervals. We first evaluate the empirical coverage probability of the confidence interval $ [\hat{K}_L, \hat{K}_U]$. We set the nominal level at $0.95$, and set the sample size at $L_0 \in \{5,50,100,150,250,300\}$ respectively. 
Figure~\ref{fig: emp cov prob} shows the empirical coverage probability at $K^* \in \{7,8,9,10,11,12\}$ for each setting of $L_0$. We can see that the empirical coverage probability of $[\hat{K}_L, \hat{K}_U]$ converges to the nominal level of 0.95 as $L_0$ increases.

Then we provide two concrete examples. To characterize the distance of the alternative from the null, for a given set $S \subseteq [n]$, we define $d(K^*, S) = \min_{i \in S} |i - K^*|$. Then for $K^* \notin \mathcal{K}_0$, $d(K^*, \mathcal{K}_0)$ characterizes the difficulty of differentiating ${\rm H}_1$ from ${\rm H}_0$.  We summarize the results of Example~\ref{exm 2} and Example~\ref{exm 5} below.
\subsubsection*{Example~\ref{exm 2}}
\begin{figure}[htbp]
	\centering
	\begin{tabular}{cc}
		\quad (a) Type I error & \quad (b) Power \\
		\includegraphics[height = 0.33\textwidth]{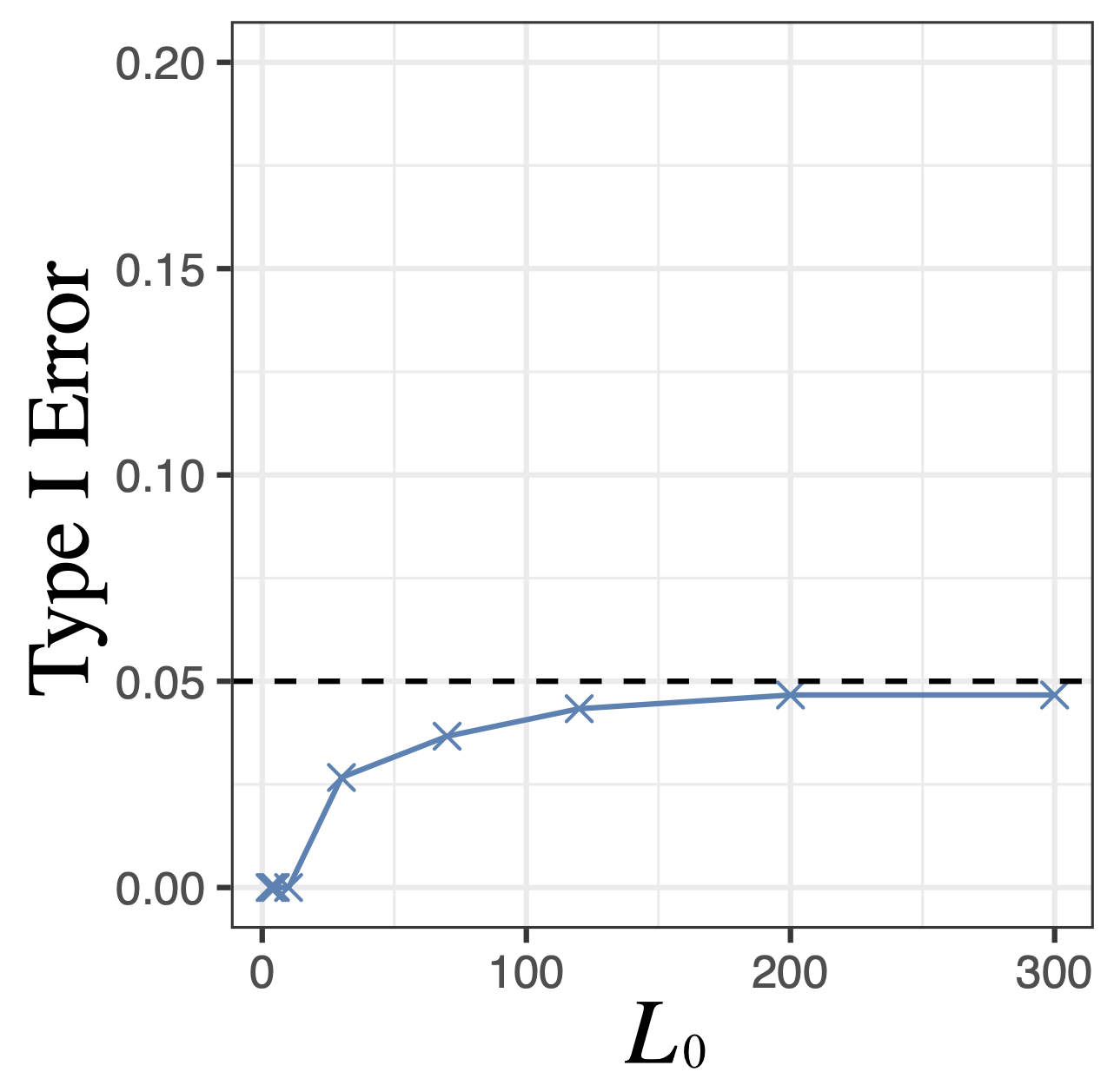}& \includegraphics[height = 0.33\textwidth]{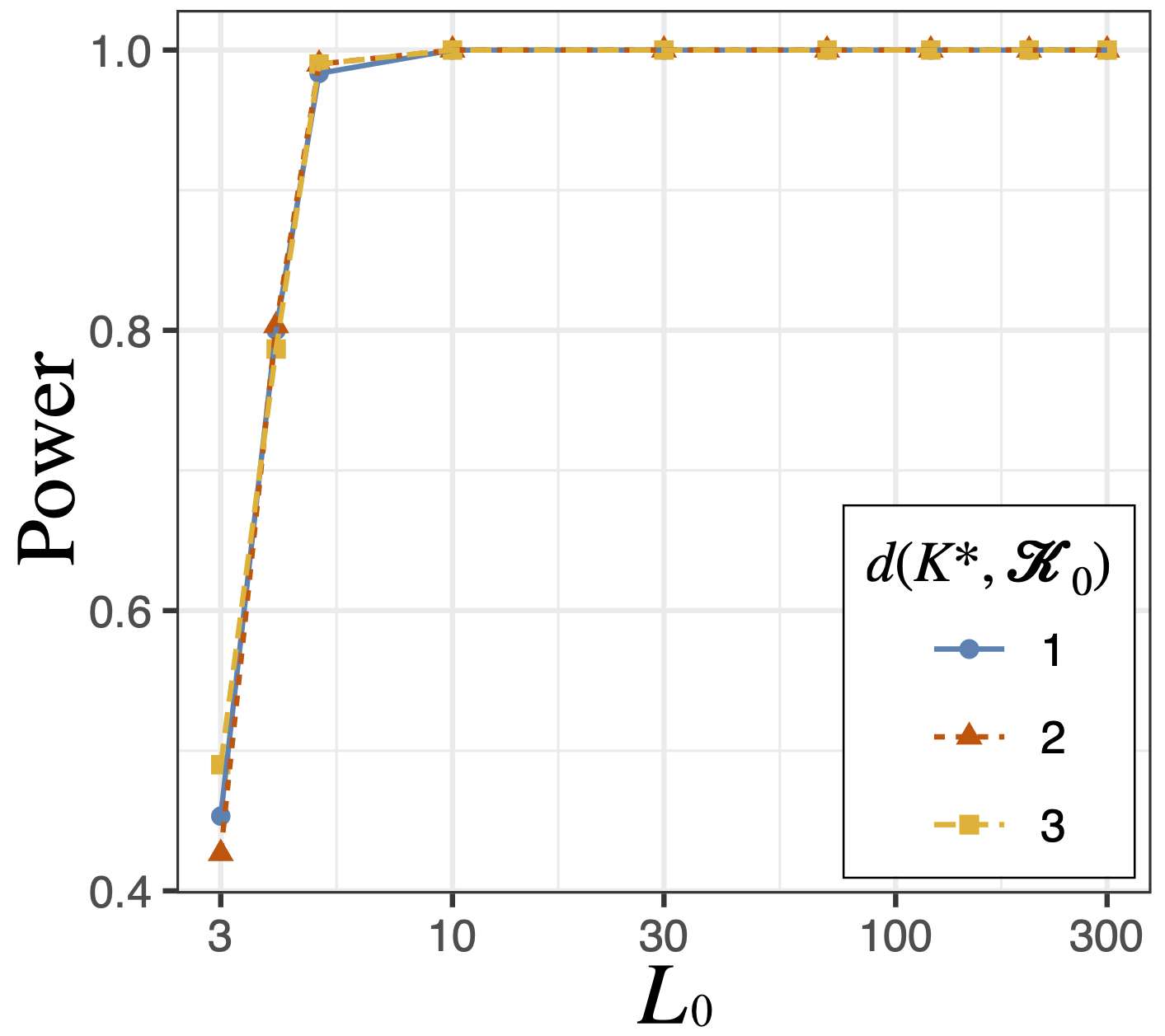}
	\end{tabular}
	\caption{Type I error and power for Example~\ref{exm 2} under different $L_0$ and $d(K^*, \mathcal{K}_0)$, where the dashed line in (a) represents the nominal 0.05 level.}
	\label{fig: exm2 simu}
\end{figure}

Recall that we are interested in testing whether a given set $A \subseteq [n]$ is a subset of the optimal set $\mathcal{S}^*$. In particular,
$$
\mathrm{H}_0: A \nsubseteq \mathcal{S}^* \text{~versus~} \mathrm{H}_1:  A \subseteq  \mathcal{S}^*.
$$
We set $A = \{2,4,6,8\}$ and 
generate $K^* \in \{7,8,9,10\}$. It can be seen that {$\mathcal{K}_0 = [7]$}. We set the null at $K^* = 7$ and choose different alternatives to evaluate the empirical Type I error and power. The results are summarized in Figure~\ref{fig: exm2 simu}. We see that the Type I error converges to the nominal 0.05 level as $L_0$ increases, while the power increases to 1 as $L_0$ grows for all settings of $d(K^*,\mathcal{K}_0)$.

\subsubsection*{Example~\ref{exm 5}}
\begin{figure}[htbp]
	\centering
	\begin{tabular}{cc}
		\quad (a) Type I error & \quad (b) Power \\
		\includegraphics[height = 0.33\textwidth]{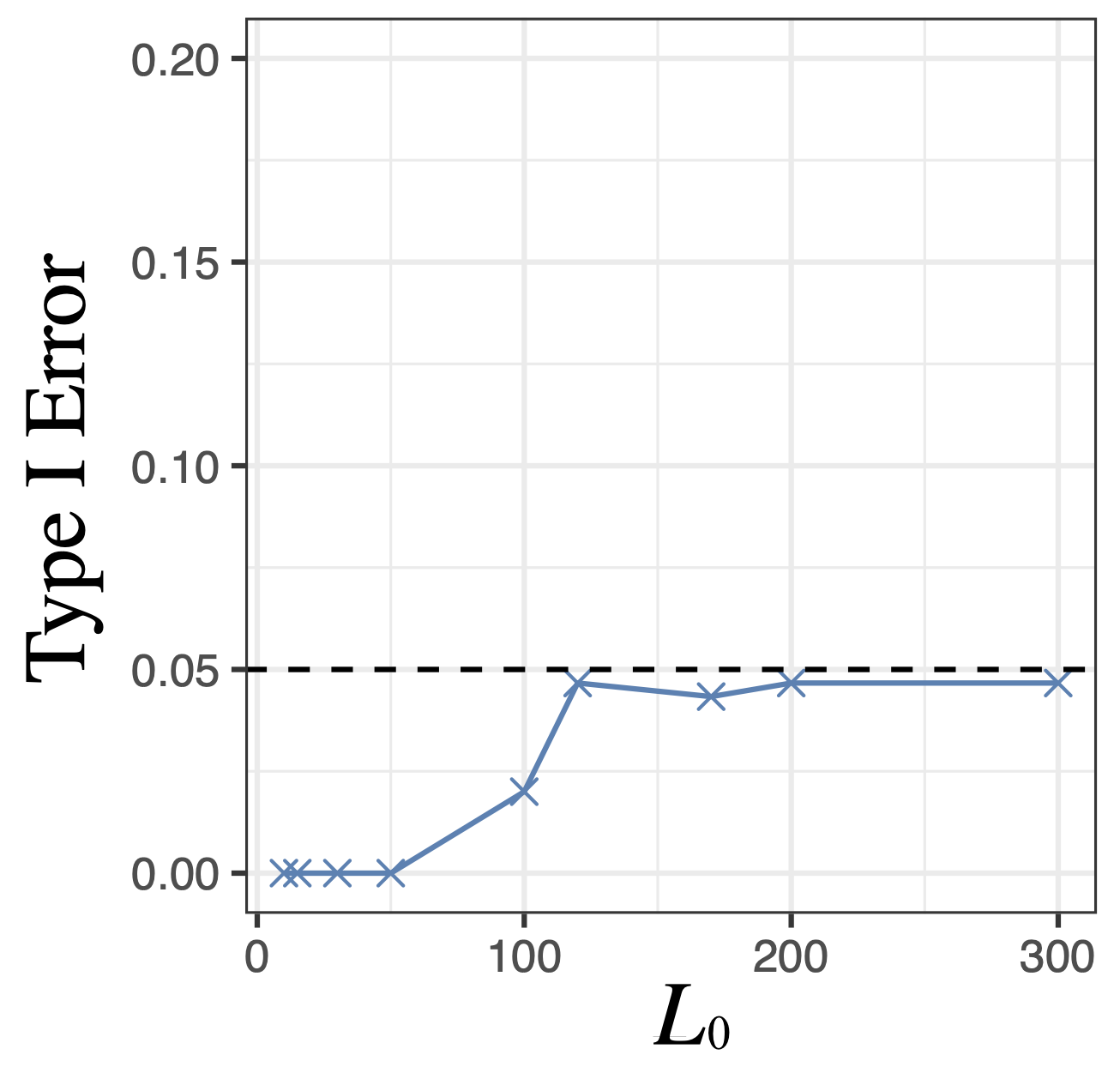}& \includegraphics[height = 0.33\textwidth]{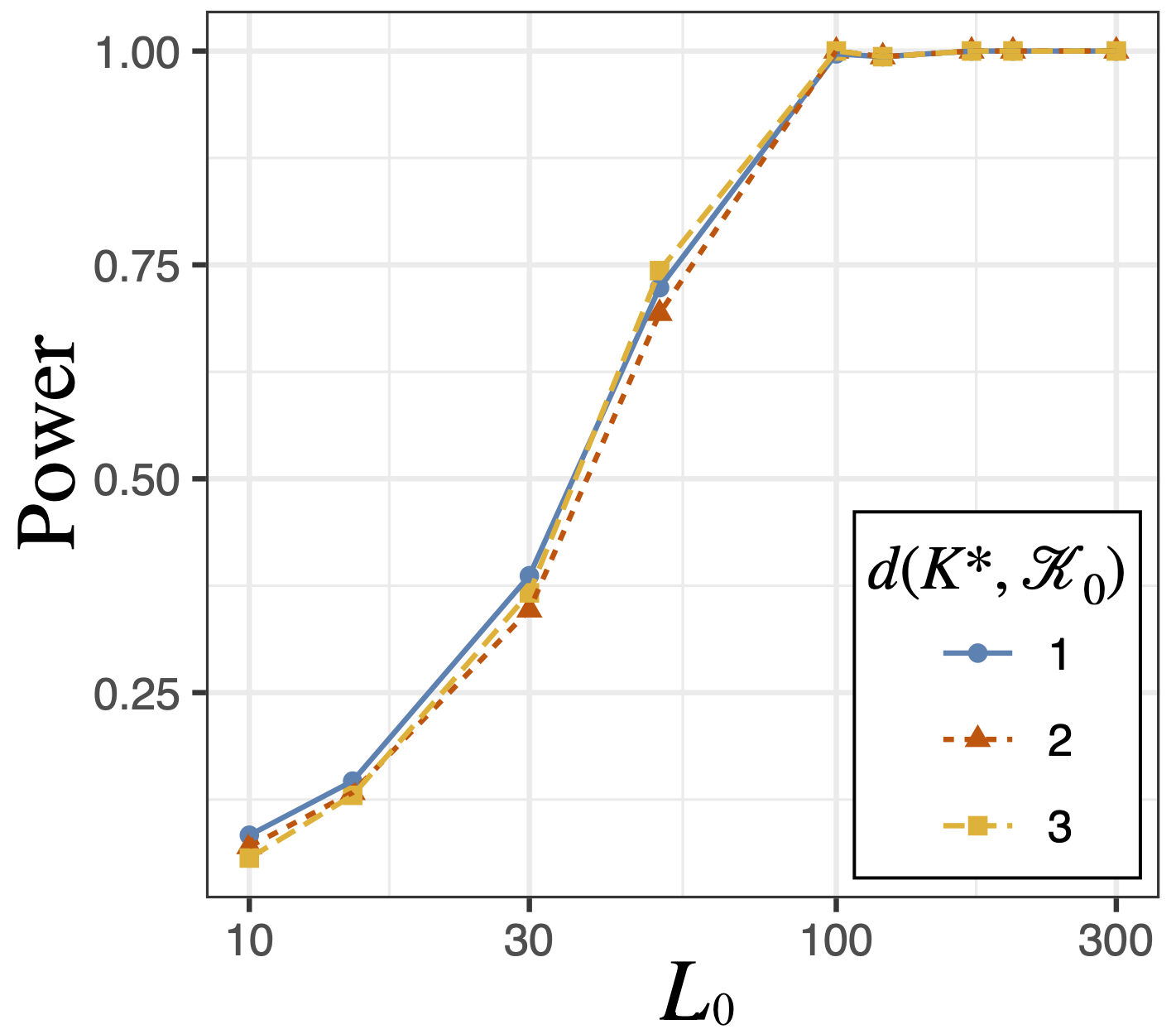}
	\end{tabular}
	\caption{Type I error and power for Example~\ref{exm 5} under different $L_0$ and $d(K^*, \mathcal{K}_0)$, where the dashed line in (a) represents the nominal 0.05 level. }
	\label{fig: exm5 simu}
\end{figure}

Recall that for a partition $\{A_1, \ldots, A_m\}$ of the products, i.e., $\bigcup_{j=1}^m A_j = [n]$ and $A_j \cap A_k = \emptyset$ for $j \neq k$, we are interested in testing whether $A_1$ contains the most elements in $\mathcal{S}^*$ compared with the other $A_j$'s. In particular, we are interested in testing the hypothesis
$$
\mathrm{H}_0: \lvert \mathcal{S}^* \cap A_1 \rvert = \max_j \lvert \mathcal{S}^* \cap A_j \rvert \text{~versus~}
\mathrm{H}_1: \lvert \mathcal{S}^* \cap A_1 \rvert < \max_j \lvert \mathcal{S}^* \cap A_j \rvert.
$$
For the simulation, we partition $[n] = [30]$ by $A_1 = [6:8] \cup [14:15] \cup [21:25]$, $A_2 = [1:3] \cup \{10,12\} \cup [16:20]$ and $A_3 = [4:5] \cup \{9,11,13\} \cup [26:30]$, and the corresponding {$\mathcal{K}_0$ is $[8:9]\cup \{15\} \cup [25:30] $}. We generate $K^* \in \{9,10,11,12\}$ and set the null at $K^* = 9$. Results are summarized in Figure~\ref{fig: exm5 simu}. The Type I error converges to the nominal level of 0.05 as $L_0$ increases, while the power increases to 1 as $L_0$ increases at $d(K^*, \mathcal{K}_0) \in \{1,2,3\}$. 

In summary, Algorithm~\ref{alg: compute ci} performs well for both examples, demonstrating the validity of our method.

\subsection{Empirical Performance under Revenue-Ordered Historical Assortments}\label{sec: simu rev ordered}
We evaluate the performance of Algorithm~\ref{alg: compute ci} when the historical assortments are revenue-ordered, and compare its Type~I error with that of directly applying the point estimator in \eqref{eq: debias mle} without uncertainty quantification. We generate \(\thetab^*\) using the same procedure as in Sections~\ref{sec: simu ind dat ci} and \ref{sec: ind dat inf}. Item revenues are drawn from a shifted exponential distribution with rate \(7.5\) and support starting at \(3\), and we set \(K^* = 7\). We fix \(n = 30\) and choose the tuning parameter \(\tau = 0.1 \sqrt{\bar\lambda \log(n)/L}\).

\begin{figure}[htbp]
	\centering
	\begin{tabular}{ccc}
		\quad (a) Type I error (Alg.~\ref{alg: compute ci}) & \quad (b) Empirical Coverage (Alg.~\ref{alg: compute ci}) &(c) Type I error (Point Estimate)  \\
		\includegraphics[height = 0.28\textwidth]{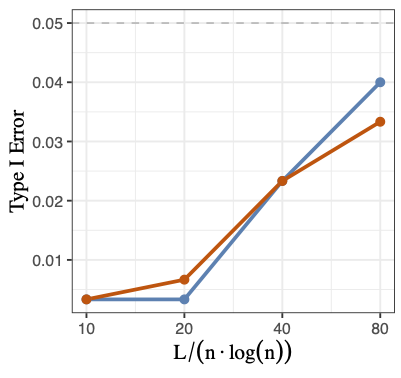}& \includegraphics[height = 0.28\textwidth]{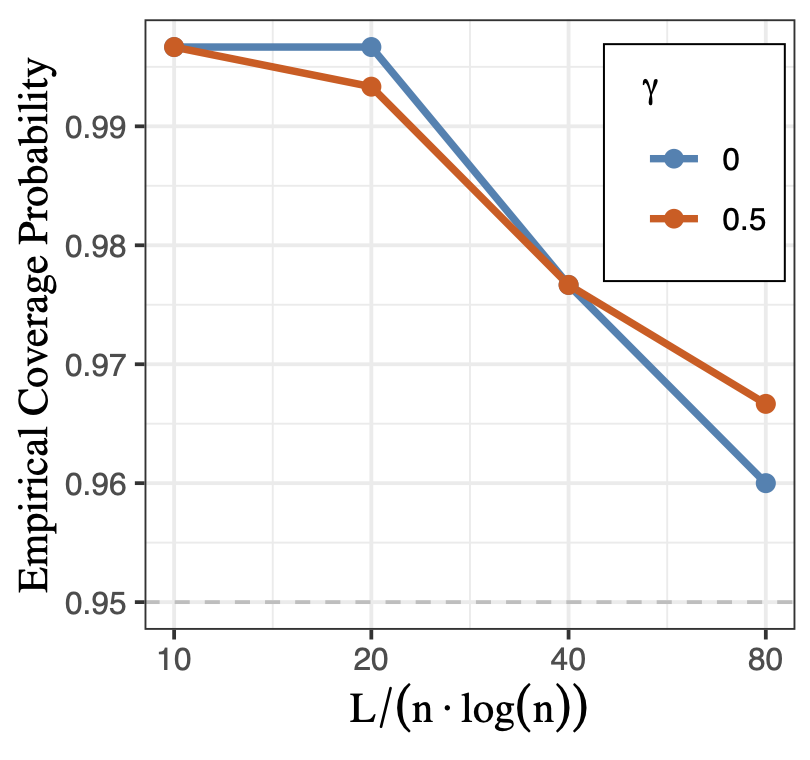} & \includegraphics[height = 0.28\textwidth]{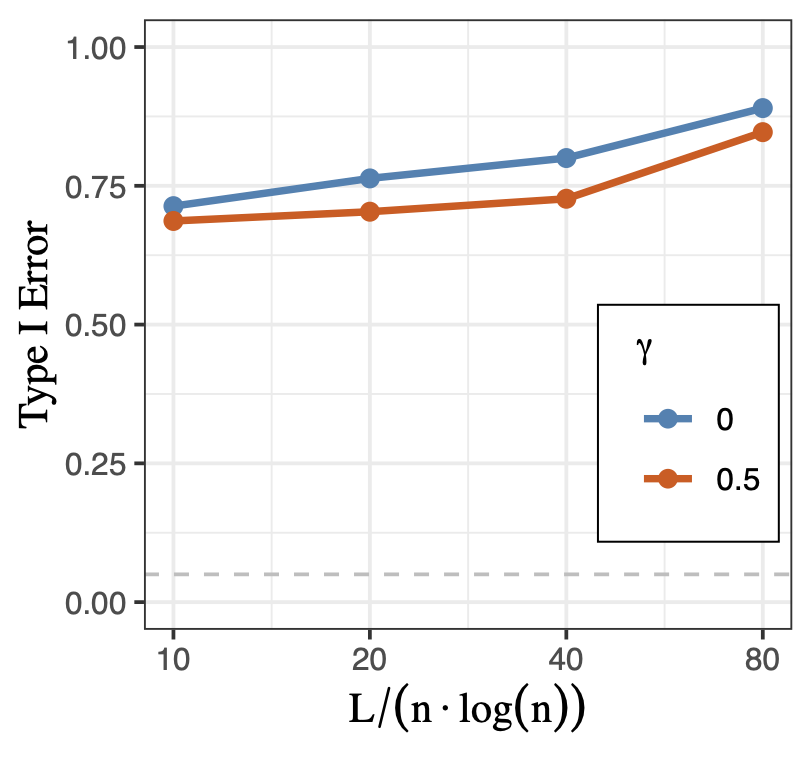}
	\end{tabular}
	\caption{Comparison of Type~I error and empirical coverage for Algorithm~\ref{alg: compute ci}, and Type~I error for the point estimator in~\eqref{eq: debias mle}, under Example~\ref{exm 2}. The total sample size \(L\) is scaled by \(n \log n\) for presentation. {For \(\gamma \in \{0,0.5\}\), the corresponding connectivity parameters are \(\kappa_{\bcE} \in \{1,2.07\}\), respectively.}}
	\label{fig: simu rev ordered}
\end{figure}

To generate revenue-ordered assortments, we set the sample size for each assortment \([k]\) to
\(L_{[k]} = L_{[n]} \cdot \gamma^{\,n-k}\), where $L_{[n]}$ denotes the sample size for the full assortment and $\gamma \in \{0,0.5\}$ controls the rate at which the sample size decays as the assortment size decreases. {Under this design, the corresponding connectivity parameters defined in Assumption~\ref{asp: connectivity graph} are $\kappa_{\bcE} \in \{1,2.07\}$, respectively.} We set \(A = \{2,4,6,8\}\) and evaluate the Type~I error under Example~\ref{exm 2} at total sample sizes \(L \in \{10, 20, 40, 80\} \cdot n \log n\). For the point estimator, we reject the null if and only if \(\hat{K} \notin \cK_0 = [7]\), where \(\hat{K} = \max\{k : \hat\Delta_k < 0\}\). The results are reported in Figure~\ref{fig: simu rev ordered}. Under this regime, Algorithm~\ref{alg: compute ci} maintains the Type~I error well below the nominal level of \(0.05\), whereas directly applying the point estimator without uncertainty quantification leads to substantially inflated Type~I error. 

We next examine the operational consequences of using point estimates to select an assortment. In each replication, we compare three decision rules. The first is the plug-in point-estimation rule
\[
    \widehat K=\sum_{k=1}^n \mathbf 1\{\widehat\Delta_k<0\}.
\]
The second selects the midpoint of the proposed confidence interval, rounded to the nearest integer:
\[
    \widehat K_{\mathrm{mid}}
    =
    \left\lfloor \frac{\widehat K_L+\widehat K_U}{2}+\frac{1}{2}\right\rfloor .
\]
The third is a best-in-confidence-set rule, which selects the assortment size in $[\widehat K_L,\widehat K_U]$ that attains the largest true revenue in the simulation. This rule is used as an oracle benchmark rather than an implementable decision rule. It measures whether the proposed confidence set retains high-revenue operational decisions that point-estimation rules may exclude due to random estimation error, thereby avoiding revenue loss from prematurely discarding such assortments.

\begin{figure}[t]
    \centering
    \includegraphics[width=0.8\linewidth]{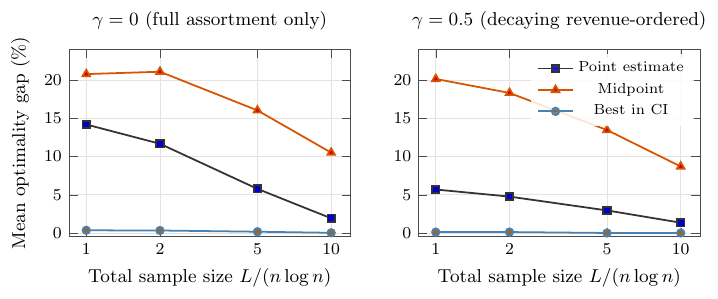}
    \caption{Mean percentage optimality gap under the point-estimation rules and the best-in-confidence-set benchmark on the constructed high-dispersion stress instance ($\kappa_r=15.0$) containing a popular high-revenue product at the boundary of the optimal assortment. The empirical coverage of the proposed confidence interval ranges from $0.982$ to $0.997$ across the displayed settings.}
    \label{fig:oper-pitfall-point-est}
\end{figure}

We report all revenue comparisons as percentage optimality gaps. For a selected assortment size $K$, define
\[
    \mathrm{gap}(K)
    =
    100 \times \frac{r(\cS^*)-r([K])}{r(\cS^*)} .
\]
We evaluate the three decision rules with 1{,}000 Monte Carlo replications per setting and focus on the limited-data grid $L \in \{1,2,5,10\}\cdot n\log n$, where the operational consequences of estimation error are most pronounced. We consider a high-dispersion regime in which revenues are drawn from the shifted exponential distribution $0.3 + 1.5\cdot\mathrm{Exp}(1)$, giving $\kappa_r$ roughly between $10$ and $20$, and examine a constructed stress instance from this regime (Figure~\ref{fig:oper-pitfall-point-est}).

The stress instance ($K^* = 5$, $\kappa_r = 15.0$) contains a popular high-revenue product at the boundary of the optimal assortment: product $K^*$ attains a $92\%$ choice share within the offered set $[K^*]$, so excluding it costs $17.1\%$ of the optimal revenue, and excluding the two boundary products costs $23.3\%$. At the same time, the boundary statistics $|\Delta_k|$ for $k \in \{K^*-2,\ldots,K^*+1\}$ are below $3\%$ of the median $|\Delta_k|$ in the instance, so the boundary decisions are statistically fragile. This combination is not accidental: since $\Delta_k$ compares $r_k$ only with the products ranked above $k$, the popularity of a boundary product raises the revenue stakes of its inclusion decision without increasing the statistical signal that identifies it. The instance is selected by a two-stage procedure that combines a structural screen with a Monte Carlo coverage check.

Figure~\ref{fig:oper-pitfall-point-est} shows that the three rules are well separated throughout the displayed sample-size range. At $\gamma=0$ and $L = n\log n$, the plug-in rule loses $14.2\%$ of the optimal revenue on average, and at least $5\%$ in $49\%$ of the replications. The midpoint rule loses $20.8\%$ on average (at least $5\%$ in $92\%$ of the replications), and its gap decays only slowly, still amounting to $16.0\%$ at $L = 5\,n\log n$ and $10.5\%$ at $L = 10\,n\log n$; the corresponding values at $\gamma=0.5$ are $20.2\%$, $13.5\%$, and $8.7\%$. In contrast, the best-in-confidence-set benchmark incurs a mean gap of at most $0.35\%$ in every cell, and the confidence interval is simultaneously informative and calibrated: its mean width decreases from about seven assortment sizes at $L = n\log n$ to about four at $L = 10\,n\log n$, and its empirical coverage ranges from $0.982$ to $0.997$ across the eight displayed cells, close to, and never below, the nominal $95\%$ level.

Taken together, these experiments identify the regimes in which relying on point estimation alone is operationally risky, namely limited or unbalanced historical data, dispersed revenues, and popular products near the optimal boundary. In precisely these regimes, the proposed confidence interval retains the high-revenue assortment sizes that point rules discard.

This experiment also clarifies the practical role of the proposed confidence interval. We do not view the confidence interval as merely a statistical object or as a replacement of the optimization problem by another point rule. Rather, the interval provides a set of statistically plausible assortment sizes that can be carried forward to downstream operational evaluation. In practice, a merchandiser may combine this set with additional business constraints, such as shelf-space limits, supplier agreements, inventory availability, or category-balance requirements. The value of the confidence interval is therefore to avoid prematurely discarding high-revenue assortment sizes that remain statistically plausible under the observed choice data.

\section{Conclusion}\label{sec: conclusion}
To conclude, we propose a general inferential framework for testing combinatorial properties of the optimal assortment, which, to the best of our knowledge, is the first such effort. Under the MNL model, we first estimate the latent preference scores from customer choice data among selected offer sets based on penalized likelihood optimization. Then we apply a Newton debiasing correction to the estimator and perform the assortment optimization algorithm by plugging in the debiased latent score estimator. Finally, we implement the Gaussian multiplier bootstrap to construct confidence intervals for the optimal offer set and then perform hypothesis testing on a given property of interest for the optimal assortment. We provide theoretical guarantees that our test is valid if the sample size is large enough under some mild conditions. 

For  future work, we plan to generalize the current results to other choice models, such as the capacitated MNL model with cardinality constraints, or the mixture of multinomial logit models (MMNL). We will also seek to develop a more computationally efficient and scalable method for constructing the confidence interval for $K^*$. 

\section{Data and Code Accessibility Statement}\label{sec: data and code}
{The results reported in this paper are reproducible from a replication package provided as supplemental material accompanying the article. The package contains three components. First, the R scripts implement the estimation and inference procedures of Section~\ref{sec: mtd}, including the penalized likelihood estimator, the Newton debiasing correction, and the bootstrap confidence interval of Algorithm~\ref{alg: compute ci}, and generate the simulation results of Sections~\ref{sec: simu ind dat ci}--\ref{sec: simu rev ordered}; the accompanying R Markdown notebooks collect the simulation outputs and produce Figures~\ref{fig:qq plot}--\ref{fig: simu rev ordered}. Second, the Python scripts produce the operational comparison in Figure~\ref{fig:oper-pitfall-point-est}, the evaluation of the connectivity parameter $\kappa_{\bcE}$ across historical-assortment designs in Table~\ref{tab:kappa-dgp-range}, the coverage experiment in Figure~\ref{fig:kappa-coverage-hard}, and the observed-assortment connectivity computations in Table~\ref{tab:public-data-kappa}. Third, a README file maps every figure and table of the paper to the script that produces it, lists the required software and packages, and describes the recommended execution order. All simulation scripts use fixed random seeds, so rerunning them reproduces the reported numbers exactly.

The numerical experiments in Section~\ref{sec: numerical results} are simulation based and require no external data. The empirical connectivity values in Table~\ref{tab:public-data-kappa} are computed from four publicly available product-choice data sets, none of which is redistributed in the package: the canned-tuna brand-choice data of \citet{kim1995modeling}, available in the R package \texttt{Ecdat} \citep{croissant2023ecdat}; the Ta Feng grocery transaction data in the construction provided by the Choice-Learn library \citep{auriau2024choicelearn}; the hotel booking data of \citet{bodea2009choice}, distributed in the R package \texttt{RMM} \citep{kim2022rmm}; and the tablet choice-based conjoint data from the ColumbiaX business analytics course materials \citep{columbiax2019tabletconjoint}. For each data set, the README file and the corresponding script document the public source and the exact construction of the observed offer sets used in the paper.
}
\clearpage
\ECswitch
\begin{center}
{\large\itshape Supplementary Materials to}\\[0.6em]
{\LARGE Combinatorial Inference on the Optimal Assortment in Multinomial Logit Models}
\end{center}
\bigskip





Before starting with the proofs, we introduce some useful notations to be used later. Recall $\cI = \{S \subseteq [n]_+: S = \cS_+, \cS \in \bcS \backslash \emptyset \}$ is the set of all non-empty offered assortments augmented by the no-purchase option, and we denote $\cS_+$ by $S$ for the convenience of notation. Define  $\cI_k = \{S \subseteq \cI : |S| = k\}$ to be the subsets of $\cI$ with cardinality equal to $k$, $2 \le k \le n+1$. Correspondingly, we let $\boldsymbol{\cE}_k = \{S \in \cI_k : \cE_S = 1\}$ for $k = 2, 3, \ldots, n+1$. Denote by $\II\{\cdot\}$ an indicator function of statements, which is equal to 1 if the statement inside $\{\cdot\}$ holds true and 0 otherwise. { For a sequence of random variables $\{X_n\}_{n=1}^{\infty}$, we write $X_n = o_P(1)$ if $X_n \overset{P}{\rightarrow} 0$, and we write $X_n = O_P(1)$ if for all $\epsilon >0$, there exists an $M > 0$ such that $\sup_n \PP(|X_n| > M) < \epsilon$.} { For a symmetric matrix $\Mb$, we define $\lambda_{\min, \perp} (\Mb)$ to be its smallest eigenvalue restricted to the eigenvectors orthogonal to $\mathbf{1}$:
$$
\lambda_{\min, \perp}(\Mb) = \min\{\lambda: \lambda = \vb^{\top} \Mb \vb, \text{ for all } \vb \text{ such that } \|\vb\|_2 = 1 \text{ and } \vb^{\top}\mathbf{1} = 0 \}.
$$} 
{
\section{Additional Empirical Evaluation Experiments}\label{sec: add simu}
We conduct additional numerical experiments to assess the robustness of the proposed method with respect to the connectivity structure of the observed assortment graph. Specifically, we first report the values of $\kappa_{\bcE}$ under a range of data-generating schemes, from uniformly generated assortments to revenue-ordered assortments and intermediate cases. We then examine how the connectivity parameter $\kappa_{\bcE}$ in Assumption~\ref{asp: connectivity graph} affects the empirical coverage of the proposed confidence interval. Finally, we report the corresponding weighted-connectivity quantities, including $\underline{\lambda}$ and $\kappa_{\bcE}$, for several publicly available choice data sets.
\subsection{Empirical Values of Connectivity Parameter Across Historical-Assortment Designs}
\label{sec:kappa-dgp-range}
To provide a more intuitive understanding of how $\kappa_{\bcE}$ varies across different data-generating regimes, we conduct an additional simulation that varies only the historical-assortment generation process and reports the corresponding values of $\kappa_{\bcE}$.
We set $n=30$, as in the main simulation study, and generate $L=\lceil 10 n\log n\rceil=1,021$ observed historical assortments in each Monte Carlo replication.  Each offer set includes the no-purchase option.  Let $\rho_0\in\{0,0.25,0.5,0.75,1\}$ denote the probability that an offer set is revenue ordered.  Conditional on $\rho_0$, each historical offer set is generated independently as follows: with probability $1-\rho_0$, we draw a nonempty subset of $[n]$ uniformly at random; with probability $\rho_0$, we draw a revenue-ordered assortment $[k]$ with $k$ uniformly distributed on $\{1,\ldots,n\}$.  For each generated historical-assortment structure, we construct the weighted comparison-graph Laplacian in Assumption~\ref{asp: connectivity graph} and compute $\kappa_{\bcE}$.
We use 200 Monte Carlo replications for each value of $\rho_0$.

\begin{table}[htbp]
\centering
\caption{Empirical Values of $\kappa_{\bcE}$ under Random, Mixture, and Revenue-Ordered Historical-Assortment Designs}
\label{tab:kappa-dgp-range}
\begin{tabular}{lccc}
\hline
Historical-assortment design & $\rho_0$ & Mean of $\kappa_{\bcE}$ & SD of $\kappa_{\bcE}$ \\
\hline
Uniform random          & 0.00 & 2.22  & 0.04  \\
25\% revenue-ordered    & 0.25 & 3.08  & 0.11  \\
50\% revenue-ordered    & 0.50 & 5.03  & 0.32  \\
75\% revenue-ordered    & 0.75 & 10.68 & 0.95  \\
Revenue-ordered         & 1.00 & 98.90 & 17.72 \\
\hline
\end{tabular}
\end{table}


Table~\ref{tab:kappa-dgp-range} shows that $\kappa_{\bcE}$ increases substantially as the historical-assortment design moves from uniformly random assortments toward revenue-ordered assortments. This pattern is consistent with the role of $\kappa_{\bcE}$ in the theory: uniformly random assortments induce a relatively well-balanced comparison graph, whereas revenue-ordered offer sets concentrate comparisons along a nested structure and lead to a more ill-conditioned historical-assortment design. In the next section, we further examine the robustness of the proposed tests with respect to changes in $\kappa_{\bcE}$.

\subsection{Sensitivity to Observed-Assortment Connectivity}\label{sec:kappa-sensitivity}

To examine the empirical role of the connectivity parameter $\kappa_{\bcE}$, we conduct an additional experiment under revenue-ordered historical assortments. We set $n=30$ and $K^*=7$, and use the same total sample-size scalings $L/(n\log n)\in\{10,20,40,80\}$ as in Section~\ref{sec: simu rev ordered}.

All historical assortments are revenue-ordered. To isolate the effect of the global connectivity structure from the amount of local information near the true optimal assortment, we allocate $40\%$ of the observations evenly across the assortments $[5],\ldots,[9]$, which bracket $[K^*]=[7]$. The remaining $60\%$ of the observations are allocated across the tail assortments $[10],\ldots,[30]$, with the sample size assigned to prefix assortment $[k]$ proportional to $\gamma^{30-k}$, where
\[
    \gamma\in\{0,0.1,0.25,0.5,0.75,0.9,0.98,1\}.
\]
This design varies the conditioning of the weighted comparison graph while keeping the amount of local boundary information approximately fixed. To make the threshold nontrivial to recover, we calibrate the two boundary revenue contrasts so that $\Delta_7=-0.02$ and $\Delta_8=0.02$; thus, the adjacent revenue contrasts around the true threshold are close to zero.

For each pair $(\gamma,L/(n\log n))$, we run 200 Monte Carlo replications using the regularization rule from Section~\ref{sec: simu rev ordered} and 200 bootstrap draws. We evaluate the empirical coverage of the proposed confidence interval by
\[
    \II\{K^*\in[\hat{K}_L,\hat{K}_U]\}.
\]
Across the designs considered here, the resulting values of $\kappa_{\bcE}$ range from $3.67$ to $90.68$. Figure~\ref{fig:kappa-coverage-hard} plots the empirical coverage probability against $\kappa_{\bcE}$. The coverage remains close to the nominal $95\%$ level over the examined range of $\kappa_{\bcE}$.

While the theoretical results in Section~\ref{sec: mtd} show that the error bound and sample-size requirements become more stringent as $\kappa_{\bcE}$ grows, such dependence serves only as a worst-case, nonasymptotic diagnostic. It does not imply that empirical coverage must decrease monotonically with $\kappa_{\bcE}$ in every finite-sample design. The present experiment shows that, across a broad observed range of $\kappa_{\bcE}$, the proposed confidence interval remains well calibrated. This stability is consistent with the fact that the confidence-interval construction incorporates increased uncertainty through the estimated standard errors and bootstrap critical values.

\begin{figure}[htbp]
    \centering
    \includegraphics[width=0.7\linewidth]{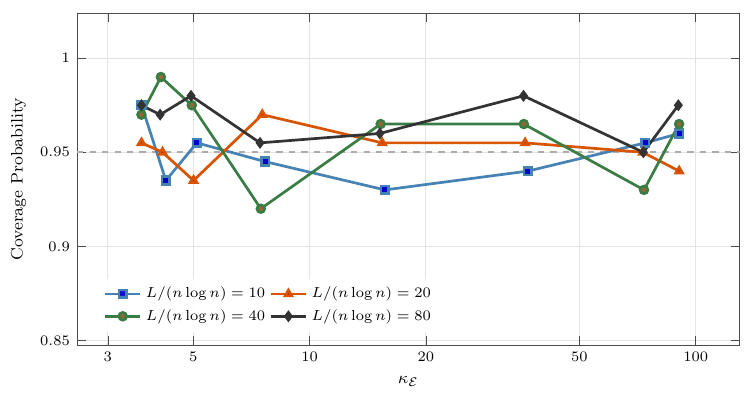}
    \caption{Empirical coverage of the proposed confidence interval as a function of $\kappa_{\bcE}$. The coverage remains close to the nominal level $0.95$ across the examined range of $\kappa_{\bcE}$.}
    \label{fig:kappa-coverage-hard}
\end{figure}
\subsection{Empirical Values of Observed-Assortment Connectivity on Public Product-Choice Data Sets}
\label{sec:public-data-kappa}

We further evaluate $\kappa_{\bcE}$ on public product-choice data sets commonly used in demand-estimation and choice-modeling applications, where observed choices are naturally modeled by multinomial-logit-type demand models and observed prices serve as revenue proxies. For each data set, we construct the weighted comparison graph from the observed or reconstructed choice sets, using the multiplicity of each historical assortment as its weight and adding the no-purchase option to each offer set, in accordance with \eqref{eq: con graph mat}.

The Tuna data set is a brand-choice data set for canned tuna.  It is available through the \texttt{Ecdat} package \citep{croissant2023ecdat} and originates from the retail-pricing application in \citet{kim1995modeling}.  The data contain prices and promotional variables for five tuna alternatives; under the standard brand-choice construction, all five alternatives are available in each observed choice occasion.  The Ta Feng grocery data, accessed through the Choice-Learn construction \citep{auriau2024choicelearn}, provide transaction-level grocery purchases and prices.  The public assortment construction used in Choice-Learn retains 25 products and marks all retained products as available in each unit-level choice.  The Hotel~1 data of \citet{bodea2009choice}, distributed through the CRAN \texttt{RMM} package \citep{kim2022rmm}, are the closest match to our revenue-management setting: each observation is a hotel booking decision with an observed exposed room-type choice set and room-specific nightly rates.  Finally, the tablet conjoint data from the ColumbiaX business analytics course materials \citep{columbiax2019tabletconjoint} consist of choice-based conjoint tasks in which respondents choose among tablet profiles described by product attributes and price, and we treat each distinct product-price profile as a separate product.

Table~\ref{tab:public-data-kappa} reports the resulting values of $\underline{\lambda}$ and $\kappa_{\bcE}$. The Tuna and Ta Feng constructions use complete assortments; hence, their comparison graphs are fully connected and yield $\kappa_{\bcE}=1$. The Hotel~1 data contain 12 distinct observed historical assortments and yield $\underline{\lambda}=0.0124$ and $\kappa_{\bcE}=12.41$. The tablet conjoint data contain genuinely varying product sets and yield a much smaller $\underline{\lambda}=0.000304$ and a much larger $\kappa_{\bcE}=620.51$, reflecting substantially weaker weighted connectivity in the induced comparison graph.

Overall, the public-data evidence supports the role of $\kappa_{\bcE}$ as a diagnostic for the informativeness of historical assortment variation.  When the public construction exposes every product in every choice occasion, as in Tuna and Ta Feng, the comparison graph is perfectly balanced and $\kappa_{\bcE}=1$.  In contrast, data sets with genuinely varying offer sets, such as Hotel~1 and the tablet conjoint data, generate nontrivial and sometimes large values of $\kappa_{\bcE}$, together with smaller values of $\underline{\lambda}$, reflecting weaker weighted connectivity. Together with the sensitivity experiment in Section~\ref{sec:kappa-sensitivity}, these results provide an empirical interpretation of $\kappa_{\bcE}$ and illustrate how the proposed method behaves across different connectivity regimes.

\begin{table}[htbp]
\centering
\caption{Empirical values of observed-assortment connectivity on public product-choice data sets.}
\label{tab:public-data-kappa}
\begin{tabular}{lcrrrr}
\hline
Data set & Assortment scheme & $n$ & $L$ & $\underline\lambda$ & $\kappa_{\bcE}$ \\
\hline
Tuna & Full assortment & 5 & 13,705 & 0.1667 & 1.00 \\
Ta Feng grocery & Full assortment & 25 & 14,215 & 0.0385 & 1.00 \\
Hotel~1 & Partially observed & 10 & 1,100 & 0.0124 & 12.41 \\
Tablet conjoint & Partially observed & 350 & 2,055 & 0.000304 & 620.51 \\
\hline
\end{tabular}
\end{table}

}
\section{Proof of Theorem \ref{thm: inf norm MLE}}\label{sec: proof thm inf norm mle}

The proof extends the results presented in Theorem 6 of \cite{chen2019topk} to the context of the multinomial logit model. { The major technical difference lies in the characterization of the new eigenstructure of the Hessian matrix, which is more complex than that of the BTL model. We employ a different approach to control the degenerating rate of the eigenvalues of the Hessian matrix and utilize the new eigenstructure to derive the convergence rate.} Note that if we take $\thetab' = \ponep \thetab$, then the convex problem \eqref{eq: convex prob 0} can be equivalently written as 
\begin{equation}\label{eq: convex prob}
	\min_{\thetab' \in \RR^{n+1}, \mathbf{1}^{\top}\thetab' = 0} \cL_{\tau}(\thetab'; \bx) := \cL(\thetab'; \bx) + \frac{\tau}{2}\|\thetab'\|_2^2.
\end{equation}
Later we will show that the minimizer of $\cL(\thetab'; \bx) + \frac{\tau}{2}\|\thetab'\|_2^2$ belongs to the subspace perpendicular to $\mathbf{1}$, and thus we can simplify \eqref{eq: convex prob} as 
\begin{equation}\label{eq: convex prob 1}
	\min_{\thetab \in \RR^{n+1}} \cL_{\tau}(\thetab; \bx) := \cL(\thetab; \bx) + \frac{\tau}{2}\|\thetab\|_2^2.
\end{equation}
It can be observed that the regularized MLE to \eqref{eq: convex prob 1} will be $\hat\thetab' := \ponep \hat\thetab$, where $\hat\thetab$ is the regularized MLE of problem \eqref{eq: convex prob 0}. In the following proof, we will consider solving \eqref{eq: convex prob 1} instead. We will show the $L_{2}$ convergence rate of the resulting MLE of \eqref{eq: convex prob 1}, i.e., $\|\hat\thetab' - \ponep \thetab^* \|_2 = \|\ponep\hat\thetab - \ponep \thetab^* \|_2 $. Since $\hat\thetab'$ can be easily transformed back to $\hat\thetab$ by taking $\hat\theta_i = \hat\theta_i' - \hat\theta'_0$ for $i \in [n]_+$, the $L_{2}$ convergence rates $\|\hat\thetab' - \ponep \thetab^* \|_2$ and $\|\hat\thetab - \thetab^* \|_2$ will be different only by a factor of constant. Besides, since $\thetab^*$ and $\ponep \thetab^*$ essentially specify the same MNL model, without loss of generality, we will replace $\thetab^*$ by $\ponep \thetab^*$ throughout the proofs, and we will abuse the notation and let $\thetab^*$ denote $\thetab^{*\prime} =  \ponep \thetab^*$ and let $\hat\thetab$ denote $\hat\thetab' = \ponep \hat\thetab$, i.e., the resulting regularized MLE solving problem \eqref{eq: convex prob 1}. Before delving into the details, we need the following lemmas to help with the proof.


The following lemmas depict the smoothness and convexity of the log-likelihood function,
{
\begin{lemma}\label{lm: eigen hessian}
    Suppose that Assumption~\ref{asp: connectivity graph} holds.
    For all $\thetab \in \RR^{n+1}$ such that $\|\thetab - \thetab^*\|_{\infty} \le C$ for some $C \ge 0$, 
    we have
	\begin{equation}\label{eq: eigen hessian}
		\lambda_{\min, \perp} \!\left(\nabla^2 \cL_{\tau}(\thetab; \bx)\!\right)\! \ge\! \tau + \frac{1}{(\kappa_{\thetab} e^{2C})^2} \underline{\lambda}, \,\, \lambda_{\max}\!\left(\nabla^2 \cL_{\tau}(\thetab; \bx)\!\right)\! \le \tau + { (\kappa_{\thetab} e^{2C})^2} \bar\lambda.
	\end{equation}
\end{lemma}
Please see Section \ref{sec: proof lm eigen hessian} for the proof of Lemma \ref{lm: eigen hessian}.
\begin{lemma}\label{lm: bound gradient}
	Under Assumption~\ref{asp: connectivity graph} and the conditions that $n \bar\lambda L \gtrsim \log n$ and $\tau \asymp \sqrt{\bar\lambda \log n/L}$, the following event  
	\begin{equation}\label{eq: bound gradient}
		\cA_{g} := \left\{\|\nabla \cL_{\tau}(\thetab^*; \bx)\|_2 \lesssim  \sqrt{\frac{n \bar\lambda\log n}{L}} \right\}
	\end{equation}
	occurs with probability at least $1 - O(n^{-10}) $.
\end{lemma}
Please refer to Section \ref{sec: proof lm bound gradient} for the proof of Lemma \ref{lm: bound gradient}.
\begin{lemma}\label{lm: smooth hessian}
	Under Assumption~\ref{asp: connectivity graph},
    we have that for any $\thetab \in \RR^{n+1}$ the following event holds
	\begin{equation}\label{eq: smooth hessian}
		\cA_{h} := \left \{\|\nabla^2 \cL_{\tau} (\thetab; \bx)\|_2 \le \tau + (n+1)^2 \bar\lambda/4\right\}. 
	\end{equation}
\end{lemma}}
Please see Section~\ref{sec: proof lm smooth hessian} for the proof of Lemma~\ref{lm: smooth hessian}.
Now we begin with the proof. As in the proof of \cite{chen2019topk}, we let $\thetab^T$ be the output of the following gradient descent
\begin{itemize}
	\item Initialize $\thetab^0 = \thetab^*$,
	\item for $t = 0,1, \ldots, T-1$,
	$$
	\thetab^{t+1} = \thetab^t - \eta \nabla \cL_{\tau}(\thetab^t; \bx),
	$$
	where $\eta$ is taken to be { $1/(\tau + (n+1)^2 \bar\lambda/4)$.}
\end{itemize}
\subsection*{Step I}
By Lemma \ref{lm: smooth hessian}, we know that 
$\cL_{\tau}(\thetab; \bx)$ is { $(\tau +  (n+1)^2 \bar\lambda/4)$-smooth} and $\tau$-strongly convex for any $\thetab \in \RR^{n+1}$, then by Theorem 3.10 in \cite{bubeck2015convex} we have 
$$
\|\thetab^t - \hat\thetab\|_2 \le \rho^t \|\thetab^* - \hat\thetab\|_2,
$$
where { $\rho = 1 - \frac{\tau}{\tau + (n+1)^2 \bar\lambda/4}$}. Note that since $\mathbf{1}^{\top} \btheta^* = 0$ and $\mathbf{1}^{\top}\nabla \cL(\thetab; \bx) = \mathbf{0}$ for any $\thetab \in \RR^{n+1}$, it follows by simple induction that $\mathbf{1}^{\top}\thetab^{t} = 0$ for all $t \ge 1$, and hence $\hat\thetab^{\top} \mathbf{1}= 0$ due to the convergence of $\thetab^t$ to $\hat\thetab$. The following lemma provides an initial upper bound for $\|\thetab^* - \hat\thetab\|_2$. 
\begin{lemma}\label{lm: initial bound MLE}
	On the event $\cA_g$ as defined in \eqref{eq: bound gradient}, there exists a constant $c > 0$ such that 
	$$
	\|\hat\thetab - \thetab^*\|_2 \le c \sqrt{n} .
	$$
\end{lemma}
The proof is deferred to Section \ref{sec: proof lm initial bound}. Recall that { $\tau \asymp  \sqrt{\frac{\bar\lambda \log n}{L}}$}, then with Lemma \ref{lm: initial bound MLE}, under the event $\cA_g$, for large enough $T$ we have 
{
\begin{align*}
	\|\thetab^T - \hat\thetab\|_2 & \le \rho^T\|\hat\thetab - \thetab^*\|_2 \lesssim \left(1 - \frac{\tau}{\tau + (n+1)^2 \bar\lambda/4}\right)^T \sqrt{n}\\
	& \le \exp\left( - \frac{T\tau}{\tau +  (n+1)^2 \bar\lambda/4} \right)\sqrt{n} \le \exp\left(-c_3 T \sqrt{\frac{\log n}{n^4 \bar\lambda  L }}\right)\sqrt{n}\\
	& \lesssim \sqrt{\frac{n \kappa_{\bcE} \log n}{\underline{\lambda}L}}.
\end{align*}}

\subsection*{Step II}
\if1\sharpbound{
	Similar as the proof of Theorem 6 in \cite{chen2019topk}, we introduce the leave-one-out sequence $\{\thetab^{t, (m)}\}$ for $m \in [n]_+$
	$$
	\thetab^{t+1,(m)} = \thetab^{t, (m)} - \eta \nabla  \cL_{\tau}^{(m)}(\thetab^{t,(m)}), 
	$$
	where $\thetab^{0,(m)} = \thetab^*$ and 
	\begin{align*}
		& \cL_{\tau}^{(m)}(\thetab; \bx) = - \sum_{S \in \bcE, m \notin S} \left\{ \sum_{i \in \cS_+} x_{S}^{(i)}\theta_i - \log \left(\sum_{i \in \cS_+}e^{\theta_i}\right) \right\}\\
		& \quad - p\!\!\!\sum_{S \in \cI, m \in S} \!\!\left\{ \sum_{i \in \cS_+} \left(\frac{e^{\theta_i^*}\theta_i}{\sum_{j \in \cS_+} e^{\theta_j^*}}\right) - \log \left(\sum_{i \in \cS_+}e^{\theta_i}\right) \right\}  + \frac{\tau}{2}\|\thetab\|^2. 
	\end{align*}
}\fi
Now we show the convergence rate of $\|\thetab^T - \thetab^*\|_2$ by induction. More specifically, we propose that with probability at least $1 - O(n^{-10})$, the following holds true for some constant $C_1 >0$ for any $t \in [T]_+$, 
{ \begin{equation}\label{eq: ind hypo 1}
	\|\thetab^t - \thetab^*\|_2 \le C_1  \sqrt{\frac{n \kappa_{\bcE} \log n}{\underline{\lambda}L}}.
\end{equation}}
Note that the base case $t = 0$ holds true trivially because of our setting of $\thetab^0 = \thetab^*$. Now suppose that \eqref{eq: ind hypo 1} holds true for the $t$-th iteration, we will show that under the event $\bcA_g$ (occurring with probability at least $1 - O(n^{-10})$), the following holds
{ $$
\|\thetab^{t+1} - \thetab^*\|_2 \le C_1  \sqrt{\frac{n \kappa_{\bcE} \log n}{\underline{\lambda}L}}.
$$}
First note that if \eqref{eq: ind hypo 1} holds for the $t$-th iteration, we have 
{ $$\|\thetab^t - \thetab^*\|_{\infty} \le \|\thetab^t - \thetab^*\|_2 \le   C_1  \sqrt{\frac{n \kappa_{\bcE} \log n}{\underline{\lambda}L}}.$$
}The rest of the proof follows the proof of Lemma 14 in \cite{chen2019topk} with modifications. 
From the gradient descent algorithm we know that 
\begin{align*}
	\boldsymbol{\theta}^{t+1}-\boldsymbol{\theta}^{*} &=\boldsymbol{\theta}^{t}-\eta \nabla \mathcal{L}_{\tau}\left(\boldsymbol{\theta}^{t};\bx\right)-\boldsymbol{\theta}^{*} \\
	&=\boldsymbol{\theta}^{t}-\eta \nabla \mathcal{L}_{\tau}\left(\boldsymbol{\theta}^{t};\bx\right)-\left[\boldsymbol{\theta}^{*}-\eta \nabla \mathcal{L}_{\tau}\left(\boldsymbol{\theta}^{*};\bx\right)\right]-\eta \nabla \mathcal{L}_{\tau}\left(\boldsymbol{\theta}^{*};\bx\right) \\
	&=\left\{\boldsymbol{I}_{n}-\eta \int_{0}^{1} \nabla^{2} \mathcal{L}_{\tau}(\boldsymbol{\theta}(\nu);\bx) \mathrm{d} \nu\right\}\left(\boldsymbol{\theta}^{t}-\boldsymbol{\theta}^{*}\right)-\eta \nabla \mathcal{L}_{\tau}\left(\boldsymbol{\theta}^{*};\bx\right),
\end{align*}
where $\thetab(\nu) := \thetab^* + \nu(\thetab^t - \thetab^*)$. Then we can see that { if $C_1  \sqrt{{n \kappa_{\bcE} \log n}/({\underline{\lambda}L})} \le \epsilon$ }for a sufficiently small constant $\epsilon$, we have that $\|\thetab(\nu) - \thetab^*\|_{\infty} \le \nu\|\thetab^t - \thetab^*\|_{\infty} \le \epsilon $. Then by Lemma \ref{lm: eigen hessian} we have that 
    for all $\nu \in [0,1]$
{ 
$$
\tau + \frac{1}{2\kappa_{\thetab}^2} \underline{\lambda}\le \lambda_{\min, \perp}\left(\nabla^{2} \mathcal{L}_{\tau}(\boldsymbol{\theta}(\nu);\bx)\right) \le \lambda_{\max}\left(\nabla^{2} \mathcal{L}_{\tau}(\boldsymbol{\theta}(\nu);\bx)\right) \le \tau + 2{\kappa_{\thetab}^2} \bar{\lambda}.
$$}
Now we denote $\Bb = \int_{0}^{1} \nabla^{2} \mathcal{L}_{\tau}(\boldsymbol{\theta}(\nu);\bx) \mathrm{d} \nu$, then we have 
\begin{align*}
	\Bb \mathbf{1}_{n+1} &= \int_{0}^{1} \nabla^{2} \mathcal{L}_{\tau}(\boldsymbol{\theta}(\nu);\bx) \mathbf{1}_{n+1} \mathrm{d} \nu = \mathbf{0};\\
	\lambda_{\min , \perp} (\Bb ) & = \min_{\vb \in \Sb^{n}, \vb^{\top} \mathbf{1} = 0} \int_{0}^{1} \vb^{\top} \nabla^{2} \mathcal{L}_{\tau}(\boldsymbol{\theta}(\nu);\bx) \vb \mathrm{d} \nu \ge  \int_{0}^{1} \lambda_{\min, \perp}\left( \nabla^{2} \mathcal{L}_{\tau}(\boldsymbol{\theta}(\nu);\bx) \right) \mathrm{d} \nu\\
	& \ge \tau + \frac{1}{2\kappa_{\thetab}^2}\underline{\lambda} ; \\
	\lambda_{\max}(\Bb) & = \max_{\vb \in \Sb^n} \int_{0}^{1} \vb^{\top} \nabla^{2} \mathcal{L}_{\tau}(\boldsymbol{\theta}(\nu);\bx) \vb \mathrm{d} \nu \le \tau + {2\kappa_{\thetab}^2} \bar{\lambda}. 
\end{align*}
Besides, by the definition of $\eta$ we have that { $1-\eta\left(\tau + \frac{1}{2\kappa_{\thetab}^2}\underline{\lambda} \right) \ge 1 - \eta \left(\tau + {2\kappa_{\thetab}^2} \bar{\lambda}\right) \ge 0$.} Combining the above results we have
{ \begin{align*}
	\|\thetab^{t+1} - \thetab^*\|_2 &\le \left(1-\frac{1}{2 \kappa_{\thetab}^2} \eta \underline{\lambda}\right)\|\thetab^t - \thetab^*\|_2 + \eta\|\nabla \mathcal{L}_{\tau}\left(\boldsymbol{\theta}^{*};\bx\right)\|_2\\
	& \le \left(1-\frac{1}{2 \kappa_{\thetab}^2} \eta \underline{\lambda}\right) C_1 \sqrt{\frac{n \kappa_{\bcE} \log n}{\underline{\lambda}L}} + C \eta \sqrt{\frac{n \bar\lambda\log n}{L}}\\
	& \le C_1  \sqrt{\frac{n \kappa_{\bcE} \log n}{\underline{\lambda}L}},
\end{align*}}
so long as $C_1$ is large enough.

\if1\sharpbound{
	We propose the following induction hypotheses to be proved later. More specifically, we propose that with probability at least $1 - O(n^{-C_L})$, the following hold for each $t \in [T]$,
	\begin{align}
		\|\thetab^t - \thetab^*\| & \le C_1 \sqrt{\frac{\log n}{M_n pL }}, \label{eq: ind hypo 1}\\
		\max_{m \in [n+1]}|\theta^{t,(m)}_m - \theta^*_m| & \le  C_2 \sqrt{\frac{\log n}{n M_n p L}} , \label{eq: ind hypo 2}\\
		\max_{m \in [n+1]}\|\thetab^t - \thetab^{t, (m)}\| & \le C_3  \sqrt{\frac{ \log n}{M_n p L}} \label{eq: ind hypo 3}\\
		\|\thetab^t - \thetab^*\|_{\infty} & \le C_4 \sqrt{\frac{\log n}{ M_n p L}} \label{eq: ind hypo 4}
	\end{align}
	
	Note that induction hypotheses \ref{eq: ind hypo 3} and \ref{eq: ind hypo 4} indicates that there exists a constant $C_5 \ge C_3 + C_4$ such that with probability at least $1 - O(n^{-C_L})$
	\begin{align*}
		\max_{m \in [n+1]} \|\thetab^{t,(m)} - \thetab^*\|_{\infty} & \le \max_{m \in [n+1]}\|\thetab^t - \thetab^{t, (m)}\| + \|\thetab^t - \thetab^*\|_{\infty}\\
		& \le  C_3 \sqrt{\frac{ \log n}{M_n p L}} + C_4 \sqrt{\frac{\log n}{ M_n p L}}\\
		& \le C_5 \sqrt{\frac{\log n}{ M_n p L}}.
	\end{align*}
	Similarly, combining induction hypotheses \ref{eq: ind hypo 1} and \ref{eq: ind hypo 3}, there exists a constant $C_6 \ge C_3 + C_1 $ such that with probability at least $1-O(n^{-C_L})$
	\begin{align*}
		\max_{m \in [n+1]} \|\thetab^{t,(m)} - \thetab^*\| & \le \max_{m \in [n+1]}\|\thetab^t - \thetab^{t, (m)}\| + \|\thetab^t - \thetab^*\|\\
		& \le  C_3 \sqrt{\frac{ \log n}{M_n p L}} + C_1 \sqrt{\frac{\log n}{M_n pL }}\\
		& \le C_6  \sqrt{\frac{\log n}{M_n pL }}.
	\end{align*}
	Since $\thetab^{0,(m)} = \thetab^0 = \thetab^*$ for all $m \in [n+1]$, the base case $t = 0$ holds true trivially. The induction steps are provided in the following lemmas. 
	\begin{lemma}\label{lm: ind hypo 1}
		If \eqref{eq: ind hypo 1} to \eqref{eq: ind hypo 4} hold true for the $t$-th iteration, then with probability at least $1- O(n^{-C_L})$, we have 
		$$
		\|\thetab^t - \thetab^*\| \le C_1 \sqrt{\frac{\log n}{M_n pL }}.
		$$
	\end{lemma}
	\begin{lemma}\label{lm: ind hypo 2}
		If \eqref{eq: ind hypo 1} to \eqref{eq: ind hypo 4} hold true for the $t$-th iteration, then with probability at least $1- O(n^{-C_L})$, we have 
		$$
		\max_{m \in [n]_+}|\theta^{t,(m)}_m - \theta^*_m| \le  C_2 \sqrt{\frac{\log n}{n M_n p L}} .
		$$
	\end{lemma}
	\begin{lemma}\label{lm: ind hypo 3}
		If \eqref{eq: ind hypo 1} to \eqref{eq: ind hypo 4} hold true for the $t$-th iteration, then with probability at least $1- O(n^{-C_L})$, we have 
		$$
		\max_{m \in [n]_+}\|\thetab^t - \thetab^{t, (m)}\| \le C_3  \sqrt{\frac{ \log n}{M_n p L}}.
		$$
	\end{lemma}
	\begin{lemma}\label{lm: ind hypo 4}
		If \eqref{eq: ind hypo 1} to \eqref{eq: ind hypo 4} hold true for the $t$-th iteration, then with probability at least $1- O(n^{-C_L})$, we have 
		$$
		\|\thetab^t - \thetab^*\|_{\infty} \le C_4  \sqrt{\frac{\log n}{ M_n p L}}.
		$$
	\end{lemma}
} \fi
\subsection*{Step III}
Combining the results in Step I and Step II, as $T$ goes to infinity we have that with probability at least $1 - O(n^{-10})$, 
{
$$
\|\thetab^T - \thetab^*\|_{2} \lesssim \sqrt{\frac{n \kappa_{\bcE} \log n}{\underline{\lambda}L}},
$$
and
$$
\|\hat\thetab - \thetab^*\|_{2} \le \|\thetab^T - \hat\thetab\|_{2} + \|\thetab^T - \thetab^*\|_{2} \lesssim \sqrt{\frac{n \kappa_{\bcE} \log n}{\underline{\lambda}L}}.
$$}

\section{Proof of Theorem \ref{thm: asymp normal}}\label{sec: proof thm asymp normal}
Before we begin with the proof, we first propose the following lemmas that provide bounds for several terms to be used in the proof. 
\begin{lemma}\label{lm: cck term bounds}
	Under the same conditions as Theorem \ref{thm: inf norm MLE}, for any $\thetab \in \RR^{n+1}$ such that $\|\thetab - \thetab^*\|_{\infty} < C$ for some constant $C > 0$, with probability at least $1 - O(n^{-10})$ we have the following bounds 
	\begin{align}
		&\|\nabla \cL(\thetab^*; \bx)\|_{\infty} \lesssim { \sqrt{\frac{ \bar\lambda\log n}{L}} + \frac{\log n}{L}}, \label{eq: cck term 1}\\
		&\|\nabla \cL(\hat\thetab; \bx) - \nabla \cL(\thetab^*; \bx) - \nabla^2 \cL(\thetab^*; \bx)(\hat\thetab - \thetab^*)\|_{\infty} \lesssim \frac{{ \kappa_{\bcE}^2} n\log n}{L},\label{eq: cck term 2}\\
		& \|\nabla^2 \cL(\hat\thetab ; \bx) - \nabla^2 \cL(\thetab^*; \bx)\|_{\infty} \lesssim { \sqrt{\frac{\kappa_{\bcE}^2 n \bar\lambda\log n}{ L}} }, \label{eq: cck term 3}\\
		& \left\|\begin{pmatrix}
			\nabla^2 \cL(\hat\thetab; \bx) & \mathbf{1} \\
			\mathbf{1}^{\top} & 0
		\end{pmatrix}^{-1} - \begin{pmatrix}
			\nabla^2 \cL(\thetab^*; \bx) & \mathbf{1} \\
			\mathbf{1}^{\top} & 0
		\end{pmatrix}^{-1}\right\|_2 \lesssim {  \frac{n}{\underline{\lambda}} \sqrt{\frac{\kappa_{\bcE}^3\log n}{\underline{\lambda} L}}}. \label{eq: cck term 5}
	\end{align}
\end{lemma}
The proof of Lemma \ref{lm: cck term bounds} is deferred to Section \ref{sec: proof lm cck term bounds}. The following corollary of Lemma \ref{lm: cck term bounds} helps characterize the Moore-Penrose inverse of the Hessian matrix. 
\begin{corollary}\label{col: bTheta entry bound}
	For any $\thetab \in \RR^{n+1}$, we have that 
	$$
	\begin{pmatrix}
		\nabla^2 \cL(\thetab; \bx) & \mathbf{1} \\ \mathbf{1}^{\top} & 0
	\end{pmatrix}^{-1} =  \begin{pmatrix}
		\nabla^2 \cL(\thetab; \bx)^{\dagger} & \frac{1}{n+1}\mathbf{1}\\ \frac{1}{n+1}\mathbf{1}^{\top} & 0 
	\end{pmatrix} .
	$$
	{ Besides, for $\thetab \in \RR^{n+1}$ such that $\|\thetab - \thetab^*\|_{\infty} < C$ for some constant $C>0$, 
    we have that
	$$\frac{1}{\bar\lambda} \lesssim [\nabla^2 \cL(\thetab; \bx)^{\dagger}]_{jj} \lesssim  \frac{1}{\underline{\lambda}} \quad \text{for $j \in [n+1]$}, \quad \text{and} \quad \big|[\nabla^2 \cL(\thetab; \bx)^{\dagger}]_{jk} \big|\lesssim \frac{1}{\underline{\lambda}} \quad \text{for $j \neq k$.}$$}
\end{corollary}

See Section \ref{sec: proof col btheta entry bound} for the proof of Corollary \ref{col: bTheta entry bound}. The following lemma provides a decomposition of the error of the debiased MLE $\hat\thetab^d$ into the leading term and the remainder term. 
\begin{lemma}\label{thm: mle decompose}
	Under the same conditions as Theorem \ref{thm: inf norm MLE} 
	, we have the following decomposition for the debiased MLE
	\begin{equation}\label{eq: decomp MLE}
		\hat\thetab^d - \thetab^{*} = - \nabla^2 \cL(\thetab^*; \bx)^{\dagger} \nabla \cL(\thetab^*; \bx)+ \Rb_0,
	\end{equation}
	where the residual term
	{ $$\|\Rb_0\|_{2} \lesssim \frac{\kappa_{\bcE}^2 n \sqrt{n} \log n}{\underline{\lambda} L}\left(n\sqrt{\frac{\kappa_{\bcE}^3 \log n}{\underline{\lambda}L}}+1\right),$$} with probability at least $1- O(n^{-10})$.
\end{lemma}

See Section \ref{sec: proof thm mle decomp} for the proof of Lemma \ref{thm: mle decompose}. Now we can begin with the proof. We abbreviate $\cL(\thetab; \bx)$ to $\cL(\thetab)$ for the convenience of notation in the following proof. For $k \in [n]$, we define the mapping $\vb_k(\cdot): \RR^{n+1} \rightarrow \RR^{n+1}$
$$\vb_k(\thetab) = \Big(\underbrace{(0-r_k) e^{\theta_0}, (r_1 - r_k)e^{\theta_1}, (r_2 - r_k)e^{\theta_2}, \ldots, (r_{k-1}- r_k)e^{\theta_{k-1}}}_{\text{first $k$ entries}} , 0, \ldots, 0\Big)^{\top},$$ 
and for the convenience of notation we let $\vb_k = \vb_k(\thetab^*)$. Define the function
$$g_k(\thetab) = \sum_{i=1}^k r_i \exp(\theta_i) - (\sum_{i=0}^k \exp(\theta_i))r_k,$$
and we have $\nabla g(\thetab) = \vb_k(\thetab)$ and $ \nabla^2 g(\thetab)= \operatorname{diag}\big(\vb_k(\thetab)\big)$, where $$\vb_k(\thetab) = \Big(\underbrace{(0-r_k) e^{\theta_0}, (r_1 - r_k)e^{\theta_1}, (r_2 - r_k)e^{\theta_2}, \ldots, (r_{k-1}- r_k)e^{\theta_{k-1}}}_{\text{first $k$ entries}} , 0, \ldots, 0\Big)^{\top}.$$
Then we have 
\begin{align*}
	\hat\Delta_k - \Delta_k& = g_k(\hat\thetab^d) - g_k(\thetab^*) = \nabla g(\thetab^*)^{\top} (\hat\thetab^d - \thetab^*) + \frac{1}{2}(\hat\thetab^d - \thetab^*)^{\top}\nabla^2 g(\tilde\thetab)(\hat\thetab^d - \thetab^*) \\
	& = \vb_k(\thetab^*)^{\top}(-\nabla^2 \cL(\thetab^*)^{\dagger} \nabla \cL(\thetab^*)+\Rb_0) + r_1, 
\end{align*}
where  $\|\tilde\thetab - \thetab^*\|_2 \le \|\hat\thetab^d - \thetab^*\|_2$, $\|\Rb_0\|_2 \lesssim { \frac{\kappa_{\bcE}^2 n \sqrt{n} \log n}{\underline{\lambda} L}\left(n\sqrt{\frac{\kappa_{\bcE}^3 \log n}{\underline{\lambda}L}}+1\right)}$ with probability at least $1-O(n^{-10})$ from Lemma \ref{thm: mle decompose}, and 
$$
|r_1| = \frac{1}{2}(\hat\thetab^d - \thetab^*)^{\top}\nabla^2 g(\tilde\thetab)(\hat\thetab^d - \thetab^*) \lesssim \|\vb_k\|_{\infty}\|\hat\thetab^d - \thetab^*\|_2^2.
$$
Note that from the proof of Lemma~\ref{thm: mle decompose} we can see that $\Rb_0^{\top} \mathbf{1} = {0}$, and hence $|\vb_k^{\top} \Rb_0| \le \|\Rb_0\|_2 \|\ponep \vb_k\|_2 $, where $\Pb_{\mathbf{1}}^{\perp}$ is the projection matrix to the perpendicular space of $\mathbf{1}$. Also from Lemma~\ref{lm: cck term bounds} and the proof of Lemma~\ref{thm: mle decompose}, we know that under the condition that { $ \underline{\lambda} L \gtrsim \kappa_{\bcE}^3 n^2 \log n$}, with high probability we have
{
$$
\|\hat\thetab^d - \thetab^*\|_2 \lesssim \frac{1}{\underline{\lambda}} \times  \sqrt{\frac{ n \bar\lambda\log n}{L}} + \frac{\kappa_{\bcE}^2 n \sqrt{n} \log n}{\underline{\lambda} L} \lesssim  \sqrt{\frac{ \kappa_{\bcE} n \log n}{\underline\lambda L}} ,
$$}
and thus we have { $|r_1| \lesssim \kappa_{\bcE} n \log n \|\vb_k\|_{\infty}/ (\underline{\lambda} L)$}. Now we consider the term $\vb_k^{\top}\nabla^2 \cL(\thetab^*)^{\dagger}  \nabla \cL(\thetab^*)$. We have 
{
\begin{align*}
	- \vb_k^{\top}\nabla^2 \cL(\thetab^*)^{\dagger}  \nabla \mathcal{L}\left(\thetab^{*}\right) &=\frac{1}{L}\vb_k^{\top}\nabla^2 \cL(\thetab^*)^{\dagger}  \sum_{\cS \in \bcE} \sum_{\ell=1}^{L_{\cS}}  \left\{ \sum_{i \in \cS_+} \left(x_{\cS}^{(i, \ell)} - \frac{e^{\theta_i^*}}{\sum_{j \in \cS_+} e^{\theta_j^*}}\right) \eb_i \right\} \\
	&=\frac{1}{L} \sum_{\cS \in \bcE} \sum_{\ell=1}^{L_{\cS}} y_{\cS}^{(\ell)},
\end{align*}}
where $y_{\cS}^{(\ell)} :=  \vb_k^{\top} \nabla^2 \cL(\thetab^*)^{\dagger}  \left\{ \sum_{i \in \cS_+} \left(x_{\cS}^{(i, \ell)} - \frac{e^{\theta_i^*}}{\sum_{j \in \cS_+} e^{\theta_j^*}}\right) \eb_i \right\} $ are independent for $\ell$ and $S$ given $\mathcal{E}$. Then we have
{
\begin{align*}
	\operatorname{Var}\left(\sum_{\cS \in \bcE} \sum_{\ell=1}^{L_{\cS}}y_{\cS}^{(\ell)}\right) =L\cdot \vb_k^{\top}\nabla^2 \cL(\thetab^*)^{\dagger}  \nabla^2 \cL(\thetab^*) \nabla^2 \cL(\thetab^*)^{\dagger}  \vb_k = L \cdot \vb_k^{\top}\nabla^2 \cL(\thetab^*)^{\dagger}  \vb_k \gtrsim  \frac{L\cdot\|\Pb_{\mathbf{1}}^{\perp} \vb_k \|_2^2}{\bar{\lambda} }.
\end{align*}}
Now consider $\left|y_{\cS}^{(\ell)}\right|$, and we have
{ 
\begin{align*}
	\left|y_{\cS}^{(\ell)}\right| & \lesssim \| \nabla^2 \cL(\thetab^*)^{\dagger} \vb_k\|_2 \left\| \sum_{i \in \cS_+} \left(x_{\cS}^{(i, \ell)} - \frac{e^{\theta_i^*}}{\sum_{j \in \cS_+} e^{\theta_j^*}}\right) \eb_i \right\|_2\lesssim  \| \nabla^2 \cL(\thetab^*)^{\dagger} \|_2\|\ponep \vb_k\|_{2} \lesssim \frac{1}{\underline{\lambda}}\|\ponep \vb_k\|_{2},
\end{align*}
}
where the last inequality is due to the fact that $\nabla^2 \cL(\thetab^*)^{\dagger}  \mathbf{1} = \mathbf{0}$. Then in turn we also have 
{
\begin{align*}
	&\sum_{\cS \in \bcE} \sum_{\ell=1}^{L_{\cS}} \EE \left| y_{\cS}^{(\ell)} \right|^3  \lesssim  \left(\max_{\cS \in \bcE} \max_{\ell \in [L_{\cS}]} \left| y_{\cS}^{(\ell)} \right|\right) \sum_{\cS \in \bcE} \sum_{\ell=1}^{L_{\cS}} \EE \left| y_{\cS}^{(\ell)} \right|^2  \lesssim \frac{L}{ \underline{\lambda}} \vb_k^{\top} \nabla^2 \cL(\thetab^*)^{\dagger}   \vb_k \|\ponep \vb_k\|_{2} \lesssim \frac{L}{ \underline{\lambda}^2} \cdot \|\ponep \vb_k\|_{2}^3.
\end{align*}}
Thus  under the condition that {  $\kappa_{\bcE}^{3}/{ (\underline{\lambda}L)}= o(1)$,} we have 
$$\max_k \left(\sum_{\cS \in \bcE} \sum_{\ell=1}^{L_{\cS}}  \operatorname{Var}\left(y_{\cS}^{(\ell)} \right)\right)^{-3 / 2}\left(\sum_{\cS \in \bcE} \sum_{\ell=1}^{L_{\cS}} \EE \left| y_{\cS}^{(\ell)} \right|^3 \right) = o(1).$$
Then by Lyapunov's Central Limit Theorem, we have 
$$\sqrt{L}(\vb_k^{\top} \nabla^2 \cL(\thetab^*)^{\dagger}  \vb_k)^{-1/2} \left(\vb_k^{\top}\nabla^2 \cL(\thetab^*)^{\dagger} \nabla\cL(\thetab^*)\right) \overset{d}{\rightarrow} N(0,1), \text{ for all } k \in [n].$$
Besides, under the condition that 
{
\begin{equation}\label{eq: condition thm asymp normal}
	\max_k \left\{ \kappa_{\bcE}^3{n}\log n\sqrt{\frac{\kappa_{\bcE} \cdot {n} }{{\underline{\lambda} L}}} + \frac{ \kappa_{\bcE}\log n \|\vb_k\|_{\infty}}{\|\Pb_{\mathbf{1}}^{\perp}\vb_k\|_2}\sqrt{\frac{\kappa_{\bcE} \cdot n^2}{\underline{\lambda} L}} \right\}= o(1),
\end{equation}}
we have $\sqrt{L}(\vb_k^{\top} \nabla^2 \cL(\thetab^*)^{\dagger}  \vb_k)^{-1/2} (\vb_k^{\top} \Rb_0 + r_1) = o_P(1)$, and by Slutsky's Theorem the claim follows. 

Now to simplify the conditions, we turn to study the norm $\|\Pb_{\mathbf{1}}^{\perp} \vb_k\|_2$ for $k \in [n]$. First note that $\Pb_{\mathbf{1}} \vb_k =(n+1)^{-1}\Delta_k \cdot \mathbf{1}$. Due to the non-decreasingness of $\Delta_k$, we consider the scenarios of $k \le K^*$ and $k > K^*$ separately. 
\begin{itemize}
	\item \textbf{$k \le K^*$}. From { Theorem~\ref{thm: alg}} we know that $\Delta_k \le 0$ when $k \le K^*$. Besides, since $(r_i - r_k)u_i^* \ge 0$ for $i < k$, we know that $|r_k u_0^*| \ge |\Delta_k|$.
	Thus we have that 
	\begin{align*}
		\|\ponep \vb_k\|_2^2 & = \left(-r_k u_0^* - \frac{\Delta_k}{n+1}\right)^2 + \sum_{i = 1}^{k-1} \left\{(r_{i} - r_k)u_i^* - \frac{\Delta_k}{n+1} \right\}^2 + \frac{n-k+1}{(n+1)^2} \Delta_k^2 \\
		& = (r_k u_0^*)^2 + \sum_{i = 1}^{k-1} \left((r_{i} - r_k)u_i^* \right)^2 - \frac{\Delta_k^2}{n+1}  \ge (r_k u_0^*)^2 - \frac{\Delta_k^2}{n+1} \\
		& \ge (r_k u_0^*)^2 - \frac{(r_k u_0^*)^2}{n+1} = \frac{n}{n+1}(r_k u_0^*)^2 \ge \frac{(r_k u_0^*)^2}{2}.
	\end{align*}
	\item \textbf{$k > K^*$}. When $k > K^*$, we know that $\Delta_K \ge 0$ from Theorem \ref{thm: alg}. Then we have 
	\begin{align*}
		\|\ponep \vb_k\|_2^2 & = \left(-r_k u_0^* - \frac{\Delta_k}{n+1}\right)^2 + \sum_{i = 1}^{k-1} \left\{(r_{i} - r_k)u_i^* - \frac{\Delta_k}{n+1} \right\}^2 + \frac{n-k+1}{(n+1)^2} \Delta_k^2 \\
		& \ge \left(-r_k u_0^* - \frac{\Delta_k}{n+1}\right)^2 \ge (r_n u_0^{*})^2.
	\end{align*}
\end{itemize}
Combining the above results we have that $\|\ponep \vb_k \|_2 \gtrsim r_n u_0^*$, and in turn we have that 
\begin{align*}
	\max_{k}\frac{\|\vb_k\|_{\infty}}{\|\ponep \vb_k \|_2} & \lesssim \frac{r_1 \max_i u_i^* }{r_n u_0^*} \le \kappa_{\thetab} \frac{r_1}{r_n} = \kappa_r \kappa_{\thetab} \lesssim \kappa_r.
\end{align*}
Therefore, under the condition that $\kappa_r \lesssim \sqrt{n}$, the condition \eqref{eq: condition thm asymp normal} simplifies to 
{
$$
\max_k \left\{ \kappa_{\bcE}^3{n}\log n\sqrt{\frac{\kappa_{\bcE} \cdot {n} }{{\underline{\lambda} L}}} + \kappa_r \kappa_{\bcE}\log n\sqrt{\frac{\kappa_{\bcE} \cdot  n^2}{\underline{\lambda} L}} \right\}  \lesssim  \kappa_{\bcE}^3{n}\log n\sqrt{\frac{\kappa_{\bcE} \cdot  {n} }{{\underline{\lambda} L}}} = o(1).
$$
}

\section{Proof of Corollary \ref{col: var est}}\label{sec: proof col var est}
By Slutsky's Theorem, it suffices for us to show that $$\max_k \left\{\lvert \hat\vb_k^{\top}  \nabla^2 \cL(\hat\thetab; \bx)^{\dagger} \hat\vb_k - \vb_k^{\top} \nabla^2 \cL(\thetab^*; \bx)^{\dagger}  \vb_k \rvert /\vb_k^{\top} \nabla^2 \cL(\thetab^*; \bx)^{\dagger}  \vb_k\right\}= o_P(1).$$ 
From \eqref{eq: cck term 5}, we know that $\| \nabla^2 \cL(\hat\thetab; \bx)^{\dagger} - \nabla^2 \cL(\thetab^*; \bx)^{\dagger} \|_2  \lesssim {  \frac{n}{\underline{\lambda}} \sqrt{\frac{\kappa_{\bcE}^3\log n}{\underline{\lambda} L}}}$ with high probability. Also from Theorem \ref{thm: inf norm MLE}, we know that $\|\hat\thetab - \thetab^{*}\|_2 \lesssim  { \sqrt{\frac{n \kappa_{\bcE} \log n}{\underline{\lambda}L}}}$, and in turn 
$$\|\hat\vb_k - \vb_k\|_2 = O_P\Big(\|\nabla^2 g(\thetab^{*}) (\hat\thetab- \thetab^{*})\|_2\Big) = O_P\Big( \|\operatorname{diag}(\vb_k)\|_2 \|\hat\thetab - \thetab^{*}\|_2\Big) = O_P\left( \|\vb_k\|_{\infty}{ \sqrt{\frac{n \kappa_{\bcE} \log n}{\underline{\lambda}L}}}
\right).$$ 
Combining the previous results, we have 
\begin{align*}
	&\lvert \hat\vb_k^{\top}  \nabla^2 \cL(\hat\thetab; \bx)^{\dagger} \hat\vb_k - \vb_k^{\top} \nabla^2 \cL(\thetab^*; \bx)^{\dagger}  \vb_k \rvert \le  \lvert\hat\vb_k^{\top}( \nabla^2 \cL(\hat\thetab; \bx)^{\dagger} - \nabla^2 \cL(\thetab^*; \bx)^{\dagger} ) \hat\vb_k\rvert \\
	&\quad + \lvert(\hat\vb_k - \vb_k)\nabla^2 \cL(\thetab^*; \bx)^{\dagger}  \vb_k\rvert + \lvert(\hat\vb_k - \vb_k)\nabla^2 \cL(\thetab^*; \bx)^{\dagger}  \hat\vb_k\rvert\\
	& \quad = O_P \Big(\| \nabla^2 \cL(\hat\thetab; \bx)^{\dagger} - \nabla^2 \cL(\thetab^*; \bx)^{\dagger} \|_2\|\Pb_{\mathbf{1}}^{\perp}\vb_k\|_2^2 + \|\Pb_{\mathbf{1}}^{\perp} \vb_k\|_2 \|\nabla^2 \cL(\thetab^*; \bx)^{\dagger}  \|_2\|\hat\vb_k - \vb_k\|_2\Big) \\
	& \quad \lesssim {  \frac{n \|\Pb_{\mathbf{1}}^{\perp} \vb_k\|_2^2}{\underline{\lambda}} \sqrt{\frac{\kappa_{\bcE}^3\log n}{\underline{\lambda} L}} + \frac{ \|\Pb_{\mathbf{1}}^{\perp} \vb_k\|_2\|\vb_k\|_{\infty}}{\underline{\lambda}} \sqrt{\frac{n \kappa_{\bcE} \log n}{\underline{\lambda}L}}}.
\end{align*}
{
Recall from the proof of Theorem \ref{thm: asymp normal} that $\vb_k^{\top} \nabla^2 \cL(\thetab^*; \bx)^{\dagger}  \vb_k \gtrsim \|\ponep \vb_k\|_2^2 / \bar{\lambda}$, then { under the condition that $\kappa_r \lesssim \sqrt{n}$ and ${n^2 \kappa_{\bcE}^5 \log n}/{(\underline{\lambda} L)} = o(1)$}, we have that with high probability
\begin{equation}\label{eq: variance estimate bound}
	\begin{aligned}
		& \max_k \left\{\frac{\lvert \hat\vb_k^{\top}  \nabla^2 \cL(\hat\thetab; \bx)^{\dagger} \hat\vb_k - \vb_k^{\top} \nabla^2 \cL(\thetab^*; \bx)^{\dagger}  \vb_k \rvert }{\vb_k^{\top} \nabla^2 \cL(\thetab^*; \bx)^{\dagger}  \vb_k } \right\} \\
		& \lesssim \max_k \left\{ \frac{n \|\Pb_{\mathbf{1}}^{\perp} \vb_k\|_2^2}{\underline{\lambda}} \sqrt{\frac{\kappa_{\bcE}^3\log n}{\underline{\lambda} L}} + \frac{ \|\Pb_{\mathbf{1}}^{\perp} \vb_k\|_2\|\vb_k\|_{\infty}}{\underline{\lambda}} \sqrt{\frac{n \kappa_{\bcE} \log n}{\underline{\lambda}L}}\right\}\!\!\frac{\bar{\lambda}}{\|\ponep \vb_k\|_2^2}\\
		&  = \max_k \kappa_{\bcE}\cdot \left\{ n \sqrt{\frac{\kappa_{\bcE}^3\log n}{\underline{\lambda} L}} + \frac{ \|\vb_k\|_{\infty}}{\|\Pb_{\mathbf{1}}^{\perp} \vb_k\|_2}\sqrt{\frac{n \kappa_{\bcE} \log n}{\underline{\lambda}L}} \right\} \lesssim n \kappa_{\bcE}^2\sqrt{\frac{\kappa_{\bcE}\log n}{\underline{\lambda} L}} + \kappa_r \kappa_{\bcE} \sqrt{\frac{n \kappa_{\bcE} \log n}{\underline{\lambda}L}}  \\
        & \lesssim n \kappa_{\bcE}^2 \sqrt{\frac{\kappa_{\bcE}\log n}{\underline{\lambda} L}} = o(1),
	\end{aligned}
\end{equation}}
and by Slutsky's theorem, the claim follows.

\section{Proof of Lemma \ref{lm: cw valid quantile cond on bcE}}\label{sec: proof thm legit ci k}
In the following proof, we denote $\cL(\thetab; \bx)$ by $\cL(\thetab)$ for the convenience of notation. When $\bcE$ is collected by a random scheme such that Assumption~\ref{asp: connectivity graph} holds with high probability (e.g., the setting of Lemma~\ref{lm: eigen of L cE}), establishing \eqref{eq: lem valid test} reduces to showing that, with high probability over $\bcE$, the following event holds. 
\begin{equation} \label{eq: valid quantile conditional}
	\bcA_{\bcE} := \left\{\sup_{\thetab^* \in \bTheta}\sup _{\alpha \in(0,1)}\left|\mathbb{P}_{\thetab^*}\left(T>c_{W}(\alpha, \bcE) \mid \bcE \right)-\alpha\right|=o(1) \right\}.
\end{equation}
In the remainder of the proof, we work in this more general setting where Assumption~\ref{asp: connectivity graph} holds with high probability with respect to the randomness of $\bcE$.
For the convenience of notation, for two positive sequences $x_n$ and $y_n$,  we use $x_n \lesssim y_n $ or $x_n = O(y_n)$ to imply that there exists a constant $C > 0$ independent of $\thetab^*$ such that $x_n \le C y_n$ for large enough $n$. 

{
For a given $\thetab^* \in \bTheta$, we define the following two auxiliary statistics,
\begin{equation}\label{eq: T0}
	T_0:=\max _{k \in [n]} \left\{\frac{\vb_k^{\top}\nabla^2 \cL(\thetab^*)^{\dagger} }{\sqrt{L \vb_k^{\top}\nabla^2 \cL(\thetab^*)^{\dagger} \vb_k}} \sum_{\cS \in \bcE}  \sum_{\ell=1}^{L_{\cS}}  \left\{ \sum_{i \in \cS_+} \left(x_{\cS}^{(i, \ell)} - \frac{e^{\theta_i^*}}{\sum_{j \in \cS_+} e^{\theta_j^*}}\right) \eb_i \right\}\right\},
\end{equation}
and 
\begin{equation}\label{eq: W0}
	W_0:=\max _{k \in [n]}  \sum_{\ell=1}^{L}\left\{\frac{\vb_k^{\top}\nabla^2 \cL(\thetab^*)^{\dagger} }{\sqrt{L \vb_k^{\top}\nabla^2 \cL(\thetab^*)^{\dagger} \vb_k}} \sum_{\cS \in \bcE}  \sum_{\ell=1}^{L_{\cS}} \left\{ \sum_{i \in \cS_+} \left(x_{\cS}^{(i, \ell)} - \frac{e^{\theta_i^*}}{\sum_{j \in \cS_+} e^{\theta_j^*}}\right) \eb_i \right\}z_{\cS, \ell}\right\} .
\end{equation}}
We define $y_{k}^{(\ell, \cS)} = \sqrt{\frac{1}{ \vb_k^{\top}\nabla^2 \cL(\thetab^*)^{\dagger} \vb_k}} \vb_k^{\top}\nabla^2 \cL(\thetab^*)^{\dagger} \left\{ \sum_{i \in \cS_+} \left(x_{\cS}^{(i, \ell)} - \frac{e^{\theta_i^*}}{\sum_{j \in \cS_+} e^{\theta_j^*}}\right) \eb_i \right\}$, for $\cS \in \bcE$, $\ell = 1, \ldots, L_{\cS}$ and $ k = 1, \ldots, n$. Then $W_0 = \max_{k \in [n]}\frac{1}{\sqrt{L}}  \sum_{\cS \in \bcE}\sum_{\ell=1}^{L_{\cS}} y_k^{(\ell, \cS)} \cdot z_{\cS,\ell}$. We will show that the following three conditions hold with high probability with respect to the randomness of $\bcE$ uniformly for all $\thetab^* \in \bTheta$.
\begin{enumerate}
	\item There exist $\zeta_1$ and $\zeta_2$ independent of $\thetab^*$ with $\zeta_{1} \sqrt{\log n}+\zeta_{2}=o(1)$  such that $\mathbb{P}\left(\lvert T-T_{0}\rvert>\zeta_{1}\right)<\zeta_{2}$, $\mathbb{P}\left(\mathbb{P}\left(\lvert W-W_{0}\rvert>\zeta_{1} \mid \boldsymbol{x}\right)>\zeta_{2}\right)<\zeta_{2}$,
	\item $c \leqslant \min_{k \in [n]} \frac{1}{L} \sum_{S \in \bcE}  \sum_{\ell=1}^{L_{\cS}} \mathbb{E}[(y_{k}^{(\ell, \cS)})^{2}]$ for some constant $c>0$,
	\item $\max_{k \in [n]}\lvert y_{k}^{(\ell, \cS)}\rvert \lesssim B$ with $B^{2}(\log (n L))^{7} / L =o(1)$ where $B$ is not necessarily a constant.
\end{enumerate}
We first verify condition (2). For any $k \in [n]$, since $y_{k}^{(\ell, \cS)}$'s are independent conditional on $\cE$, we have
\begin{align*}
	\frac{1}{L} \sum_{S \in \bcE} \sum_{\ell=1}^{L_{\cS}}  \mathbb{E}\big[(y_{k}^{(\ell, \cS)})^{2}\big] & = \frac{1}{{{L \vb_k^{\top}\nabla^2 \cL(\thetab^*)^{\dagger} \vb_k}}}\!\operatorname{Var}\!\left[{\vb_k^{\top}\nabla^2 \cL(\thetab^*)^{\dagger} } \sum_{\cS \in \bcE} \sum_{\ell=1}^{L_{\cS}} \! \sum_{i \in \cS_+}\!\! \left(\!\!x_{\cS}^{(i, \ell)} \!- \frac{e^{\theta_i^*}}{\sum_{j \in \cS_+} e^{\theta_j^*}}\right)\! \!\eb_i\!\right]\\ 
	& =\frac{\vb_k^{\top}\nabla^2 \cL(\thetab^*)^{\dagger} \vb_k}{\vb_k^{\top}\nabla^2 \cL(\thetab^*)^{\dagger} \vb_k} = 1,
\end{align*}
and thus condition (2) holds trivially. As for condition (3), with high probability with respect to the randomness of $\bcE$, we have that
{\begin{align*}
	\left| y_{k}^{(\ell, \cS)} \right| &= \sqrt{\frac{1}{ \vb_k^{\top}\nabla^2 \cL(\thetab^*)^{\dagger} \vb_k}} \left|\vb_k^{\top}\nabla^2 \cL(\thetab^*)^{\dagger} \left\{ \sum_{i \in \cS_+} \left(x_{\cS}^{(i, \ell)} - \frac{e^{\theta_i^*}}{\sum_{j \in \cS_+} e^{\theta_j^*}}\right) \eb_i \right\}\right| \\
	& = \sqrt{\frac{1}{ \vb_k^{\top}\nabla^2 \cL(\thetab^*)^{\dagger} \vb_k}} \left|(\Pb_{\mathbf{1}}^{\perp}\vb_k)^{\top}\nabla^2 \cL(\thetab^*)^{\dagger} \left\{ \sum_{i \in \cS_+} x_{\cS}^{(i, \ell)}\eb_i - \sum_{i \in \cS_+} \frac{e^{\theta_i^*}}{\sum_{j \in \cS_+} e^{\theta_j^*}} \eb_i \right\}\right|\\
	& \lesssim \frac{1}{\|\Pb_{\mathbf{1}}^{\perp} \vb_k\|_2/\sqrt{\bar\lambda}} \|\Pb_{\mathbf{1}}^{\perp} \vb_k\|_2 \left\|\nabla^2 \cL(\thetab^*)^{\dagger} \right\|_2 \left\| \sum_{i \in \cS_+} x_{\cS}^{(i, \ell)}\eb_i - \sum_{i \in \cS_+} \frac{e^{\theta_i^*}}{\sum_{j \in \cS_+} e^{\theta_j^*}} \eb_i \right\|_2 \\
	& \lesssim \sqrt{\kappa_{\bcE} / \underline{\lambda}}.
\end{align*}}
Thus we can take $B = \sqrt{\kappa_{\bcE} / \underline{\lambda}}$. Note that 
{ $$n \bar\lambda = n \|\Lb_{\bcE}\|_2 \ge n \cdot (\Lb_{\bcE})_{1,1} \gtrsim n \cdot \sum_{\cS \in \bcE} \frac{L_{\cS}}{L \cdot |\cS_+|} \gtrsim \sum_{\cS \in \bcE} \frac{L_{\cS}}{L  } = 1,$$}
where the second inequality is due to the fact that $0 \in \cS_+$ for any $\bcE$ by definition. 
Then condition (3) holds uniformly for $\thetab^* \in \bTheta$ with high probability regarding $\bcE$ under the condition that { $\kappa_{\bcE}^2 \log^7 n/ (\underline{\lambda} L ) = o(1)$.}

Now we move on to verify condition (1). First recall from previous proof that the following hold with probability at least $1- O(n^{-10})$,
{
\begin{align*}
	& \|\hat\thetab^d - \thetab^*\|_2 \lesssim  \sqrt{\frac{ \kappa_{\bcE} n \log n}{\underline\lambda L}} ,  \quad \|\hat\vb_k - \vb_k\|_2 \lesssim  \|\vb_k\|_{\infty} \sqrt{\frac{n \kappa_{\bcE} \log n}{\underline{\lambda}L}},\\
	& \|\nabla \cL(\hat\thetab ) - \nabla \cL(\thetab^*)\|_{\infty} \lesssim \sqrt{\frac{\kappa_{\bcE}^2 n \bar\lambda\log n}{ L}},\quad \| \nabla^2 \cL(\hat\thetab)^{\dagger} - \nabla^2 \cL(\thetab^*)^{\dagger} \|_2 \lesssim \frac{n}{\underline{\lambda}} \sqrt{\frac{\kappa_{\bcE}^3\log n}{\underline{\lambda} L}}, \\
	& \left|\hat\vb_k^{\top} \nabla^2 \cL(\hat\thetab)^{\dagger}\hat\vb_k - \vb_k^{\top}\nabla^2 \cL(\thetab^*)^{\dagger} \vb_k\right| \lesssim \frac{n \|\Pb_{\mathbf{1}}^{\perp} \vb_k\|_2^2}{\underline{\lambda}} \sqrt{\frac{\kappa_{\bcE}^3\log n}{\underline{\lambda} L}} + \frac{ \|\Pb_{\mathbf{1}}^{\perp} \vb_k\|_2\|\vb_k\|_{\infty}}{\underline{\lambda}} \sqrt{\frac{n \kappa_{\bcE} \log n}{\underline{\lambda}L}}.
\end{align*}}
And based on the above upper bounds we have 
{
\begin{align*}
	&\bigg\lvert \frac{1}{\sqrt{\hat\vb_k^{\top} \nabla^2 \cL(\hat\thetab)^{\dagger}\hat\vb_k}} \!-\! \frac{1}{\sqrt{\vb_k^{\top}\nabla^2 \cL(\thetab^*)^{\dagger}  \vb_k}} \bigg\rvert\! \lesssim\! \frac{1}{\sqrt{\vb_k^{\top}\nabla^2 \cL(\thetab^*)^{\dagger} \vb_k}}\!\bigg\lvert \!\frac{\hat\vb_k^{\top} \nabla^2 \cL(\hat\thetab)^{\dagger}\hat\vb_k \!- \!\vb_k^{\top}\nabla^2 \cL(\thetab^*)^{\dagger} \vb_k}{\vb_k^{\top}\nabla^2 \cL(\thetab^*)^{\dagger} \vb_k}\bigg\rvert\\
	& \quad \lesssim \frac{1}{\sqrt{\vb_k^{\top}\nabla^2 \cL(\thetab^*)^{\dagger} \vb_k}} \left( n \sqrt{\frac{\kappa_{\bcE}^3\log n}{\underline{\lambda} L}} + \frac{ \|\vb_k\|_{\infty}}{\|\Pb_{\mathbf{1}}^{\perp} \vb_k\|_2}\sqrt{\frac{n \kappa_{\bcE} \log n}{\underline{\lambda}L}}\right),\\
	& \|\hat\vb_k^{\top} \nabla^2 \cL(\hat\thetab)^{\dagger} - \vb_k^{\top}\nabla^2 \cL(\thetab^*)^{\dagger} \|_2 \lesssim \|(\hat\vb_k - \vb_k)^{\top}  \nabla^2 \cL(\hat\thetab)^{\dagger}\|_2+\|\vb_k^{\top}( \nabla^2 \cL(\hat\thetab)^{\dagger} - \nabla^2 \cL(\thetab^*)^{\dagger} )\|_2\\
	& \quad \lesssim \frac{ \|\vb_k\|_{\infty}}{\underline{\lambda}}\sqrt{\frac{n \kappa_{\bcE} \log n}{\underline{\lambda}L}} +   \frac{n \cdot \|\Pb_{\mathbf{1}}^{\perp}\vb_k\|_2}{\underline{\lambda}} \sqrt{\frac{\kappa_{\bcE}^3\log n}{\underline{\lambda} L}},
\end{align*}}
and in turn, we have 
{
\begin{align*}
	& \left\| \frac{\hat\vb_k^{\top} \nabla^2 \cL(\hat\thetab)^{\dagger}}{\sqrt{\hat\vb_k^{\top} \nabla^2 \cL(\hat\thetab)^{\dagger}\hat\vb_k}} - \frac{\vb_k^{\top}\nabla^2 \cL(\thetab^*)^{\dagger} }{\sqrt{\vb_k^{\top}\nabla^2 \cL(\thetab^*)^{\dagger}  \vb_k}} \right\|_2 \lesssim \frac{\|\hat\vb_k^{\top} \nabla^2 \cL(\hat\thetab)^{\dagger} - \vb_k^{\top}\nabla^2 \cL(\thetab^*)^{\dagger} \|_2}{\sqrt{\vb_k^{\top}\nabla^2 \cL(\thetab^*)^{\dagger} \vb_k}} \\
	& \quad + \bigg\lvert \frac{\|\vb_k^{\top}\nabla^2 \cL(\thetab^*)^{\dagger} \|_2}{\sqrt{\hat\vb_k^{\top} \nabla^2 \cL(\hat\thetab)^{\dagger}\hat\vb_k}} - \frac{\|\vb_k^{\top}\nabla^2 \cL(\thetab^*)^{\dagger} \|_2}{\sqrt{\vb_k^{\top}\nabla^2 \cL(\thetab^*)^{\dagger}  \vb_k}} \bigg\rvert \\
	& \quad \lesssim \frac{1}{\sqrt{\vb_k^{\top}\nabla^2 \cL(\thetab^*)^{\dagger} \vb_k}} \left(\frac{ \|\vb_k\|_{\infty}}{\underline{\lambda}}\sqrt{\frac{n \kappa_{\bcE} \log n}{\underline{\lambda}L}} +   \frac{n \cdot \|\Pb_{\mathbf{1}}^{\perp}\vb_k\|_2}{\underline{\lambda}} \sqrt{\frac{\kappa_{\bcE}^3\log n}{\underline{\lambda} L}}\right).
\end{align*}}

Now we consider 
{
\begin{align*}
	W-W_0 & \le \max _{k \in [n]}  \sum_{S \in \bcE} \sum_{\ell=1}^{L_{\cS}} \sum_{i \in \cS_+}\Bigg\{\frac{\hat\vb_k^{\top} \nabla^2 \cL(\hat\thetab)^{\dagger}}{\sqrt{L \hat\vb_k^{\top} \nabla^2 \cL(\hat\thetab)^{\dagger}\hat\vb_k}} \left(x_{\cS}^{(i, \ell)} - \frac{e^{\hat\theta_i}}{\sum_{j \in \cS_+} e^{\hat\theta_j}}\right)\\
	&\quad - \frac{\vb_k^{\top}\nabla^2 \cL(\thetab^*)^{\dagger} }{\sqrt{L \vb_k^{\top}\nabla^2 \cL(\thetab^*)^{\dagger} \vb_k}} \left(x_{\cS}^{(i, \ell)} - \frac{e^{\theta_i^*}}{\sum_{j \in \cS_+} e^{\theta_j^*}}\right)\Bigg\}\eb_i z_{\cS, \ell},
\end{align*}}
where the RHS is the maximum of $n$ centered Gaussian random variables conditioning on the data $\bx$ and the comparison graph $\cE$. We have the following upper bound on the conditional variance 
\begin{align*}
	& \max _{k \in [n]}  \frac{1}{L}\sum_{S \in \bcE}\! \sum_{\ell=1}^{L_{\cS}}\!\left\{\!\!\sum_{i \in \cS_+}\!\Bigg\{\frac{\hat\vb_k^{\top} \nabla^2 \cL(\hat\thetab)^{\dagger}}{\sqrt{ \hat\vb_k^{\top} \nabla^2 \cL(\hat\thetab)^{\dagger}\hat\vb_k}} \!\left(\!\!x_{\cS}^{(i, \ell)} \!- \!\frac{e^{\hat\theta_i}}{\sum_{j \in \cS_+} e^{\hat\theta_j}}\!\right)\!-\! \frac{\vb_k^{\top}\nabla^2 \cL(\thetab^*)^{\dagger} }{\sqrt{ \vb_k^{\top}\nabla^2 \cL(\thetab^*)^{\dagger} \vb_k}} \!\!\left(\!\!x_{\cS}^{(i, \ell)} - \frac{e^{\theta_i^*}}{\sum_{j \in \cS_+} e^{\theta_j^*}}\!\right)\!\!\!\Bigg\}\eb_i\!\right\}^2\\
	& \lesssim \underbrace{\max _{k \in [n]}  \frac{1}{L}\sum_{S \in \bcE}\! \sum_{\ell=1}^{L_{\cS}} \left\{\sum_{i \in \cS_+}\Bigg\{\frac{\vb_k^{\top}\nabla^2 \cL(\thetab^*)^{\dagger} }{\sqrt{ \vb_k^{\top}\nabla^2 \cL(\thetab^*)^{\dagger} \vb_k}} \left( \frac{e^{\hat\theta_i}}{\sum_{j \in \cS_+} e^{\hat\theta_j}} - \frac{e^{\theta_i^*}}{\sum_{j \in \cS_+} e^{\theta_j^*}}\right)\Bigg\}\eb_i \right\}^2}_{\text{I}}\\
	& \quad + \underbrace{\max _{k \in [n]}  \frac{1}{L}\sum_{S \in \bcE}\! \sum_{\ell=1}^{L_{\cS}}\left\{\sum_{i \in \cS_+}\Bigg\{\left(\frac{\hat\vb_k^{\top} \nabla^2 \cL(\hat\thetab)^{\dagger}}{\sqrt{ \hat\vb_k^{\top} \nabla^2 \cL(\hat\thetab)^{\dagger}\hat\vb_k}} - \frac{\vb_k^{\top}\nabla^2 \cL(\thetab^*)^{\dagger} }{\sqrt{ \vb_k^{\top}\nabla^2 \cL(\thetab^*)^{\dagger} \vb_k}}\right) \left(x_{\cS}^{(i, \ell)} - \frac{e^{\hat\theta_i}}{\sum_{j \in \cS_+} e^{\hat\theta_j}}\right)\Bigg\}\eb_i\right\}^2}_{\text{II}}.
\end{align*}
For term I, recall from previous proof that by the mean theorem, with probability at least $1 - O(n^{-10})$, for all $S \in \bcE$ and $i \in S$ we have that, 
$$
\left|  \frac{e^{\hat\theta_i}}{\sum_{j \in \cS_+} e^{\hat\theta_j}} - \frac{e^{\theta_i^*}}{\sum_{j \in \cS_+} e^{\theta_j^*}} \right| \lesssim \frac{e^{\theta_i^*}}{\sum_{j \in \cS_+} e^{\theta_j^*}} \|\hat\thetab - \thetab^*\|_{\infty} \lesssim \frac{1}{|\cS_+|} \|\hat\thetab - \thetab^*\|_{\infty},
$$
and thus with probability at least $1 - O(n^{-10})$, we have 
{
\begin{align*}
	\text{I} & = \max _{k \in [n]}  \frac{1}{L}\sum_{S \in \bcE}\! \sum_{\ell=1}^{L_{\cS}}\left\{ \sum_{i \in \cS_+}\Bigg\{\frac{\vb_k^{\top}\nabla^2 \cL(\thetab^*)^{\dagger} }{\sqrt{ \vb_k^{\top}\nabla^2 \cL(\thetab^*)^{\dagger} \vb_k}} \left( \frac{e^{\hat\theta_i}}{\sum_{j \in \cS_+} e^{\hat\theta_j}} - \frac{e^{\theta_i^*}}{\sum_{j \in \cS_+} e^{\theta_j^*}}\right)\Bigg\}\eb_i \right\}^2\\
	& \lesssim \max _{k \in [n]}  \frac{1}{L}\sum_{S \in \bcE}\! \sum_{\ell=1}^{L_{\cS}} \frac{1}{\vb_k^{\top}\nabla^2 \cL(\thetab^*)^{\dagger} \vb_k}\left\{(\Pb_{\mathbf{1}}^{\perp}\vb_k)^{\top} \sum_{i \in \cS_+}\Bigg\{ \left( \frac{e^{\hat\theta_i}}{\sum_{j \in \cS_+} e^{\hat\theta_j}} - \frac{e^{\theta_i^*}}{\sum_{j \in \cS_+} e^{\theta_j^*}}\right)\left[\nabla^2 \cL(\thetab^*)^{\dagger} \right]_i\Bigg\} \right\}^2\\
	& \lesssim \max _{k \in [n]}  \frac{1}{L}\sum_{S \in \bcE}\! \sum_{\ell=1}^{L_{\cS}}\frac{1}{\|\Pb_{\mathbf{1}}^{\perp}\vb_k\|_2^2/\bar\lambda} \|\Pb_{\mathbf{1}}^{\perp}\vb_k\|_2^2 \|\nabla^2 \cL(\thetab^*)^{\dagger} \|_2^2 \|\hat\thetab - \thetab^*\|_{\infty}^2  \\
	& \lesssim \frac{\kappa_{\bcE} \cdot \|\hat\thetab - \thetab^*\|_{\infty}^2}{\underline{\lambda}} \cdot n\bar\lambda \lesssim \frac{\kappa_{\bcE}^3 n^2 \log n}{\underline{\lambda}L},
\end{align*}}
where the last but one inequality follows from 
{ 
\begin{equation}\label{eq: sum L bound by lambda}
  \sum_{S \in \bcE} \frac{L_{\cS}}{L} \lesssim n \cdot \sum_{S \in \bcE} \frac{L_{\cS}}{L \cdot |\cS_+|} \lesssim n \cdot  (\Lb_{\bcE})_{1,1} \le n \cdot \|\Lb_{\bcE}\|_2 = n \bar\lambda ,
\end{equation}
}
and for term II we have 
{
\begin{align*}
	\text{II} & = \max _{k \in [n]}  \frac{1}{L}\sum_{S \in \bcE}\! \sum_{\ell=1}^{L_{\cS}} \left\{ \sum_{i \in \cS_+}\Bigg\{\left(\frac{\hat\vb_k^{\top} \nabla^2 \cL(\hat\thetab)^{\dagger}}{\sqrt{ \hat\vb_k^{\top} \nabla^2 \cL(\hat\thetab)^{\dagger}\hat\vb_k}} - \frac{\vb_k^{\top}\nabla^2 \cL(\thetab^*)^{\dagger} }{\sqrt{ \vb_k^{\top}\nabla^2 \cL(\thetab^*)^{\dagger} \vb_k}}\right) \left(x_{\cS}^{(i, \ell)} - \frac{e^{\hat\theta_i}}{\sum_{j \in \cS_+} e^{\hat\theta_j}}\right)\Bigg\}\eb_i \right\}^2\\
	& \le \max _{k \in [n]}  \frac{1}{L}\sum_{S \in \bcE}\! \sum_{\ell=1}^{L_{\cS}}  \left\|\frac{\hat\vb_k^{\top} \nabla^2 \cL(\hat\thetab)^{\dagger}}{\sqrt{ \hat\vb_k^{\top} \nabla^2 \cL(\hat\thetab)^{\dagger}\hat\vb_k}} - \frac{\vb_k^{\top}\nabla^2 \cL(\thetab^*)^{\dagger} }{\sqrt{ \vb_k^{\top}\nabla^2 \cL(\thetab^*)^{\dagger} \vb_k}}\right\|_2^2 \left\| \sum_{i \in \cS_+} \left(x_{\cS}^{(i, \ell)} - \frac{e^{\hat\theta_i}}{\sum_{j \in \cS_+} e^{\hat\theta_j}}\right)\eb_i \right\|_2^2\\
	& \lesssim  \max_{k \in [n]}\frac{n \bar\lambda}{{\vb_k^{\top}\nabla^2 \cL(\thetab^*)^{\dagger} \vb_k}} \left(\frac{ \|\vb_k\|_{\infty}}{\underline{\lambda}}\sqrt{\frac{n \kappa_{\bcE} \log n}{\underline{\lambda}L}} +   \frac{n \cdot \|\Pb_{\mathbf{1}}^{\perp}\vb_k\|_2}{\underline{\lambda}} \sqrt{\frac{\kappa_{\bcE}^3\log n}{\underline{\lambda} L}}\right)^2  \\
	& \lesssim n \max_{k \in [n]} \left(\frac{ \|\vb_k\|_{\infty}}{\|\Pb_{\mathbf{1}}^{\perp}\vb_k\|_2}\sqrt{\frac{n \kappa_{\bcE}^3 \log n}{\underline{\lambda}L}} +   n\kappa_{\bcE}\sqrt{\frac{\kappa_{\bcE}^3\log n}{\underline{\lambda} L}}\right)^2 \lesssim \frac{n^3 \kappa_{\bcE}^5 \log n}{\underline{\lambda} L},
\end{align*}
where the last but one inequality follows from \eqref{eq: sum L bound by lambda}.}
Denote { $\epsilon_0 = \frac{n^3 \kappa_{\bcE}^5 \log n}{\underline{\lambda} L}$}, and define the event $E_{\cE}$ to be 
{ 
\begin{align*}
	E_{\cE}  &= \Bigg\{ \max _{k \in [n]}  \frac{1}{L}\sum_{S \in \bcE}\! \sum_{\ell=1}^{L_{\cS}}\Bigg\{ \sum_{i \in \cS_+}\Bigg\{\frac{\hat\vb_k^{\top} \nabla^2 \cL(\hat\thetab)^{\dagger}}{\sqrt{ \hat\vb_k^{\top} \nabla^2 \cL(\hat\thetab)^{\dagger}\hat\vb_k}} \Big(x_{\cS}^{(i, \ell)} - \frac{e^{\hat\theta_i}}{\sum_{j \in \cS_+} e^{\hat\theta_j}} \Big)\\
	& \quad - \frac{\vb_k^{\top}\nabla^2 \cL(\thetab^*)^{\dagger} }{\sqrt{ \vb_k^{\top}\nabla^2 \cL(\thetab^*)^{\dagger} \vb_k}} \Big(x_{\cS}^{(i, \ell)} - \frac{e^{\theta_i^*}}{\sum_{j \in \cS_+} e^{\theta_j^*}}\!\Big)\!\!\!\Bigg\}\!\eb_i \Bigg\}^2 \le C_0 \epsilon_0 \Bigg\},
\end{align*}}
where the constant $C_0$ is chosen properly such that $\sup_{\thetab^* \in \bTheta}\PP_{\thetab^*}\left((E_{\cE})^{\rm c}\right) \le n^{-10}$. By maximal inequality we have that under the event $E_{\cE}$, there exists a large enough constant $C >0$ such that 
{
\begin{align*}
	& \PP_{\thetab^*}\left(W-W_0 \ge C\sqrt{\epsilon_0 \log n} \,|\,\bx, \cE \right) \le \PP_{\thetab^*}\Bigg(\max _{k \in [n]} \sum_{S \in \bcE}\! \sum_{\ell=1}^{L_{\cS}} \sum_{i \in \cS_+}\Bigg\{\frac{\hat\vb_k^{\top} \nabla^2 \cL(\hat\thetab)^{\dagger}}{\sqrt{L \hat\vb_k^{\top} \nabla^2 \cL(\hat\thetab)^{\dagger}\hat\vb_k}} \left(x_{\cS}^{(i, \ell)} - \frac{e^{\hat\theta_i}}{\sum_{j \in \cS_+} e^{\hat\theta_j}}\right)\\
	&\quad - \frac{\vb_k^{\top}\nabla^2 \cL(\thetab^*)^{\dagger} }{\sqrt{L \vb_k^{\top}\nabla^2 \cL(\thetab^*)^{\dagger} \vb_k}} \left(x_{\cS}^{(i, \ell)} - \frac{e^{\theta_i^*}}{\sum_{j \in \cS_+} e^{\theta_j^*}}\right)\Bigg\}\eb_i z_{\cS,\ell}\ge C\sqrt{\epsilon_0 \log n} \,|\,\bx, \cE \Bigg) \le n^{-10}/2.
\end{align*}}
Very similarly, under the event $\E_{\cE}$, we can also show that 
$$
\PP_{\thetab^*}\left(W_0-W \ge C\sqrt{\epsilon_0 \log n} \,|\,\bx, \cE \right) \le n^{-10}/2,
$$
and in turn 
$$
\PP_{\thetab^*}\left(|W-W_0| \ge C\sqrt{\epsilon_0 \log n} \,|\,\bx, \cE \right) \le n^{-10}.
$$
Combining the above results we have 
$$
\sup_{\thetab^* \in \bTheta}\PP_{\thetab^*}\left( \PP_{\thetab^*}\left(|W-W_0| \ge C\sqrt{\epsilon_0 \log n} \,|\,\bx, \cE \right) > n^{-10} \right) \le \sup_{\thetab^* \in \bTheta}\PP_{\thetab^*}\left((E_{\cE})^{\rm c}\right) \le n^{-10}.
$$

Now we consider $|T-T_0|$. Under the condition that $\kappa_r \lesssim \sqrt{n}$, with probability at least $1-O(n^{-10})$, we have that  
{
\begin{align*}
	|T-T_0| & \lesssim \max_{k \in [n]} \bigg\{\!\!(\hat\Delta_k \!-\! \Delta_k)\bigg\lvert \frac{\sqrt{L}}{\sqrt{\hat\vb_k^{\top} \nabla^2 \cL(\hat\thetab)^{\dagger}\hat\vb_k}} \!-\! \frac{\sqrt{L}}{\sqrt{\vb_k^{\top}\nabla^2 \cL(\thetab^*)^{\dagger}  \vb_k}} \bigg\rvert \!+\! \frac{\|\Pb_{\mathbf{1}}^{\perp}\vb_k\|_2\|\Rb_0 \|_2+ \|\vb_k\|_{\infty}\|\hat\thetab^d \!-\! \thetab^*\|_2^2}{\sqrt{\vb_k^{\top}\nabla^2 \cL(\thetab^*)^{\dagger} \vb_k/L}}\bigg\}\\
	& \lesssim  \max_{k \in [n]} \bigg\{\|\Pb_{\mathbf{1}}^{\perp}\vb_k\|_2\|\hat\thetab^d - \thetab^*\|_2\bigg\lvert \frac{\sqrt{L}}{\sqrt{\hat\vb_k^{\top} \nabla^2 \cL(\hat\thetab)^{\dagger}\hat\vb_k}} - \frac{\sqrt{L}}{\sqrt{\vb_k^{\top}\nabla^2 \cL(\thetab^*)^{\dagger}  \vb_k}} \bigg\rvert\\
	&\quad +\frac{ \|\Pb_{\mathbf{1}}^{\perp}\vb_k\|_2\|\Rb_0 \|_2+ \|\vb_k\|_{\infty}\|\hat\thetab^d - \thetab^*\|_2^2}{\sqrt{\vb_k^{\top}\nabla^2 \cL(\thetab^*)^{\dagger} \vb_k/L}}\bigg\}\\
	& \lesssim \max_{k \in [n]} \left\{ \sqrt{\frac{ \kappa_{\bcE}^2 n \log n}{ L}} \left( n \sqrt{\frac{\kappa_{\bcE}^3\log n}{\underline{\lambda} }} + \frac{ \|\vb_k\|_{\infty}}{\|\Pb_{\mathbf{1}}^{\perp} \vb_k\|_2}\sqrt{\frac{n \kappa_{\bcE} \log n}{\underline{\lambda}}}\right) + \kappa_{\bcE}^3{n}\log n\sqrt{\frac{\kappa_{\bcE} \cdot  {n} }{{\underline{\lambda} L}}} \right\}\\
    & \le \kappa_{\bcE}^3{n}\log n\sqrt{\frac{\kappa_{\bcE} \cdot  {n} }{{\underline{\lambda} L}}}.
\end{align*}}
Thus if we take { $\zeta_1 = C_1 \kappa_{\bcE}^3{n}\log n\sqrt{\frac{\kappa_{\bcE} \cdot  {n} }{{\underline{\lambda} L}}}$}
and $\zeta_2 = C_2 n^{-10}$ with large enough constants $C_1$ and $C_2$ independent of $\thetab^*$, under the condition that { $\kappa_{\bcE}^3{n}\log^{3/2} n\sqrt{\frac{\kappa_{\bcE} \cdot  {n} }{{\underline{\lambda} L}}} = o(1)$}, condition (1) also holds, and by Corollary 3.1 in \cite{cck2013aos} we have that with high probability with respect to $\bcE$ the event \eqref{eq: valid quantile conditional} holds.  

Then by Jensen's inequality and the dominated convergence theorem, we have 
\begin{align*}
	&\lim_{n, L \rightarrow \infty} \sup_{\thetab^* \in \bTheta}\sup _{\alpha \in(0,1)}\left|\mathbb{P}_{\thetab^*}\left(T>c_{W}(\alpha, \bcE) \right)-\alpha\right|  = \lim_{n, L \rightarrow \infty}  \sup_{\thetab^* \in \bTheta} \sup _{\alpha \in(0,1)}\left| \int \mathbb{P}_{\thetab^*}\left(T>c_{W}(\alpha, \bcE) \mid \bcE\right) {\rm d} \mu(\bcE)-\alpha\right| \\
	&  = \!\!\!\lim_{n, L \rightarrow \infty} \sup_{\thetab^* \in \bTheta} \!\sup _{\alpha \in(0,1)}\!\left| \int\!\! \left(\mathbb{P}_{\thetab^*}\!\left(T\!>\!c_{W}(\!\alpha, \!\bcE) \!\mid \!\bcE\right) - \alpha \right) {\rm d} \mu(\bcE)\right|\! \le \!\!\lim_{n, L \rightarrow \infty}  \!\int \!\!\!\sup_{\thetab^* \in \bTheta}\!\sup _{\alpha \in(0,1)}\!\!\left| \mathbb{P}_{\thetab^*}\!\left(T\!>\!c_{W}(\alpha, \bcE) \!\mid\! \bcE\right) \!-\! \alpha \right| \!{\rm d} \mu(\bcE)\\
	& = \!\!\!\lim_{n, L \rightarrow \infty}  \!\!\int_{\bcA_{\bcE}}\!\sup_{\thetab^* \in \bTheta}\!\sup _{\alpha \in(0,1)}\!\!\!\left| \mathbb{P}_{\thetab^*}\!\left(T\!>\!c_{W}(\alpha, \bcE)\! \mid\! \bcE\!\right) \!- \alpha \right| \!{\rm d} \mu(\bcE) \!+\!\!\! \lim_{n, L \rightarrow \infty}  \!\int_{\bcA_{\bcE}^c} \!\sup_{\thetab^* \in \bTheta}\!\sup _{\alpha \in(0,1)}\!\!\!\left| \mathbb{P}_{\thetab^*}\!\left(T\!>\!c_{W}(\alpha, \bcE) \!\mid \!\bcE\right) \!- \!\alpha \right| \!{\rm d} \mu(\bcE)\\
	& \le \int_{\bcA_{\bcE}} \lim_{n, L \rightarrow \infty} \sup_{\thetab^* \in \bTheta} \sup _{\alpha \in(0,1)}\left| \mathbb{P}_{\thetab^*}\left(T>c_{W}(\alpha, \bcE) \mid \bcE\right) - \alpha \right| {\rm d} \mu(\bcE) + \lim_{n,L\rightarrow \infty}\int_{\bcA_{\bcE}^c} {\rm d} \mu(\bcE) = 0.
\end{align*}
Thus $\sup_{\thetab^* \in \bTheta}\sup _{\alpha \in(0,1)}\left|\mathbb{P}_{\thetab^*}\left(T>c_{W}(\alpha, \bcE) \right)-\alpha\right|  = o(1)$ as $n, L \rightarrow \infty$.

\section{Proof of Theorem~\ref{thm: legit ci for K}}\label{sec: proof thm legit CI}
We will first show that $[\hat{K}_L, \hat{K}_U]$ is a valid confidence interval for $K^*$ based on Lemma~\ref{lm: cw valid quantile cond on bcE}. Note that by definition, if $\hat{K}_U < n$, we have that $ \hat\Delta_{\hat{K}_U + 1} > c_{W}(\alpha/2, \bcE)\sqrt{{\hat\vb_k^{\top} \nabla^2 \cL(\hat\thetab)^{\dagger} \hat\vb_k }/{L}}$.
Then we have 
\begin{align*}
	& \PP_{\thetab^*}(\hat{K}_{L} \!\!\le \!\!K^* \!\!\le \!\!\hat{K}_U) \ge 1\! -\! \PP_{\thetab^*}(\hat{K}_L \!>\! K^*) \!- \!\PP_{\thetab^*}(\hat{K}_U \!<\! K^*) = 1\! -\! \PP_{\thetab^*}(\hat{K}_L \!>\! K^*) \!-\! \PP_{\thetab^*}(\hat{K}_U\! +\! 1\! \le \!K^*) \\
	&= 1 \!-\! \PP_{\thetab^*}(\Delta_{\hat{K}_L}\! \ge\!0)  - \PP_{\thetab^*}\left(\left\{\Delta_{\hat{K}_U + 1} \!<\! 0\right\}\cap \{\hat{K}_U<n\}\right) \\
	&= 1\!-\! \PP_{\thetab^*}\left(\!\{\Delta_{\hat{K}_L}\! \ge \!0\} \!\cap \!\left\{\hat\Delta_{\hat{K}_L} \! \! < \!-c_{W}(\alpha/2, \bcE)\sqrt{{\hat\vb_{\hat{K}_L}^{\top} \nabla^2 \cL(\hat\thetab)^{\dagger} \hat\vb_{\hat{K}_L}}/{L} }\right\}\!\right)\\
	& \quad  - \PP_{\thetab^*}\left(\!\{\Delta_{\hat{K}_U + 1}\! < \!0\} \!\cap \!\left\{\hat{K}_U<n, \,\hat\Delta_{\hat{K}_U + 1} \! \! > \!c_{W}(\alpha/2, \bcE)\sqrt{{\hat\vb_{\hat{K}_U + 1}^{\top} \nabla^2 \cL(\hat\thetab)^{\dagger} \hat\vb_{\hat{K}_U + 1}}/{L} }\right\}\!\right) \\
	&  \ge 1 \!-\!  \PP_{\thetab^*}\Bigg( \!\min_{k > K^*} \frac{\hat\Delta_k - \Delta_k}{\sqrt{\hat\vb_k^{\top} \nabla^2 \cL(\hat\thetab)^{\dagger}\hat\vb_k/L}} \!\le\! -c_{W}(\alpha/2, \bcE)\!\!\Bigg) \!-\!  \PP_{\thetab^*}\Bigg(\!\max_{k \le K^*} \frac{\hat\Delta_k - \Delta_k}{\sqrt{\hat\vb_k^{\top} \nabla^2 \cL(\hat\thetab)^{\dagger}\hat\vb_k/L}} \ge c_{W}(\alpha/2, \bcE)\!\!\Bigg)\\
	&\ge 1 \!-\!  \PP_{\thetab^*}\Bigg( \!\min_{k \in [n]} \frac{\hat\Delta_k - \Delta_k}{\sqrt{\hat\vb_k^{\top} \nabla^2 \cL(\hat\thetab)^{\dagger}\hat\vb_k/L}} \!\le\! -c_{W}(\alpha/2, \bcE)\!\!\Bigg) \!-\!  \PP_{\thetab^*}\Bigg(\!\max_{k \in [n]} \frac{\hat\Delta_k - \Delta_k}{\sqrt{\hat\vb_k^{\top} \nabla^2 \cL(\hat\thetab)^{\dagger}\hat\vb_k/L}} \ge c_{W}(\alpha/2, \bcE)\!\!\Bigg).
\end{align*}
By Lemma~\ref{lm: cw valid quantile cond on bcE}, we know that 

$$
\!\lim_{n, L \rightarrow \infty} \sup_{\thetab^* \in \bTheta} \PP_{\thetab^*}\left(\max_{k \in [n]} \frac{\hat\Delta_k - \Delta_k}{\sqrt{\hat\vb_k^{\top} \nabla^2 \cL(\hat\thetab)^{\dagger}\hat\vb_k/L}} \ge c_{W}(\alpha/2, \bcE)\right) = \frac{\alpha}{2}.
$$
Also, following similar proof and the symmetry of Gaussian distribution we have
\begin{align*}
	&\lim_{n, L \rightarrow \infty} \sup_{\thetab^* \in \bTheta} \PP_{\thetab^*} \left( \min_{k \in [n]} \frac{\hat\Delta_k - \Delta_k}{\sqrt{\hat\vb_k^{\top} \nabla^2 \cL(\hat\thetab)^{\dagger}\hat\vb_k/L}}  \le -c_{W}(\alpha/2, \bcE) \right) \\
	& = \lim_{n, L \rightarrow \infty} \sup_{\thetab^* \in \bTheta} \PP_{\thetab^*} \left( \max_{k \in [n]} - \frac{\hat\Delta_k - \Delta_k}{\sqrt{\hat\vb_k^{\top} \nabla^2 \cL(\hat\thetab)^{\dagger}\hat\vb_k/L}} \ge c_{W}(\alpha/2, \bcE) \right) = \frac{\alpha}{2}.
\end{align*}
Thus by taking limits and the supremum on both sides of the inequality, we have that \eqref{eq: legit CI} follows.

Now we move on to show \eqref{eq: valid test}. By the equivalence of \eqref{eq: hypo test S S0} and \eqref{eq: hypo test K S0}, we have that
\begin{align*}
	&\sup_{\cS^* \in \bcS_0}\PP_{\thetab^*}(\text{Reject } H_0)  = \sup_{\thetab^*: K^* \in \cK_0}\PP_{\thetab^*}(\text{Reject } H_0) = \sup_{\thetab^*: K^* \in \cK_0}\PP_{\thetab^*}( [\hat{K}_L, \hat{K}_U] \cap \cK_0 = \emptyset) \\
	& = \sup_{\thetab^*: K^* \in \cK_0}\PP_{\thetab^*}\Big( \{K^* \in \cK_0\} \cap \big\{[\hat{K}_L, \hat{K}_U] \cap \cK_0 = \emptyset\big\}\Big) \le \sup_{\thetab^*: K^* \in \cK_0}\PP_{\thetab^*}\Big( \{K^* \in \cK_0\} \cap \{K^* \notin [\hat{K}_L, \hat{K}_U] \}\Big)  \\
	& \le 1 - \inf_{\thetab^*: K^* \in \cK_0}\PP_{\thetab^*}(K^* \in [\hat{K}_L, \hat{K}_U]) \le 1 - \inf_{\thetab^* \in \bTheta}\PP_{\thetab^*}(K^* \in [\hat{K}_L, \hat{K}_U]).
\end{align*}
Then from \eqref{eq: legit CI}, by taking limits on both sides we have that \eqref{eq: valid test} holds.
{ 
	\section{Proof of Theorem~\ref{thm: power and compare with recovery}}\label{sec: proof of thm power and compare with recovery}
	{ Similar as in the proof of Lemma~\ref{lm: cw valid quantile cond on bcE}, we work in this more general setting where $\bcE$ is allowed to be random and Assumption~\ref{asp: connectivity graph} holds with high probability with respect to the randomness of $\bcE$.}
    
	We will show the result in (1) first. We begin by establishing an upper bound for $T$ and $ c_{W}(\alpha/2, \bcE) $ with high probability. Conditional on \(\bcE\), define the matrix \(\bSigma_{\Delta} \in \RR^{n \times n}\) as follows:
	\[
	[\bSigma_{\Delta}]_{jk} = \frac{\vb_j^{\top}\nabla^2 \cL(\thetab^*)^{\dagger} \vb_k}{ \sqrt{ \left(\vb_j^{\top}\nabla^2 \cL(\thetab^*)^{\dagger} \vb_j \right) \cdot \left(\vb_k^{\top}\nabla^2 \cL(\thetab^*)^{\dagger} \vb_k \right) } }, \quad j,k \in [n],
	\]
	and let \(\boldsymbol{\xi} = (\xi_1,\ldots, \xi_n)^{\top} \in \RR^n\) be a multivariate Gaussian random vector conditional on \(\bcE\), such that \(\boldsymbol{\xi} \sim \cN(\mathbf{0}, \bSigma_{\Delta}) | \bcE\).
	
	By the verification of conditions (1)-(3) in the proof of Lemma~\ref{lm: cw valid quantile cond on bcE} and applying Theorem~2.2 in \citet{cck2013aos}, with probability at least \(1-o(1)\) with respect to the randomness of \(\bcE\), we have:
	\begin{equation}\label{eq: tv T W xi}
		\max\left\{ \sup_{t \in \RR}\big|\PP(T \le t | \bcE) - \PP(\max_{j \in [n]} \xi_j \le t  | \bcE)\big|,  \ \sup_{t \in \RR}\big|\PP(W \le t  | \bcE) - \PP(\max_{j \in [n]} \xi_j \le t  | \bcE)\big|\right\} \le \rho_n,
	\end{equation}
	where \(\rho_n = o(1)\) is a rate independent of \(\bcE\). Applying the maximal inequality for $\boldsymbol{\xi}$ and taking $t = 2\sqrt{\log n}$, we have that $\PP(\max_{j \in [n]} \xi_j \le 2\sqrt{\log n}  | \bcE)  \le n^{-1}$. Then applying the law of total probability and \eqref{eq: tv T W xi}, with probability at least $1 - o(1)$, we have that $\max(T, W) \le 2\sqrt{\log n}$, which also gives $ c_{W}(\alpha/2, \bcE) \le 2\sqrt{\log n}$ with high probability. 
	
	Next, we use the upper bounds for \(T\) and \(c_{W}(\alpha/2, \bcE)\) to construct an upper bound for the Type II error. For any $\thetab^* \in \bTheta$ such that $\cS^* \notin \bcS_0$ and $\delta(\thetab^*, \cK_0) \ge C \delta_n $, we have that
	\begin{equation}\label{eq: type II decomp}
		\PP_{\thetab^*} (\text{Reject } {\rm H}_0 ) = \PP_{\thetab^*} (\cK_0 \cap [\hat{K}_L, \hat{K}_U] = \emptyset) \ge \PP_{\thetab^*} \big(\big\{\cK_0^{\rm L} < \hat{K}_L \big\} \cap \big\{\cK_0^{\rm R} > \hat{K}_U \big\}\big).
	\end{equation}
	We first consider the event \(\{\cK_0^{\rm L} < \hat{K}_L \}\). Since \(\PP_{\thetab^*}(\cK_0^{\rm L} < \hat{K}_L) = 1\) when \(\cK_0^{\rm L} = -1\), we focus on the case where \(\cK_0^{\rm L} \geq 1\). By the definition of $\hat{K}_L$, if the following holds:
	$$
	\hat\Delta_{\cK_0^{\rm L} + 1} < -c_{W}\big(\frac{\alpha}{2}, \bcE\big)\sqrt{ \frac{\hat\vb_{\cK_0^{\rm L} + 1 }^{\top} \nabla^2 \cL(\hat\thetab)^{\dagger} \hat\vb_{\cK_0^{\rm L} + 1 }}{L} },
	$$
	then we have 
	$$
	\hat{K}_{L} =\max \Bigg\{ k: \hat\Delta_k < -c_{W}\big(\frac{\alpha}{2}, \bcE\big)\sqrt{ \frac{\hat\vb_k^{\top} \nabla^2 \cL(\hat\thetab)^{\dagger} \hat\vb_k}{L} } \Bigg\} \ge \cK_0^{\rm L} + 1 > \cK_0^{\rm L}.
	$$
	This indicates that
	\begin{align*}
		\PP(\cK_0^{\rm L} \ge \hat{K}_L) &\le \PP\left(\hat\Delta_{\cK_0^{\rm L} + 1} \ge -c_{W}\big(\frac{\alpha}{2}, \bcE\big)\sqrt{ \frac{\hat\vb_{\cK_0^{\rm L} + 1 }^{\top} \nabla^2 \cL(\hat\thetab)^{\dagger} \hat\vb_{\cK_0^{\rm L} + 1 }}{L} }\right)\\
		& = \PP\left(\Delta_{\cK_0^{\rm L} + 1} 
		+ \hat\Delta_{\cK_0^{\rm L} + 1}  -\Delta_{\cK_0^{\rm L} + 1} \ge -c_{W}\big(\frac{\alpha}{2}, \bcE\big)\sqrt{ \frac{\hat\vb_{\cK_0^{\rm L} + 1 }^{\top} \nabla^2 \cL(\hat\thetab)^{\dagger} \hat\vb_{\cK_0^{\rm L} + 1 }}{L} }\right)\\
		& \le \PP\left(\Delta_{\cK_0^{\rm L} + 1} 
		\ge -\Big(c_{W}\big(\frac{\alpha}{2}, \bcE\big) + T \Big)\sqrt{ \frac{\hat\vb_{\cK_0^{\rm L} + 1 }^{\top} \nabla^2 \cL(\hat\thetab)^{\dagger} \hat\vb_{\cK_0^{\rm L} + 1 }}{L} }\right) \\
		& \le \PP\left(\Delta_{\cK_0^{\rm L} + 1} 
		\ge -4\sqrt{\log n} \cdot \sqrt{ \frac{\hat\vb_{\cK_0^{\rm L} + 1 }^{\top} \nabla^2 \cL(\hat\thetab)^{\dagger} \hat\vb_{\cK_0^{\rm L} + 1 }}{L} }\right) + o(1)\\
		& \le \PP\left(\Delta_{\cK_0^{\rm L} + 1} 
		\ge -5 \sqrt{\log n} \cdot \sqrt{ \frac{\vb_{\cK_0^{\rm L} + 1 }^{\top} \nabla^2 \cL(\thetab^*)^{\dagger} \vb_{\cK_0^{\rm L} + 1 }}{L} }\right) + o(1),
	\end{align*}
	where the last inequality follows from \eqref{eq: variance estimate bound} under the condition that $\kappa_r \lesssim \sqrt{n}$ and { ${n^2 \kappa_{\bcE}^5 \log n}/{(\underline{\lambda} L)} = o(1)$}. Furthermore, when $\sup_{\thetab^* \in \bTheta}\kappa_{\thetab^*} = O(1)$, when Assumption~\ref{asp: connectivity graph} holds, we have that { $\sup_{\thetab^* \in \bTheta}\|\nabla^2 \cL(\thetab^*)^{\dagger} \|_2 \lesssim \underline{\lambda}^{-1}$} and $\|\vb_k\|_2^2 \asymp \sum_{j = 0}^{k-1} (r_j - r_k)^2$ for all $k \in [n]$. Hence there exists a properly large constant $C' > 0$ such that as long as $C > C' > 0$, we have 
	\begin{align*}
		\sup_{\thetab^*: \cS^* \notin \bcS_0, \delta(\thetab^*, \cK_0) \ge C \delta_n} \PP(\cK_0^{\rm L} \ge \hat{K}_L) &\le \PP\left(\Big[\sum_{j = 0}^{\cK_0^{\rm L} } (r_j - r_{\cK_0^{\rm L} + 1})^2\Big]^{-1/2}\Delta_{\cK_0^{\rm L} + 1} 
		\ge - C'  { \sqrt{ \frac{\log n}{\underline{\lambda}L } }}\right) + o(1) = o(1).
	\end{align*}
	Next, we consider the event  $\big\{\cK_0^{\rm R} > \hat{K}_U \big\} $. When $\cK_0^{\rm R} = n + 1$, we have $\PP_{\thetab^*}(\cK_0^{\rm R} > \hat{K}_U) = 1$, and we will focus on when $\cK_0^{\rm R} \le n$. By the definition of $\hat{K}_U$, if the following holds:
	$$
	\hat\Delta_{\cK_0^{\rm R} } > c_{W}\big(\frac{\alpha}{2}, \bcE\big)\sqrt{ \frac{\hat\vb_{\cK_0^{\rm R}  }^{\top} \nabla^2 \cL(\hat\thetab)^{\dagger} \hat\vb_{\cK_0^{\rm R} }}{L} },
	$$
	then we will have 
	$$
	\hat{K}_{U} =\min \Bigg\{ k: \hat\Delta_k > c_{W}\big(\frac{\alpha}{2}, \bcE\big)\sqrt{ \frac{\hat\vb_k^{\top} \nabla^2 \cL(\hat\thetab)^{\dagger} \hat\vb_k}{L} } \Bigg\} -1 \le  \cK_0^{\rm R} -1 < \cK_0^{\rm R},
	$$
	which in turn indicates that
	$$
	\PP(\cK_0^{\rm R} \le \hat{K}_U) \le \PP\left(\hat\Delta_{\cK_0^{\rm R} } \le c_{W}\big(\frac{\alpha}{2}, \bcE\big)\sqrt{ \frac{\hat\vb_{\cK_0^{\rm R}  }^{\top} \nabla^2 \cL(\hat\thetab)^{\dagger} \hat\vb_{\cK_0^{\rm R} }}{L} }\right).
	$$
	Then following similar arguments regarding the event $\big\{\cK_0^{\rm L} < \hat{K}_L \big\} $, we have that $ \sup_{\thetab^*: \cS^* \notin \bcS_0, \delta(\thetab^*, \cK_0) \ge C \delta_n} \PP(\cK_0^{\rm R} \le \hat{K}_U) = o(1)$. Combining with \eqref{eq: type II decomp}, we have that 
	$$
	\inf_{\thetab^*: \cS^* \notin \bcS_0, \delta(\thetab^*, \cK_0) \ge C \delta_n} \PP_{\thetab^*} (\text{Reject } {\rm H}_0 ) \rightarrow 1.
	$$
	
	Now we move on to show the results in (2). By Theorem~\ref{thm: alg}, the optimal assortment estimate $\hat\cS$ is of the form $\hat\cS = \{1,2,\ldots, \hat{K}\}$, where $\hat{K}= \max\{k \in [n]: \hat\Delta_k < 0\}$.  Also, it can be checked that $\hat\Delta_k$ is also non-decreasing with respect to $k \in [n]$. Hence we have the following equivalence
	\begin{align*}
		\PP(\hat\cS = \cS^*) =  \PP(\hat{K} = K^*) = \PP\big(\{\hat\Delta_{K^*} < 0\}\cap \{\hat\Delta_{K^* + 1} \ge 0\} \big).
	\end{align*}
	By Theorem~\ref{thm: asymp normal}, we have that 
	\begin{align*}
		\PP(\hat\Delta_{K^*} < 0) &=  \PP\left\{\sqrt{\frac{L}{\vb_{K^*}^{\top} \nabla^2 \cL(\thetab^*)^{\dagger} \vb_{K^*}}}\left(\hat\Delta_{K^*} - \Delta_{K^*} \right)< - \Delta_{K^*} \sqrt{\frac{L}{\vb_{K^*}^{\top} \nabla^2 \cL(\thetab^*)^{\dagger} \vb_{K^*}}}\right\}\\
		& = \Phi\left(- \Delta_{K^*} \sqrt{\frac{L}{\vb_{K^*}^{\top} \nabla^2 \cL(\thetab^*)^{\dagger} \vb_{K^*}}}\right) + o(1),\\
		\PP(\hat\Delta_{K^* + 1} \ge 0) &=  \PP\left\{\sqrt{\frac{L}{\vb_{K^* + 1}^{\top} \nabla^2 \cL(\thetab^*)^{\dagger} \vb_{K^* + 1}}}\left(\hat\Delta_{K^* + 1} - \Delta_{K^* + 1} \right) \ge - \Delta_{K^* + 1} \sqrt{\frac{L}{\vb_{K^* + 1}^{\top} \nabla^2 \cL(\thetab^*)^{\dagger} \vb_{K^* + 1}}}\right\}\\
		& = \Phi\left(\Delta_{K^* + 1} \sqrt{\frac{L}{\vb_{K^* + 1}^{\top} \nabla^2 \cL(\thetab^*)^{\dagger} \vb_{K^* + 1}}}\right) + o(1),
	\end{align*}
	where $\Phi(\cdot)$ denotes the cumulative distribution function of the standard normal distribution. Under Assumption~\ref{asp: connectivity graph}, for all $k \in [n]$, we have that
	{ \begin{equation}\label{eq: ponep vk ub}
		\bar\lambda^{-1} \|\ponep \vb_k\|_2^2  \lesssim \vb_{k}^{\top} \nabla^2 \cL(\thetab^*)^{\dagger} \vb_{k} \lesssim \underline{\lambda}^{-1} \sum_{j=0}^{k-1} (r_j - r_k)^2,
	\end{equation}}
	where $\Pb_{\mathbf{1}}^{\perp}$ is the projection matrix to the perpendicular space of $\mathbf{1}$. Furthermore, 
	\begin{equation}\label{eq: ponep vk lb}
		\begin{aligned}
			\|\ponep \vb_k\|_2^2 &= \| \vb_k\|_2^2 - \|\Pb_{\mathbf{1}} \vb_k\|_2^2 = \sum_{j=1}^k v_{kj}^2 - \frac{1}{n+1} (\sum_{j=1}^k v_{kj})^2\\
			& \ge \sum_{j=1}^k v_{kj}^2 - \frac{\underline{k}+1}{n+1} \Big(\sum_{j=1}^k v_{kj}^2\Big) \gtrsim \left(1-\frac{\underline{k}+1}{n+1}\right) \sum_{j=0}^{k-1} (r_j - r_k)^2,
		\end{aligned}
	\end{equation}
	where  $v_{kj}$ denotes the $j$-th entry of $\vb_k$, $\underline{k} := |\{j \in [n]: r_j > r_k\}|$ is the number of items with revenues larger than $r_k$, and the last but one inequality follows from Cauchy–Schwarz inequality. Combining \eqref{eq: ponep vk ub} and \eqref{eq: ponep vk lb} gives
	{
    $$
	\bar\lambda^{-1} \left(1-\frac{\underline{k}+1}{n+1}\right) \sum_{j=0}^{k-1} (r_j - r_k)^2 \lesssim \vb_{k}^{\top} \nabla^2 \cL(\thetab^*)^{\dagger} \vb_{k} \lesssim \underline{\lambda}^{-1} \sum_{j=0}^{k-1} (r_j - r_k)^2, \quad \text{for all } k \in [n].
	$$}
	Therefore, when $\min_{k \in \{K^*, K^*+1\}} [\sum_{j=0}^{k-1} (r_j - r_k)^2]^{-1/2} |\Delta_k| \ge C \delta_n $ for some large enough $C > 0$, by Chernoff bound, we have that 
	$$
	\PP(\hat\cS = \cS^*) \ge 1- \PP\big(\hat\Delta_{K^*} \ge 0 \big) - \PP\big(\hat\Delta_{K^* + 1} < 0 \big) \ge 1 - O(n^{-1}) - o(1) = 1 - o(1).
	$$
	On the other hand, if $\min_{k \in \{K^*, K^*+1\}} [\sum_{j=0}^{k-1} (r_j - r_k)^2]^{-1/2} |\Delta_k| \le c ({ \kappa_{\bcE}}\log n/[1-(\underline{K} + 2)/(n+1)])^{-1/2}\delta_n $ for some small enough constant $c > 0$, where $\underline{K} := |\{j \in [n]: r_j > r_{K^*}\}|$, we have that
	$$
	\PP(\hat\cS \ne \cS^*) \ge \max \{\PP\big(\hat\Delta_{K^*} \ge 0 \big) ,\, \PP\big(\hat\Delta_{K^* + 1} < 0 \big) \} \ge 1/3.
	$$
}
{
\section{Analogous Result to the Bradley--Terry--Luce (BTL) Model}\label{sec: lm eigen of L cE}
Lemma~\ref{lm: eigen of L cE} establishes conditions under which Assumption~\ref{asp: connectivity graph} holds when the sets \(\cS \in \bcE\) are sampled independently. We emphasize that this random-sampling assumption is not required, and we include it solely to facilitate comparison of optimal scaling rates with prior work.
\begin{lemma}\label{lm: eigen of L cE}
Let $\bcS=\{\cS\subseteq[n]:\cS\neq\emptyset\}$ denote the family of all nonempty offer sets. Suppose each $\cS\in\bcS$ is included in $\bcE$ independently with probability $p_{\cS}$ and yields sample size $L_{\cS}$. Let $\underline p:=\min_{\cS\in\bcS}p_{\cS}$ and assume there exists a constant $C\ge1$ such that
\begin{equation}\label{eq: avg p_S cond}
\max_{i\in[n]_+}
\frac{\sum_{\cS\in\bcS}\frac{p_{\cS}}{\underline p}\,|\cS_+|^{-1}\,\II\{i\in\cS_+\}}
     {\sum_{\cS\in\bcS}|\cS_+|^{-1}\,\II\{i\in\cS_+\}}
\;\le\; C,
\qquad
2^{n}\underline p\;\ge\; C\,n\log n.
\end{equation}
Define
\[
M_n=\sum_{k=2}^{n+1}\frac{1}{k^{2}}\binom{n-2}{k-3}\;\asymp\;\frac{2^{n}}{n^{2}},
\qquad
N_n=\sum_{k=2}^{n+1}\frac{1}{k^{2}}\binom{n-1}{k-2}\;\asymp\;\frac{2^{n+1}}{n^{2}}.
\]
Let $\underline L:=\min_{\cS\in\bcS}L_{\cS}$ and $\bar L:=\max_{\cS\in\bcS}L_{\cS}$. Then there exist absolute constants $C_0\ge c_0>0$ and $C_1>0$ such that, with probability at least $1-O(n^{-10})$,
\begin{equation}\label{eq: eigen of L cE}
c_0\,\frac{\underline L}{L}\,(nM_n+N_n)\,\underline p
\;\le\;
\underline\lambda \;\le\;
\bar\lambda
\;\le\;
C_0\,\frac{\bar L}{L}\,n N_n\,\underline p, \quad \kappa_{\bcE}\;\le\;C_1\,\frac{\bar L}{\underline L}.
\end{equation}
\end{lemma}
\begin{remark}
    The proof of Lemma \ref{lm: eigen of L cE} is deferred to Section~\ref{sec: proof lm eigen L cE}. The lower bound on $\underline p$ in \eqref{eq: avg p_S cond} guarantees that the comparison graph is connected with high probability. 
 To understand the heterogeneity condition in \eqref{eq: avg p_S cond}, consider the simple sufficient case where \(\max_{\cS \in \bcS} p_{\cS} \le C \underline{p}\) for some \(C \ge 1\), then the first inequality in \eqref{eq: avg p_S cond} holds trivially. Moreover, consider a more heterogeneous sufficient case where for each \(i \in [n]_+\), there exists a subset \(\tilde{\bcS}_i \subset \bcS_i\) with cardinality \(|\tilde{\bcS}_i| \le n\), and suppose that \(\max_{\cS \in \tilde{\bcS}_i} p_{\cS} / \underline{p} \le {2^n}/{n^2}\) and \(\max_{\cS \in \bcS_i \backslash \tilde{\bcS}_i} p_{\cS} / \underline{p} \le C'\) for some constant \(C' \ge 1\). Then, because \(\sum_{\cS \in \bcS_i} (|\cS| + 1)^{-1} \asymp {2^n}/{n}\) for all \(i \in [n]_+\), we have that
		\begin{align*}
			\max_{i \in [n]_+} \sum_{\cS \in \bcS_i} \frac{ (|\cS| + 1)^{-1} }{\sum_{\cS' \in \bcS_i} (|\cS'| + 1)^{-1}} \cdot \frac{p_{\cS}}{\underline{p}}  &\le \max_{i \in [n]_+} \frac{\sum_{\cS \in \tilde{\bcS}_i} (|\cS| + 1)^{-1} \cdot \frac{p_{\cS}}{\underline{p}} + \sum_{\cS \in \bcS_i \backslash \tilde{\bcS}_i} (|\cS| + 1)^{-1} \cdot \frac{p_{\cS}}{\underline{p}}}{\sum_{\cS \in \bcS_i} (|\cS| + 1)^{-1}} \\
			&\lesssim \frac{2^n}{n^2} \cdot \frac{n}{2^n / n} + C' \lesssim C' + 1,
		\end{align*}
		and \eqref{eq: avg p_S cond} holds. Hence, the heterogeneity condition in \eqref{eq: avg p_S cond} allows for \(p_{\cS}\) to differ by up to an exponential order of \(n\) for a small proportion of sets \(\cS \in \bcS_i\), which can be useful in practical scenarios where certain assortments are offered much more frequently than others.
\end{remark}
}
\section{Proof of Technical Lemmas}\label{sec: proof tech lems}
{
\subsection{Proof of Lemma~\ref{prop: rev order assort}}\label{sec: proof prop rev order assort}
The matrix $\Lb_{\bcE}$ can be rewritten as follows, 
\begin{align*}
    \Lb_{\bcE} = \sum_{k=2}^{n+1} \frac{m_{k-1}}{k^2} \diag \left(k \Ib_k - \mathbf{1}_k \mathbf{1}_k^{\top}, \mathbf{0}_{n+1-k}\right), 
\end{align*}
where $\mathbf{0}_m$ denotes the $m \times m$ zero matrix. It is immediate that $\Lb_{\bcE} \mathbf{1} = \mathbf{0}$ and $\lambda_{n+1} (\Lb_{\bcE}) = 0$. 

We show $\underline{\lambda} \asymp m_n / n$ first. Recall $\lambda_{\min, \perp}$ is the smallest eigenvalue restricted to the eigenvectors orthogonal to $\mathbf{1}$.
Then $\underline{\lambda} = \lambda_n (\Lb_{\bcE} ) = \lambda_{\min, \perp} (\Lb_{\bcE})$. For any $\vb \in \RR^{n+1}$ such that $\|\vb\|_2 = 1$ and $\vb^{\top} \mathbf{1} = 0$, the following holds
\begin{align*}
    \vb^{\top} \Lb_{\bcE} \vb \ge \frac{m_n}{(n+1)^2} \cdot \vb^{\top} \Big((n+1)\cdot \Ib_{n+1} - \mathbf{1}_{n+1} \mathbf{1}_{n+1}^{\top}\Big) \vb = \frac{m_n}{(n+1)^2} \cdot (n+1) \gtrsim m_n / n,
\end{align*}
which implies that $\underline{\lambda}  \gtrsim m_n / n$. On the other hand, taking $\vb_n = \eb_{n+1} - \frac{1}{n+1} \mathbf{1}_{n+1}$, we have that $\vb_n^{\top} \mathbf{1} = 0$ and that $\vb_n^{\top} \Lb_{\bcE} \vb_n = \eb_{n+1}^{\top} \Lb_{\bcE} \eb_{n+1}  = m_n / (n+1) $. Hence $\underline{\lambda}  = \lambda_{\min, \perp} (\Lb_{\bcE}) \lesssim \vb_n^{\top} \Lb_{\bcE} \vb_n  = m_n / (n+1) $, which in turn gives that $\underline{\lambda}   \asymp m_n / n$

 Next, we show that $\bar\lambda \asymp \sum_{k=1}^n {m_k}/{k}$. We have that the following holds
 $$\sum_{k=1}^n {m_k}/{k} \lesssim \eb_1^{\top} \Lb_{\bcE} \eb_1 \le \bar\lambda  = \lambda_1 (\Lb_{\bcE}) \le \sum_{k=2}^{n+1} \frac{m_{k-1}}{k^2} \|k \Ib_k - \mathbf{1}_k \mathbf{1}_k^{\top}\|_2  \lesssim \sum_{k=1}^n {m_k}/{k},$$
 which gives $\bar\lambda  \asymp \sum_{k=1}^n {m_k}/{k}$. 

 When $\sum_{k=1}^{n-1} m_k / k \lesssim n_n/ n$, $\bar\lambda \asymp \underline\lambda$ follows trivially, and the proof is complete. 
 }
{ \subsection{Proof of Lemma \ref{lm: eigen of L cE}}\label{sec: proof lm eigen L cE}
Let $\Ab_{\cS} = \cE_{\cS} \cdot \frac{1}{|\cS_+|^2} \sum_{\substack{i,j\in \cS_+\\i < j}} (\eb_i - \eb_j)(\eb_i - \eb_j)^{\top}$, where $\cE_{\cS} = \II\{\cS \in \bcE\}$, then we have that
\begin{equation}\label{eq: L eps range}
 \frac{\underline{L}}{L} \sum_{\cS \in \bcS} \Ab_{\cS}   \preceq \Lb_{\bcE} \preceq \frac{\bar{L}}{L} \sum_{\cS \in \bcS} \Ab_{\cS}.
\end{equation}
Denote $\widetilde\Lb_{\bcE} = \sum_{\cS \in \bcS} \Ab_{\cS}$. Recall $M_n = \sum_{k=2}^{n+1} \frac{1}{k^2} {n - 2 \choose k - 3} $ and $N_n = \sum_{k=2}^{n+1} \frac{1}{k^2} {n - 1 \choose k - 2} $, and we have that 
\begin{align*}
	\EE(\widetilde\Lb_{\bcE}) & = \underline{p}  \sum_{\cS \in \bcS} \frac{p_{\cS}}{\underline{p}} \cdot \frac{1}{|\cS_+|^2} \sum_{\substack{i,j\in \cS_+ \\i < j}} (\eb_i - \eb_j)(\eb_i - \eb_j)^{\top} 
	\succeq \underline{p} \begin{pmatrix}
		n N_n & \quad  - N_n \mathbf{1}_n^{\top} \\
		- N_n \mathbf{1}_n & \quad  (nM_n\! +\! N_n) \Ib_n \!- \!M_n \mathbf{1}_n \mathbf{1}_n^{\top}
	\end{pmatrix} \\
	& = \underline{p} \left( (n+1)M_n \Ib_{n+1} - M_n \mathbf{1}_{n+1} \mathbf{1}_{n+1}^{\top} + (N_n - M_n)\begin{pmatrix}
		n & -\mathbf{1}_n^{\top}\\
		-\mathbf{1}_n & \Ib_n
	\end{pmatrix}\right)\\
	& = \underline{p} \left( (n+1)M_n \Ib_{n+1} - M_n \mathbf{1}_{n+1} \mathbf{1}_{n+1}^{\top} + (N_n - M_n)\left(\Ib_{n+1} + \diag(n, \mathbf{1}_n\mathbf{1}_n^{\top}) - \mathbf{1}_{n+1} \mathbf{1}_{n+1}^{\top}\right)\right)\notag\\
	& = \underline{p} \left( (nM_n + N_n)\Ib_{n+1} + (N_n - M_n)\diag(n, \mathbf{1}_n\mathbf{1}_n^{\top})  - N_n \mathbf{1}_{n+1} \mathbf{1}_{n+1}^{\top}  \right).
\end{align*}
Consider the matrix $(N_n - M_n)\diag(n, \mathbf{1}_n\mathbf{1}_n^{\top})  - N_n \mathbf{1}_{n+1} \mathbf{1}_{n+1}^{\top} $. Define $\wb_1 = (1, \mathbf{0}^{\top})^{\top}$, $\wb_2 = (0,\mathbf{1}_n^{\top}/\sqrt{n})^{\top}$ and $\Wb = (\wb_1, \wb_2)$. Then it can be seen that $\mathbf{1}_{n+1} = \wb_1 + \sqrt{n} \wb_2$ and we have that
\begin{align*}
	& (N_n - M_n)\diag(n, \mathbf{1}_n\mathbf{1}_n^{\top})  - N_n \mathbf{1}_{n+1} \mathbf{1}_{n+1}^{\top} 
	= n(N_n - M_n)\Wb \Wb^{\top} - \Wb \begin{pmatrix}
		N_n & \sqrt{n} N_n\\
		\sqrt{n} N_n & N_n
	\end{pmatrix} \Wb^{\top}\\
	& = \Wb \begin{pmatrix}
		(n-1)N_n - nM_n & -\sqrt{n}N_n\\
		-\sqrt{n} N_n & -n M_n
	\end{pmatrix} \Wb^{\top} = \Wb \Gb \diag \left(n(N_n - M_n) , -nM_n - N_n\right) \Gb^{\top} \Wb^{\top},
\end{align*}
where $\Gb = \frac{1}{\sqrt{n+1}}\begin{pmatrix}
	\sqrt{n} & 1\\
	-1 & \sqrt{n}
\end{pmatrix}$ and the last equality follows from the following eigendecomposition:
$$\begin{pmatrix}
	(n-1)N_n - nM_n & -\sqrt{n}N_n\\
	-\sqrt{n} N_n & -n M_n
\end{pmatrix} = \Gb \diag \left(n(N_n - M_n) , -nM_n - N_n\right) \Gb^{\top}.$$
Then combining the above results we have that 
\begin{equation}\label{eq: spike of E L cE}
	\EE(\widetilde\Lb_{\bcE}) \succeq \underline{p} \left( (nM_n + N_n)\Ib_{n+1} + \Wb \Gb \diag \left(n(N_n - M_n) , -nM_n - N_n\right) \Gb^{\top} \Wb^{\top}\right),
\end{equation}
which indicates that 
$\lambda_{n} \big(\EE(\widetilde\Lb_{\bcE}) \big) \ge (nM_n + N_n)\underline{p}$.
Also since $ \EE(\widetilde\Lb_{\bcE})\mathbf{1}_{n+1} = \mathbf{0}$, we have that $\mathbf{1}_{n+1}$ corresponds to the $(n+1)$-th eigenvalue of $\EE(\widetilde\Lb_{\bcE})$. 
Moreover, we have the following upper bound holds,
\begin{equation}\label{eq: L eps upper bd}
    \begin{aligned}
			\EE(\widetilde\Lb_{\bcE})  &= \underline{p}  \sum_{\cS \in \bcS} \frac{p_{\cS}}{\underline{p}} \cdot \frac{1}{|\cS_+|^2} \sum_{\substack{i,j\in \cS_+\\i < j}} (\eb_i - \eb_j)(\eb_i - \eb_j)^{\top} \\
			& \preceq \underline{p} \sum_{S \in \bcS} \frac{p_{\cS}}{\underline{p}} \cdot |\cS_+|^{-1} \Ib_{\cS_+} = \underline{p} \cdot \diag( s_0, s_1, \ldots, s_n),
		\end{aligned}
\end{equation}
        where $\Ib_{\cS_+}$ is a diagonal matrix with $(\Ib_{\cS_+})_{ii} = 1$ for $i \in \cS_+$ and  $(\Ib_{\cS_+})_{ii} = 0$ for $i \notin \cS_+$, and $s_i = \sum_{\cS \in \bcS} \frac{p_{\cS}}{\underline{p}}\cdot |\cS_+|^{-1} \II\{i \in \cS_+\}$ for $i \in [n]_+$; the last inequality follows from the fact that for any $\cS \in \bcS$ we have
		$$
		\sum_{\substack{i,j\in \cS_+\\i < j}} (\eb_i - \eb_j)(\eb_i - \eb_j)^{\top}  = |\cS_+| \cdot \Ib_{\cS_+} - \mathbf{1}_{\cS_+} \mathbf{1}_{\cS_+}^{\top} \preceq |\cS_+| \cdot \Ib_{\cS_+},
		$$
		where $\mathbf{1}_{\cS_+} \in \RR^{n+1}$ with $(\mathbf{1}_{\cS_+})_i = 1$ for $i \in \cS_+$ and $(\mathbf{1}_{\cS_+})_i = 0$ for $i \notin \cS_+$. When the condition \eqref{eq: avg p_S cond} holds, we have that $s_i \le C \sum_{\cS \in \bcS}  |\cS_+|^{-1} \II\{i \in \cS_+\} \lesssim (n+1) N_n \asymp 2^n / n$ for any $i \in [n]_+$, and in turn we have $\lambda_{1} \big(\EE(\widetilde\Lb_{\bcE}) \big) \lesssim (n+1)N_n \underline{p}$. Hence
        $$
        (nM_n + N_n)\underline{p} \le \lambda_{n} \big(\EE(\widetilde\Lb_{\bcE}) \big) \le \lambda_{1} \big(\EE(\widetilde\Lb_{\bcE}) \big) \lesssim (n+1)N_n \underline{p}.
        $$

Let $\Rb_{\bcE} \in \RR^{(n+1) \times n}$ denote the stacking of the top $n$ normalized eigenvectors of $\EE(\widetilde\Lb_{\bcE})$, which is unique up to rotation, and we have $\Rb_{\bcE}^{\top} \Rb_{\bcE} = \Ib_n$ and $\Rb_{\bcE}^{\top} \mathbf{1}_{n+1} = \mathbf{0}$. Then from \eqref{eq: spike of E L cE} and \eqref{eq: L eps upper bd}, it can be seen that 
$$
\diag \big((n+1)N_n\underline{p}, (nM_n + N_n)\underline{p}\Ib_{n-1}\big) \preceq \EE(\Rb_{\bcE}^{\top} \widetilde\Lb_{\bcE} \Rb_{\bcE}) =\Rb_{\bcE}^{\top} \EE(\widetilde\Lb_{\bcE}) \Rb_{\bcE} \preceq C' nN_n\underline{p}\cdot \Ib_n,
$$
where $C' > 0$ is a large enough constant. We have that 
$$\lambda_{\min} (\Ab_{\cS}) \ge 0, \quad \lambda_{\max}(\Ab_{\cS} ) \le |\cS_+|/|\cS_+|^2 = 1/|\cS_+| \le 1,$$ 
and 
$$\lambda_{\min} (\Rb_{\bcE}^{\top} \Ab_{\cS} \Rb_{\bcE}) \ge 0, \quad \lambda_{\max}(\Rb_{\bcE}^{\top}\Ab_{\cS} \Rb_{\bcE}) \le \lambda_{\max}(\Ab_{\cS} )  \le 1.$$ Then by Theorem 5.1.1 in \cite{tropp2015intromatineq}, under the condition that $2^n \underline{p} \ge C n \log n$ for some large enough constant $C > 0 $, there exists a constant $c>0$ such that for large enough $n$ we have
\begin{align*}
	&\PP\left(\lambda_{\min, \perp}(\widetilde\Lb_{\bcE}) \le  (nM_n + N_n)\underline{p} /2 \right) \le \PP\left(\lambda_{\min}(\Rb_{\bcE}^{\top}\widetilde\Lb_{\bcE}\Rb_{\bcE}) \le \lambda_{\min}\left(\EE(\Rb_{\bcE}^{\top} \widetilde\Lb_{\bcE} \Rb_{\bcE}) \right) /2 \right)\\
	& \quad \le n\exp(-cnM_n \underline{p} ) \le n^{-10}, 
\end{align*}
and 
\begin{align*}
	&\PP\left(\lambda_{\max}(\widetilde\Lb_{\bcE}) \ge  3C' n N_n \underline{p} /2 \right) \le \PP\left(\lambda_{\max}(\widetilde\Lb_{\bcE}) \ge 3\lambda_{\max}\left(\EE( \widetilde\Lb_{\bcE} ) \right) /2 \right)\\
	& \quad \le (n+1)\exp(-cnM_n \underline{p} ) \le n^{-10}.
\end{align*}
Then we know that \eqref{eq: eigen of L cE} holds with probability at least $1 - O(n^{-10})$ as the result of \eqref{eq: L eps range}.}
\subsection{Proof of Lemma \ref{lm: eigen hessian}}\label{sec: proof lm eigen hessian}

{
First, we can rewrite 
$$
\nabla^2 \cL(\thetab; \bx) = \sum_{\cS \in \bcE} \frac{L_{\cS}}{L}\left\{ \sum_{\substack{i,l\in \cS_+\\i < l}} \frac{e^{\theta_i}e^{\theta_l}}{\left(\sum_{j \in \cS_+} e^{\theta_j}\right)^2}(\eb_i - \eb_l) (\eb_i - \eb_l)^{\top} \right\}.
$$
Then we have
$$
\frac{e^{\theta_i}}{\sum_{j \in \cS_+} e^{\theta_j}}  = \frac{1}{\sum_{j \in \cS_+} e^{\theta_j - \theta_i}} = \frac{1}{\sum_{j \in \cS_+} e^{(\theta_j - \theta_j^*) + (\theta_j^* - \theta_i^*) + (\theta_i^*- \theta_i)}} \ge \frac{1}{\sum_{j \in \cS_+} e^{\log \kappa_{\thetab} + 2\|\thetab^*- \thetab\|_{\infty}}} = \frac{1}{|\cS_+|\kappa_{\thetab} e^{2C}}.
$$
Similarly, we also have 
$$
\frac{e^{\theta_i}}{\sum_{j \in \cS_+} e^{\theta_j}} = \frac{1}{\sum_{j \in \cS_+} e^{(\theta_j - \theta_j^*) + (\theta_j^* - \theta_i^*) + (\theta_i^*- \theta_i)}} \le \frac{1}{\sum_{j \in \cS_+} e^{-\log \kappa_{\thetab} - 2\|\thetab^*- \thetab\|_{\infty}}} = \frac{\kappa_{\thetab} e^{2C}}{|\cS_+|}.
$$
Since $(\eb_i - \eb_l) (\eb_i - \eb_l)^{\top} \succeq \mathbf{0}$ for any $i < l$, we have that 
$$
\left(\frac{e^{\theta_i}e^{\theta_l}}{\left(\sum_{j \in \cS_+} e^{\theta_j}\right)^2} - \frac{1}{|\cS_+|^2 (\kappa_{\thetab} e^{2C})^2}\right)(\eb_i - \eb_l) (\eb_i - \eb_l)^{\top} \succeq \mathbf{0},
$$
$$
\text{and} \quad \left(\frac{e^{\theta_i}e^{\theta_l}}{\left(\sum_{j \in \cS_+} e^{\theta_j}\right)^2} - \frac{(\kappa_{\thetab} e^{2C})^2}{|\cS_+|^2 }\right)(\eb_i - \eb_l) (\eb_i - \eb_l)^{\top} \preceq \mathbf{0}.
$$
Hence,
\begin{align*}
	\nabla^2 \cL(\thetab; \bx) & = \sum_{\cS \in \bcE} \frac{L_{\cS}}{L}\bigg\{ \sum_{\substack{i,l\in \cS_+\\i < l}} \frac{e^{\theta_i}e^{\theta_l}}{\left(\sum_{j \in \cS_+} e^{\theta_j}\right)^2}(\eb_i - \eb_l) (\eb_i - \eb_l)^{\top} \bigg\} \\
	& \succeq\sum_{\cS \in \bcE} \frac{L_{\cS}}{L}\bigg\{ \sum_{\substack{i,l\in \cS_+\\i < l}} \frac{1}{|\cS_+|^2 (\kappa_{\thetab} e^{2C})^2} (\eb_i - \eb_l) (\eb_i - \eb_l)^{\top} \bigg\}\\
	& = \frac{1}{(\kappa_{\thetab} e^{2C})^2}\sum_{\cS \in \bcE} \frac{L_{\cS}}{L} \bigg\{ \sum_{\substack{i,l\in \cS_+\\i < l}} \frac{1}{|\cS_+|^2} (\eb_i - \eb_l) (\eb_i - \eb_l)^{\top} \bigg\} = \frac{1}{(\kappa_{\thetab} e^{2C})^2} \Lb_{\bcE},
\end{align*}
and similarly $\nabla^2 \cL(\thetab; \bx)  \preceq (\kappa_{\thetab} e^{2C})^2 \Lb_{\bcE}$. 
Then by the fact that $\nabla^2 \cL_{\tau}(\thetab; \bx)  = \nabla^2 \cL(\thetab; \bx) + \tau \Ib_{n+1}$,
the inequalities in \eqref{eq: eigen hessian} follow.}
\subsection{Proof of Lemma \ref{lm: bound gradient}}\label{sec: proof lm bound gradient}
{ 
We know that 
\begin{align*}
	\nabla \cL_{\tau} (\thetab^*; \bx) &= \tau \thetab^* - \sum_{\cS \in \bcE} \left\{ \sum_{i \in \cS_+} \left(x_{\cS}^{(i)} - \frac{L_{\cS}}{L} \frac{e^{\theta^*_i}}{\sum_{j \in \cS_+} e^{\theta^*_j}}\right) \eb_i \right\}\\
	& =  \tau \thetab^* - \frac{1}{L}\sum_{\cS \in \bcE}\sum_{\ell = 1}^{L_{\cS}} \underbrace{\left\{ \sum_{i \in \cS_+}\left(x_{\cS}^{(i,\ell)} - \frac{e^{\theta^*_i}}{\sum_{j \in \cS_+} e^{\theta^*_j}}\right) \eb_i \right\}}_{:= \bz_{\cS}^{(\ell)}},
\end{align*}
and we have $\EE\left(\bz_{\cS}^{(\ell)}\right) = \mathbf{0}$ and 
$$
\left\|\bz_{\cS}^{(\ell)}\right\|_2^2 = \sum_{i \in \cS_+}\left(x_{\cS}^{(i,\ell)} - \frac{e^{\theta^*_i}}{\sum_{j \in \cS_+} e^{\theta^*_j}}\right)^2 \!\!\le \sum_{i \in \cS_+} (x_{\cS}^{(i,\ell)})^2 + \sum_{i \in \cS_+} \frac{(e^{\theta^*_i})^2 }{\left(\sum_{j \in \cS_+} e^{\theta^*_j}\right)^2} \le 1 + \sum_{i \in \cS_+} \frac{e^{\theta^*_i} }{\sum_{j \in \cS_+} e^{\theta^*_j}} =2,
$$
\begin{align*}
	\EE\left(\bz_{\cS}^{(\ell)} \bz_{\cS}^{(\ell) \top}\right) & = \sum_{\substack{i,l\in \cS_+\\i < l}} \frac{e^{\theta_i^*}e^{\theta_l^*}}{\left(\sum_{j \in \cS_+} e^{\theta_j^*}\right)^2}(\eb_i - \eb_l) (\eb_i - \eb_l)^{\top} \preceq \kappa_{\thetab}^2 \sum_{\substack{i,l\in \cS_+\\i < l}} \frac{1}{|\cS_+|^2}(\eb_i - \eb_l) (\eb_i - \eb_l)^{\top} ;\\
	\EE\left(\bz_{\cS}^{(\ell) \top} \bz_{\cS}^{(\ell)}\right) & =\sum_{i \in \cS_+} \frac{e^{\theta^*_i} }{\sum_{j \in \cS_+} e^{\theta^*_j}} - \sum_{i \in \cS_+} \left( \frac{e^{\theta^*_i} }{\sum_{j \in \cS_+} e^{\theta^*_j}} \right)^2 \le 1,
\end{align*}
and thus under Assumption~\ref{asp: connectivity graph}
we have 
\begin{align*}
	&\left\| \frac{1}{L^2}\sum_{\cS \in \bcE} \sum_{\ell=1}^{L_{\cS}} \EE\left(\bz_{\cS}^{(\ell)} \bz_{\cS}^{(\ell) \top}\right) \right\|_2 \lesssim \frac{\kappa_{\thetab}^2}{L} \bigg\| \sum_{\cS \in \bcE} \frac{L_{\cS}}{L}\Big\{\frac{1}{|\cS_+|^2} \sum_{\substack{i,l\in \cS_+\\i < l}} (\eb_i - \eb_j)(\eb_i - \eb_j)^{\top}\Big\}\bigg\|_2 \lesssim \frac{1}{L} \|\Lb_{\bcE}\|_2 = \frac{\bar\lambda}{L} ;\\
	&\Big|\frac{1}{L^2}\sum_{S \in \bcE} \sum_{\ell=1}^{L_{\cS}} \EE\left(\bz_{\cS}^{(\ell)\top} \bz_{\cS}^{(\ell)}\right)\Big|  \le \left|\frac{1}{L}\sum_{S \in \bcE} \frac{L_{\cS}}{L}\right| \le \left|\frac{n+1}{L}\sum_{S \in \bcE} \frac{L_{\cS}}{L\cdot |\cS_+|}\right| \lesssim  \frac{n}{L} (\Lb_{\bcE})_{1,1} \le \frac{n}{L} \|\Lb_{\bcE}\|_2  =  \frac{n\bar\lambda}{L}.
\end{align*}
Thus we can take $V = \frac{n \bar\lambda}{L}$ and $B = 1/L $, then by the matrix Bernstein inequality \citep{tropp2012user}, with probability at least $1-O(n^{-10})$ we have that 
$$
\|\nabla \cL_{\tau} (\thetab^*; \bx) - \tau\thetab^*\|_2 \lesssim \sqrt{V \log n} + B\log n =  \sqrt{\frac{n \bar\lambda\log n}{L}} + \frac{\log n}{L} \lesssim \sqrt{\frac{n \bar\lambda\log n}{L}},
$$
where the last inequality follows by the condition that $n \bar\lambda L \gtrsim \log n$.

Thus when $\tau \asymp  \sqrt{\frac{\bar\lambda\log n}{L}} $, since we know that $\|\thetab^*\|_2 \le \sqrt{n+1} \log \kappa_{\thetab}$, with probability at least $1-O(n^{-10})$ we have
$$
\|\nabla \cL_{\tau} (\thetab^*; \bx) \|_2 \le \|\tau\thetab^*\|_2 + \|\nabla \cL_{\tau} (\thetab^*; \bx) - \tau\thetab^*\|_2  \lesssim   \sqrt{\frac{n \bar\lambda\log n}{L}}.
$$}
\subsection{Proof of Lemma \ref{lm: smooth hessian}}\label{sec: proof lm smooth hessian}
{ 
Note that for any $\thetab \in \RR^{n+1}$ and any $\cS \in \bcS$, we have that 
$$e^{\theta_i}e^{\theta_l}\Big/\Big(\sum_{j \in \cS_+}e^{\theta_j}\Big)^2 \le e^{\theta_i}\Big/\Big(\sum_{j \in \cS_+}e^{\theta_j}\Big) \Big(1-e^{\theta_i}\Big/\Big(\sum_{j \in \cS_+}e^{\theta_j}\Big)\Big) \le 1/4, \quad \forall i, l \in \cS_+,$$ 
and thus 
\begin{align*}
    \nabla^2 \cL(\thetab; \bx) &= \sum_{S \in \bcE} \frac{L_{\cS}}{L}\left\{ \sum_{\substack{i,l\in \cS_+\\i < l}} \frac{e^{\theta_i}e^{\theta_l}}{\left(\sum_{j \in \cS_+} e^{\theta_j}\right)^2}(\eb_i - \eb_l) (\eb_i - \eb_l)^{\top} \right\} \preceq \frac{1}{4} \sum_{S \in \bcE}\frac{L_{\cS}}{L} \left\{ \sum_{\substack{i,l\in \cS_+\\i < l}} (\eb_i - \eb_l) (\eb_i - \eb_l)^{\top} \right\}\\
    & \preceq \frac{(n+1)^2}{4} \sum_{S \in \bcE}\frac{L_{\cS}}{L \cdot|\cS_+|^2} \left\{ \sum_{\substack{i,l\in \cS_+\\i < l}} (\eb_i - \eb_l) (\eb_i - \eb_l)^{\top} \right\} = \frac{(n+1)^2}{4} \cdot \Lb_{\bcE}.
\end{align*}
Thus under Assumption~\ref{asp: connectivity graph},
we have that 
$$
\|\nabla^2 \cL_{\tau} (\thetab; \bx)\|_2 \le \tau +  \frac{(n+1)^2}{4} \|\Lb_{\bcE}\|_2  = \tau +  \frac{(n+1)^2 \bar\lambda}{4}. 
$$}
\subsection{Proof of Lemma \ref{lm: initial bound MLE}}\label{sec: proof lm initial bound}
By the optimality of $\hat\thetab$ and Taylor's expansion, we have 
\begin{align*}
	\cL_{\tau}(\hat\thetab; \bx) & = \cL_{\tau}(\thetab^*; \bx) +  (\hat\thetab - \thetab^*)^{\top}\nabla\cL_{\tau}(\thetab^*; \bx) + \frac{1}{2}(\hat\thetab - \thetab^*)^{\top}\nabla^2  \cL_{\tau}(\tilde\thetab; \bx) (\hat\thetab - \thetab^*)\le \cL_{\tau}(\thetab^*; \bx),
\end{align*}
where $\tilde\thetab$ lies between $\hat\thetab$ and $\thetab^*$. Thus 
\begin{align*}
	&\frac{1}{2}\lambda_{\min}\left(\nabla^2  \cL_{\tau}(\tilde\thetab; \bx)\right) \|\hat\thetab - \thetab^*\|_2^2 \le \frac{1}{2}(\hat\thetab - \thetab^*)^{\top}\nabla^2  \cL_{\tau}(\tilde\thetab; \bx) (\hat\thetab - \thetab^*)\\
	&\quad \le - (\hat\thetab - \thetab^*)^{\top}\nabla\cL_{\tau}(\thetab^*; \bx) \le \|\nabla\cL_{\tau}(\thetab^*; \bx)\|_2\|\hat\thetab - \thetab^*\|_2,
\end{align*}
and thus by Lemma \ref{lm: bound gradient}
$$
\|\hat\thetab - \thetab^*\|_2 \le \frac{2\|\nabla\cL_{\tau}(\thetab^*; \bx)\|_2}{\lambda_{\min}\left(\nabla^2  \cL_{\tau}(\tilde\thetab; \bx)\right)} \lesssim  \frac{2\|\nabla\cL_{\tau}(\thetab^*; \bx)\|_2}{\tau}  \lesssim \sqrt{n}.
$$
\if1\sharpbound{
	\subsection{Proof of Lemma \ref{lm: ind hypo 1}}
	The proof follows the proof of Lemma 14 in \cite{chen2019topk} with modifications. For the convenience of notations, we abbreviate $\cL_{\tau} (\thetab;\bx)$ to $\cL_{\tau}(\thetab)$ in the following proof. From the gradient descent algorithm we know that 
	\begin{align*}
		\boldsymbol{\theta}^{t+1}-\boldsymbol{\theta}^{*} &=\boldsymbol{\theta}^{t}-\eta \nabla \mathcal{L}_{\tau}\left(\boldsymbol{\theta}^{t}\right)-\boldsymbol{\theta}^{*} \\
		&=\boldsymbol{\theta}^{t}-\eta \nabla \mathcal{L}_{\tau}\left(\boldsymbol{\theta}^{t}\right)-\left[\boldsymbol{\theta}^{*}-\eta \nabla \mathcal{L}_{\tau}\left(\boldsymbol{\theta}^{*}\right)\right]-\eta \nabla \mathcal{L}_{\tau}\left(\boldsymbol{\theta}^{*}\right) \\
		&=\left\{\boldsymbol{I}_{n}-\eta \int_{0}^{1} \nabla^{2} \mathcal{L}_{\tau}(\boldsymbol{\theta}(\nu)) \mathrm{d} \nu\right\}\left(\boldsymbol{\theta}^{t}-\boldsymbol{\theta}^{*}\right)-\eta \nabla \mathcal{L}_{\tau}\left(\boldsymbol{\theta}^{*}\right),
	\end{align*}
	where $\thetab(\nu) := \thetab^* + \nu(\thetab^t - \thetab^*)$. Then we can see that if 
	$$
	C_4  \sqrt{\frac{\log n}{ M_n p L}} \le \epsilon,
	$$
	for a sufficiently small constant $\epsilon$, we have that $\|\thetab(\nu) - \thetab^*\|_{\infty} \le \nu\|\thetab^t - \thetab^*\|_{\infty} \le \epsilon $. Then by Lemma \ref{lm: eigen hessian} we have that under event $\cA_{\Lb_{\bcE}}$ defined in \eqref{eq: eigen of L cE} (which occurs with probability at least $1-O(n^{-C_L})$), for all $\nu \in [0,1]$
	$$
	\lambda + \frac{1}{3 \kappa_{\thetab}^2} nM_n p \le \lambda_{\min, \perp}\left(\nabla^{2} \mathcal{L}_{\tau}(\boldsymbol{\theta}(\nu))\right) \le \lambda_{\max}\left(\nabla^{2} \mathcal{L}_{\tau}(\boldsymbol{\theta}(\nu))\right) \le \lambda + n 2^{n} p/4 .
	$$
	Now we denote $\Bb = \int_{0}^{1} \nabla^{2} \mathcal{L}_{\tau}(\boldsymbol{\theta}(\nu)) \mathrm{d} \nu$, then we have 
	\begin{align*}
		\Bb \mathbf{1} &= \int_{0}^{1} \nabla^{2} \mathcal{L}_{\tau}(\boldsymbol{\theta}(\nu)) \mathbf{1} \mathrm{d} \nu = \mathbf{0};\\
		\lambda_{\min , \perp} (\Bb ) & = \min_{\vb \in \Sb^{n}, \vb^{\top} \mathbf{1} = 0} \int_{0}^{1} \vb^{\top} \nabla^{2} \mathcal{L}_{\tau}(\boldsymbol{\theta}(\nu)) \vb \mathrm{d} \nu \ge  \int_{0}^{1} \lambda_{\min, \perp}\left( \nabla^{2} \mathcal{L}_{\tau}(\boldsymbol{\theta}(\nu)) \right) \mathrm{d} \nu\\
		& \ge \lambda + \frac{1}{3 \kappa_{\thetab}^2} nM_n p; \\
		\lambda_{\max}(\Bb) & = \max_{\vb \in \Sb^n} \int_{0}^{1} \vb^{\top} \nabla^{2} \mathcal{L}_{\tau}(\boldsymbol{\theta}(\nu)) \vb \mathrm{d} \nu \le \lambda + n 2^n p /4. 
	\end{align*}
	Besides, by the definition of $\eta$ we have that $1-\eta\left(\lambda + \frac{1}{3 \kappa_{\thetab}^2} nM_n p\right) \ge 1 - \eta \left(\lambda + n 2^n p /4\right) \ge 0$. Combining the above results we have
	\begin{align*}
		\|\thetab^{t+1} - \thetab^*\| &\le \left(1-\frac{1}{3 \kappa_{\thetab}^2} \eta n M_n p\right)\|\thetab^t - \thetab^*\| + \eta\|\nabla \mathcal{L}_{\tau}\left(\boldsymbol{\theta}^{*}\right)\|\\
		& \le \left(1-\frac{1}{3 \kappa_{\thetab}^2} \eta n M_n p\right) C_1 \sqrt{\frac{\log n}{M_n pL }} + C \eta n \sqrt{\frac{M_n p \log n}{L}} \\
		& \le C_1  \sqrt{\frac{\log n}{M_n pL }},
	\end{align*}
	so long as $C_1$ is large enough.
	\subsection{Proof of Lemma \ref{lm: ind hypo 2}}
	According to the definition of the leave-one-out sequence defined in Section \ref{sec: proof thm inf norm mle}, we have 
	\begin{align*}
		\theta_m^{t+1,(m)} - \theta_m^* & = \theta_m^{t,(m)} - \eta\left[\nabla \cL_{\tau}^{(m)} (\thetab^{t,(m)})\right]_m - \eta\lambda \theta_{m}^{t,(m)} - \theta_m^*\\
		& = \theta_m^{t,(m)} - \theta_m^* - \eta p\!\!\!\sum_{S \in \cI, m \in S} \!\!\left\{ \frac{e^{\theta^{t,(m)}_m}}{\sum_{j \in \cS_+} e^{\theta_j^{t,(m)}}} - \frac{e^{\theta_m^*}}{\sum_{j \in \cS_+} e^{\theta_j^*}}\right\}  - \eta \lambda\theta_m^{t,(m)},
	\end{align*}
	where by the mean theorem we have 
	\begin{align*}
		& \sum_{S \in \cI, m \in S} \!\!\left\{ \frac{e^{\theta^{t,(m)}_m}}{\sum_{j \in S e^{\theta_j^{t,(m)}}}} - \frac{e^{\theta_m^*}}{\sum_{j \in \cS_+} e^{\theta_j^*}}\right\} = \underbrace{\left\{ \sum_{S \in \cI, m \in S}  \frac{e^{\tilde{\theta}_m \left(\sum_{j \in S, j \neq m}e^{\tilde{\theta}_j}\right)}}{\left(\sum_{j \in \cS_+}e^{\tilde{\theta}_j}\right)^2} \right\}}_{\rm I} (\theta_m^{t,(m)} - \theta_m^*)\\
		& \quad - \underbrace{\sum_{S \in \cI, m \in S} \sum_{l \in S, l \neq m} \left\{\frac{e^{\tilde{\theta}_m}e^{\tilde{\theta}_l}}{\left(\sum_{j \in \cS_+}e^{\tilde{\theta}_j}\right)^2} (\theta_l^{t, (m)} - \theta^*_l)\right\}}_{\rm II},
	\end{align*}
	where we abuse the notation and let $\tilde{\thetab} = \{\tilde{\theta}_j\}_{j=0}^{n}$ represent the vector that lies between $\thetab^{t, (m)}$ and $\thetab^*$. Now from the proof of Theorem \ref{thm: inf norm MLE} in Section \ref{sec: proof thm inf norm mle}, we know that { there exists some small enough constant $\epsilon >0$ such that 
		$\|\tilde{\thetab} - \thetab^*\|_{\infty} \le \|\thetab^{t, (m)} - \thetab^*\|_{\infty} \le \epsilon$}, and thus we have 
	$$
	\frac{1}{2 \kappa_{\thetab} |S|}\le \frac{1}{e^{2\epsilon}\kappa_{\thetab} |S|}\le \frac{e^{\tilde{\theta}_i}}{\sum_{j \in \cS_+}e^{\tilde{\theta}_j}} \le \frac{\kappa_{\thetab} e^{2\epsilon}}{|S|} \le \frac{2 \kappa_{\thetab} }{|S|}.
	$$
	Then we have the following bounds on term I and term II
	\begin{align*}
		& I =  \sum_{k=2}^{n+1} \sum_{S \in \cI_k, m \in S}  \frac{e^{\tilde{\theta}_m \left(\sum_{j \in S, j \neq m}e^{\tilde{\theta}_j}\right)}}{\left(\sum_{j \in \cS_+}e^{\tilde{\theta}_j}\right)^2} \ge \frac{1}{4\kappa_{\thetab}^2} \sum_{k=2}^{n+1} \frac{k-1}{k^2} {n \choose k-1} = \frac{n}{4\kappa_{\thetab}^2} \sum_{k=2}^{n+1} \frac{1}{k^2} {n-1 \choose k-2} = \frac{n M_n}{4\kappa_{\thetab}^2};\\
		& II = \sum_{k=2}^{n+1} \sum_{S \in \cI_k, m \in S} \sum_{l \in S, l \neq m} \left\{\frac{e^{\tilde{\theta}_m}e^{\tilde{\theta}_l}}{\left(\sum_{j \in \cS_+}e^{\tilde{\theta}_j}\right)^2} (\theta_l^{t, (m)} - \theta^*_l)\right\} \\
		& \quad \le 4\kappa_{\thetab}^2 M_n \sum_{i \in [n]_+, i \neq m} |\theta_i^{t,(m)} - \theta_i^*| \le 5 \kappa_{\thetab}^2 \sqrt{n}M_n \|\thetab^{t,(m)} - \thetab^*\|.
	\end{align*}
	Note that 
	$$
	1 - \eta \lambda - \frac{\eta n M_n p }{4 \kappa_{\thetab}^2} \ge 1 - \eta \left( \lambda + \frac{ n 2^n p }{4}\right) = 0,
	$$
	and in turn we have 
	\begin{align*}
		& \theta_m^{t+1, (m)} - \theta_m^* = (1 - \eta\lambda)(\theta_m^{t,(m)} - \theta_m^*) - \eta p\!\!\!\sum_{S \in \cI, m \in S} \!\!\left\{ \frac{e^{\theta^{t,(m)}_m}}{\sum_{j \in S e^{\theta_j^{t,(m)}}}} - \frac{e^{\theta_m^*}}{\sum_{j \in \cS_+} e^{\theta_j^*}}\right\}  - \eta \lambda\theta_m^*\\
		&\quad \le \left(1 - \eta \lambda - \frac{\eta n M_n p }{4 \kappa_{\thetab}^2}\right)\left|\theta_m^{t,(m)} - \theta_m^*\right| + 5 \eta \kappa_{\thetab}^2 \sqrt{n}M_n p \|\thetab^{t,(m)} - \thetab^*\| + \eta \lambda \|\thetab^*\|_{\infty}\\
		& \quad \le \left(1 - \eta \lambda - \frac{\eta n M_n p }{4 \kappa_{\thetab}^2}\right)C_2 \sqrt{\frac{\log n}{n M_n p L}}  + 5 C_6 \eta \kappa_{\thetab}^2 \sqrt{n}M_n p \sqrt{\frac{\log n}{M_n p L}} + c_{\lambda} \eta \log \kappa_{\thetab} \sqrt{\frac{n M_n p \log n}{L}}\\
		& \quad \le C_2 \sqrt{\frac{\log n}{n M_n p L}} ,
	\end{align*}
	so long as $C_2$ is large enough, where $c_{\lambda} : = \lambda \left(\sqrt{\frac{n M_n p \log n}{L}}\right)^{-1}$ depends on the setting of $\lambda$.
	\subsection{Proof of Lemma \ref{lm: ind hypo 3}}
	Similar to the proof of Lemma 16 in \cite{chen2019topk}, we have 
	\begin{align*}
		& \boldsymbol{\theta}^{t+1}-\boldsymbol{\theta}^{t+1,(m)}= \boldsymbol{\theta}^{t}-\eta \nabla \mathcal{L}_{\tau}\left(\boldsymbol{\theta}^{t}\right)-\left[\boldsymbol{\theta}^{t,(m)}-\eta \nabla \mathcal{L}_{\tau}^{(m)}\left(\boldsymbol{\theta}^{t,(m)}\right)\right] \\
		&\quad = \boldsymbol{\theta}^{t}-\eta \nabla \mathcal{L}_{\tau}\left(\boldsymbol{\theta}^{t}\right)-\left[\boldsymbol{\theta}^{t,(m)}-\eta \nabla \mathcal{L}_{\tau}\left(\boldsymbol{\theta}^{t,(m)}\right)\right] -\eta\left(\nabla \mathcal{L}_{\tau}\left(\boldsymbol{\theta}^{t,(m)}\right)-\nabla \mathcal{L}_{\tau}^{(m)}\left(\boldsymbol{\theta}^{t,(m)}\right)\right) \\
		& \quad = \underbrace{\left(\boldsymbol{I}_{n}-\eta \int_{0}^{1} \nabla^{2} \mathcal{L}_{\tau}(\boldsymbol{\theta}^{(m)}(\nu)) \mathrm{d} \nu\right)\left(\boldsymbol{\theta}^{t}-\boldsymbol{\theta}^{t,(m)}\right)}_{:=\boldsymbol{v}_{1}} -\underbrace{\eta\left(\nabla \mathcal{L}_{\tau}\left(\boldsymbol{\theta}^{t,(m)}\right)-\nabla \mathcal{L}_{\tau}^{(m)}\left(\boldsymbol{\theta}^{t,(m)}\right)\right)}_{:=\boldsymbol{v}_{2}},
	\end{align*}
	where $\thetab^{(m)}(\nu) : = \thetab^{t,(m)} + \nu(\thetab^t - \thetab^{t,(m)})$. Then for $\bv_1$, with similar argument as the proof of Lemma \ref{lm: ind hypo 1}, under the event $\cA_{\Lb_{\bcE}}$ we have that
	$$
	\|\bv_1\| \le \left(1-\frac{1}{3 \kappa_{\thetab}^2} \eta n M_n p\right)\|\thetab^t - \thetab^{t,(m)}\|.
	$$
	As for $\bv_2$ we have 
	\begin{align*}
		-\eta^{-1}\bv_2 & = \sum_{S \in \cI, m \in S} \left\{\cE_{S} \sum_{i \in \cS_+} \left(x_{S}^{(i)} - \frac{e^{\theta_i^{t,(m)}}}{\sum_{j \in \cS_+}e^{\theta_j^{t,(m)}}}\right)\eb_i\right\} \\
		& \quad - p \sum_{S \in \cI, m \in S} \left\{\sum_{i \in \cS_+} \left(\frac{e^{\theta_i^{*}}}{\sum_{j \in \cS_+}e^{\theta_j^{*}}} - \frac{e^{\theta_i^{t,(m)}}}{\sum_{j \in \cS_+}e^{\theta_j^{t,(m)}}}\right)\eb_i\right\} \\
		& = \underbrace{ \sum_{S \in \bcE, m \in S} \left\{\frac{1}{L} \sum_{i \in \cS_+} \sum_{\ell = 1}^L\left(x_{\cS}^{(i,\ell)} - \frac{e^{\theta_i^{*}}}{\sum_{j \in \cS_+}e^{\theta_j^{*}}}\right)\eb_i\right\} }_{\bu^m} \\
		& \quad + \underbrace{\sum_{S \in \cI, m \in S} \left\{\sum_{i \in \cS_+} \left(\frac{e^{\theta_i^{*}}}{\sum_{j \in \cS_+}e^{\theta_j^{*}}} - \frac{e^{\theta_i^{t,(m)}}}{\sum_{j \in \cS_+}e^{\theta_j^{t,(m)}}}\right)\eb_i\right\} (\cE_{S} - p)}_{\bv^m}.
	\end{align*}
	For $i \in [n]_+$ and $k = 2, 3, \ldots, n + 1 $, define 
	\begin{align*}
		n_{i,k} =\left\{\begin{array}{ll}
			\sum_{\substack{S \in \cI_k\\ m, i \in S }} \cE_{S} & \text { if } i \neq m \\
			\sum_{\substack{S \in \cI_k,  m \in S }} \cE_{S} & \text { if } i = m,
		\end{array}\right.
	\end{align*}
	Then conditional on $\bcE$, we first aim to prove the following inequality holds true with probability at least $1 - n^{-(C_L + 1)}$ for each $i \in [n]_+$
	\begin{equation}\label{eq: ind hypo 4 ineq um}
		\bigg|\frac{1}{L}\sum_{k =2}^{n+1} \sum_{\substack{S \in \bcE_k \\ m, i \in S }} \sum_{\ell = 1}^L \left(x_{\cS}^{(i,\ell)} - \frac{e^{\theta_i^{*}}}{\sum_{j \in \cS_+}e^{\theta_j^{*}}}\right)\bigg| \lesssim \sqrt{\frac{ \sum_{k=2}^{n+1} k^{-1} n_{i,k} \log n}{L}}.
	\end{equation}
	First, for the case $\sum_{k=2}^{n+1} n_{i,k} = 0$, \eqref{eq: ind hypo 4 ineq um} holds true trivially. When $\sum_{k=2}^{n+1} n_{i,k} > 0$, observe that $x_{\cS}^{(i,\ell)}$'s are independent for any $\ell \in [L]$ and $S \in \bcE$. Also since $ \bigg| x_{\cS}^{(i,\ell)} - \frac{e^{\theta_i^{*}}}{\sum_{j \in \cS_+}e^{\theta_j^{*}}}\bigg| \le 1$ and $\operatorname{Var}(x_{\cS}^{(i,\ell)}) \le e^{\theta_i^*}/(\sum_{j \in \cS_+}e^{\theta_j^*}) \le \kappa_{\thetab} / |S|$, by Bernstein's inequality we have
	\begin{align*}
		& \bigg|\frac{1}{L}\sum_{k =2}^{n+1} \sum_{\substack{S \in \bcE_k \\ m, i \in S }} \sum_{\ell = 1}^L \left(x_{\cS}^{(i,\ell)} - \frac{e^{\theta_i^{*}}}{\sum_{j \in \cS_+}e^{\theta_j^{*}}}\right)\bigg|  \lesssim \sqrt{\frac{ \sum_{k=2}^{n+1} k^{-1} n_{i,k} \log n}{L}} + \frac{\log n}{L} \lesssim \sqrt{\frac{ \sum_{k=2}^{n+1} k^{-1} n_{i,k} \log n}{L}},
	\end{align*}
	under the condition that $ L  \gtrsim n\log n$.
	Also, by Bernstein’s inequality, under the condition that $c_p \ge 1$, with probability at least $1 - O(n^{-C_L})$ we have
	\begin{align*}
		\sum_{i \in [n]_+} \sum_{k=2}^{n+1} k^{-1} n_{i,k} & = \sum_{k=2}^{n+1} k^{-1} \sum_{i \in [n]_+} n_{i,k} = \sum_{k=2}^{n+1} k^{-1} \cdot k \cdot \sum_{S \in \cI_k : m \in S} \cE_{S} \\
		& \lesssim p\sum_{k=2}^{n+1} {n \choose k-1} \le  n^2 p M_n,
	\end{align*}
	and thus with probability at least $1 - O(n^{-C_L})$ we have that the event 
	\begin{align*}
		\cA_{\bu} := \left\{\|\bu^m\| \lesssim \sqrt{\frac{ \sum_{i \in [n]_+}\sum_{k=2}^{n+1} k^{-1} n_{i,k} \log n}{L}} \lesssim n\sqrt{\frac{   M_n p \log n}{L}}\right\}
	\end{align*}
	holds for any $t \in [T]$. Now we move on to $\bv^m$. We denote $$\bv_{S}^m = \left\{\sum_{i \in \cS_+} \left(\frac{e^{\theta_i^{*}}}{\sum_{j \in \cS_+}e^{\theta_j^{*}}} - \frac{e^{\theta_i^{t,(m)}}}{\sum_{j \in \cS_+}e^{\theta_j^{t,(m)}}}\right)\eb_i\right\} (\cE_{S} - p),$$ then note that $\bv_{S}^m$ are independent across $S \in \cI$ with $m \in S$. Under the condition that { $\|\thetab^{t,(m)} - \thetab^*\|_{\infty} \le C_5  \sqrt{\frac{\log n}{ M_n p L}} \le \epsilon$ for small enough constant $\epsilon > 0$, } by the mean value theorem we have that for each $ i \in S$
	\begin{align*}
		&\Bigg|\frac{e^{\theta_i^{*}}}{\sum_{j \in \cS_+}e^{\theta_j^{*}}} - \frac{e^{\theta_i^{t,(m)}}}{\sum_{j \in \cS_+}e^{\theta_j^{t,(m)}}} \Bigg|= \Bigg|\sum_{l \neq i} \frac{e^{\tilde{\theta}_i \tilde{\theta}_l}}{(\sum_{j \in \cS_+} e^{\tilde{\theta}_j})^2} (\theta_l^{t,(m)} - \theta_l^*) - \frac{e^{\tilde{\theta}_i} (\sum_{j \in S, j \neq i} e^{\tilde{\theta}_j})}{(\sum_{j \in \cS_+} e^{\tilde{\theta}_j})^2}(\theta_i^{t,(m)} - \theta_i^*)\Bigg|\\
		& \le  \left(\sum_{l \neq i} \frac{e^{\tilde{\theta}_i \tilde{\theta}_l}}{(\sum_{j \in \cS_+} e^{\tilde{\theta}_j})^2} + \frac{e^{\tilde{\theta}_i} (\sum_{j \in S, j \neq i} e^{\tilde{\theta}_j})}{(\sum_{j \in \cS_+} e^{\tilde{\theta}_j})^2}\right) \|\thetab^{t,(m)} - \thetab^*\|_{\infty} \le \frac{2 e^{\tilde{\theta}_i}}{\sum_{j \in \cS_+} e^{\tilde{\theta}_j}} \|\thetab^{t,(m)} - \thetab^*\|_{\infty} \\
		& \le \frac{3 \kappa_{\thetab}}{|S|}\|\thetab^{t,(m)} - \thetab^*\|_{\infty} ,
	\end{align*}
	where $\tilde{\thetab}$ lies between $\thetab^{t, (m)}$ and $\thetab^*$. And in turn we have that 
	\begin{align*}
		&\|\bv_{S}^m\|^2  \le  \sum_{i \in \cS_+} \left(\frac{e^{\theta_i^{*}}}{\sum_{j \in \cS_+}e^{\theta_j^{*}}} - \frac{e^{\theta_i^{t,(m)}}}{\sum_{j \in \cS_+}e^{\theta_j^{t,(m)}}}\right)^2 \lesssim \frac{1}{|S|} \|\thetab^{t,(m)} - \thetab^*\|_{\infty}^2;\\
		&\sum_{S \in \cI: m \in S} \EE \|\bv_{S}^m\|^2 \lesssim p \|\thetab^{t,(m)} - \thetab^*\|_{\infty}^2 \sum_{k = 2}^n \frac{1}{k} {n \choose k-1} \lesssim  n p \|\thetab^{t,(m)} - \thetab^*\|_{\infty}^2 \sum_{k=2}^{n+1} \frac{1}{k^2} {n - 1 \choose k -2} \\
		& \quad =  n M_n p \|\thetab^{t,(m)} - \thetab^*\|_{\infty}^2.
	\end{align*}
	Also by Jensen's inequality we have
	\begin{align*}
		\left\|\sum_{S \in \cI, m \in S} \EE (\bv_{S}^m)(\bv_{S}^m)^{\top} \right\|& \le \sum_{S \in \cI, m \in S} \EE \| (\bv_{S}^m)(\bv_{S}^m)^{\top} \| \le \sum_{S \in \cI, m \in S} \EE \|\bv_{S}^m\|^2\\
		& \lesssim  n M_n p \|\thetab^{t,(m)} - \thetab^*\|_{\infty}^2.
	\end{align*}
	Thus we can take $V =  n M_n p \|\thetab^{t,(m)} - \thetab^*\|_{\infty}^2$ and $B =  \|\thetab^{t,(m)} - \thetab^*\|_{\infty}$, then by the matrix Bernstein inequality \citep{tropp2012user}, with probability at least $1 - O(n^{-C_L})$ we have 
	\begin{align*}
		\|\bv^m\| & = \left\|\sum_{S \in \cI, m \in S} \bv_{S}^m\right\| \lesssim \sqrt{V \log n} + B\log n \lesssim  \sqrt{nM_np \log n}  \|\thetab^{t,(m)} - \thetab^*\|_{\infty}.
	\end{align*}
	Combining the above results we have $\|\bv_2\| \lesssim \eta\left(n\sqrt{\frac{  M_n p \log n}{L}} +  \sqrt{nM_np \log n}  \|\thetab^{t,(m)} - \thetab^*\|_{\infty}\right)$, and thus with probability at least $1-O(n^{-C_L})$, there exists some constant $C>0$ such that 
	\begin{align*}
		\|\thetab^{t+1} - \thetab^{t+1,(m)}\| & \le \left(1-\frac{1}{3 \kappa_{\thetab}^2} \eta n M_n p\right)\|\thetab^t - \thetab^{t,(m)}\| \\
		&\quad + C\eta\left(n\sqrt{\frac{ M_n p \log n}{L}} +  \sqrt{nM_np \log n}  \|\thetab^{t,(m)} - \thetab^*\|_{\infty}\right)\\
		& \le \left(1-\frac{1}{3 \kappa_{\thetab}^2} \eta n M_n p\right)C_3 \sqrt{\frac{ \log n}{M_n p L}}\\
		& \quad + C\eta\left(n\sqrt{\frac{ M_n p \log n}{L}} + C_5  \sqrt{nM_np \log n}  \cdot  \sqrt{\frac{\log n}{M_n p L}}  \right)\\
		& \le C_3 \sqrt{\frac{ \log n}{M_n p L}},
	\end{align*}
	under the condition that $C_3$ is sufficiently large { and $c_p > 1$}. 
	\subsection{Proof of Lemma \ref{lm: ind hypo 4}}
	For any $m \in [n]_+$, we have that with probability at least $1 - O(n^{-C_L})$ there exists a sufficiently large constant $C_4 > C_3 > 0$,
	\begin{align*}
		|\theta^{t+1}_m - \theta_m^*| &\le |\theta^{t+1}_m - \theta^{t, (m)}_m| + |\theta^{t, (m)}_m - \theta_m^*| \le  \max_{m \in [n]_+}\|\thetab^{t+1}- \thetab^{t, (m)}\| + \max_{m \in [n]_+}|\theta^{t, (m)}_m - \theta_m^*|\\
		& \le C_3 \sqrt{\frac{ \log n}{M_n p L}} +  C_2 \sqrt{\frac{\log n}{n M_n p L}} \\
		& \le C_4  \sqrt{\frac{\log n}{ M_n p L}}.
	\end{align*}
}\fi
\subsection{Proof of Lemma \ref{lm: cck term bounds}}\label{sec: proof lm cck term bounds}
In this section, we provide the proof for the upper bounds given in Lemma \ref{lm: cck term bounds}.
\subsubsection{Proof of \ref{eq: cck term 1}}
{ 
Note that for any $i \in [n]_+$, we have that $\left|x_{\cS}^{(i,\ell)} - \frac{e^{\theta_i^*}}{\sum_{j \in \cS_+}e^{\theta_j^*}} \right| \le 1$,  $\operatorname{Var}(x_{\cS}^{(i,\ell)} ) \le {e^{\theta_i^*}}/\left({\sum_{j \in \cS_+}e^{\theta_j^*}}\right) \lesssim 1/|\cS_+| $, and
$$\frac{1}{L^2}  \sum_{\cS \in \bcE: i \in \cS_+} \sum_{\ell =1}^{L_{\cS}} \operatorname{Var}\left(x_{\cS}^{(i, \ell)} - \frac{e^{\theta_i^*}}{\sum_{j \in \cS_+}e^{\theta_j^*}}\right) \lesssim \frac{1}{L} \sum_{\cS \in \bcE: i \in \cS_+} \frac{L_{\cS}}{L \cdot |\cS_+|} \lesssim \frac{1}{L}(\Lb_{\bcE} )_{i+1,i+1} \le \frac{\bar\lambda}{L}.$$
By Bernstein's inequality, conditional on $\bcE$, with probability at least $1 - O(n^{-11})$, we have
\begin{align*}
	& \left|\frac{1}{L} \sum_{\cS \in \bcE: i \in \cS_+} \sum_{\ell =1}^{L_{\cS}} \left(x_{\cS}^{(i, \ell)} - \frac{e^{\theta_i^*}}{\sum_{j \in \cS_+}e^{\theta_j^*}}\right) \right| \lesssim \sqrt{\frac{ \bar\lambda\log n}{L}} + \frac{\log n}{L}.
\end{align*}
Hence taking the union bound over $i \in [n]_+$, with probability at least $1 - O(n^{-10})$, we have 
\begin{align*}
	\|\nabla \cL(\thetab^*; \bx)\|_{\infty} & = \max_{i \in [n]_+}  \left|\frac{1}{L} \sum_{\cS \in \bcE: i \in \cS_+} \sum_{\ell =1}^{L_{\cS}} \left(x_{\cS}^{(i, \ell)} - \frac{e^{\theta_i^*}}{\sum_{j \in \cS_+}e^{\theta_j^*}}\right) \right| \lesssim \sqrt{\frac{ \bar\lambda\log n}{L}} + \frac{\log n}{L}.
\end{align*}}
\subsubsection{Proof of \ref{eq: cck term 2}}\label{sec: proof of cck term 2}
{
We first consider $\nabla \cL(\hat\thetab; \bx) - \nabla \cL(\thetab^*; \bx)$. For $i \in [n]_+$, by the mean value theorem we have
\begin{align*}
	&\left[\nabla \cL(\hat\thetab; \bx) - \nabla \cL(\thetab^*; \bx) \right]_{i+1}=  \sum_{\cS \in \bcE: i \in \cS_+}  \frac{L_{\cS}}{L}\left(\frac{e^{\hat\theta_i}}{\sum_{j \in \cS_+} e^{\hat\theta_j}} - \frac{e^{\theta^*_i}}{\sum_{j \in \cS_+} e^{\theta^*_j}}\right) \\
	& = \sum_{\cS \in \bcE: i \in \cS_+} \frac{L_{\cS}}{L}\left\{\frac{e^{\tilde{\theta}_i}\left(\sum_{j \in \cS_+ \backslash \{i\}} e^{\tilde{\theta}_j}\right)}{\left(\sum_{j \in \cS_+} e^{\tilde{\theta}_j}\right)^2} \eb_i -  \sum_{l \in \cS_+ \backslash\{i\}} \frac{e^{\tilde{\theta}_i} e^{\tilde{\theta}_l} }{\left(\sum_{j \in \cS_+} e^{\tilde{\theta}_j}\right)^2}\eb_l \right\}^{\top} (\hat\thetab - \thetab^*),
\end{align*}
where $\tilde\thetab$ lies on the line between $\hat\thetab - \thetab^*$. Then we have 
\begin{align*}
	& \left|\left[\nabla \cL(\hat\thetab; \bx) - \nabla \cL(\thetab^*; \bx) - \nabla^2 \cL(\thetab^* ; \bx)(\hat\thetab - \thetab^*) \right]_{i+1}\right|\\
	&\le \left\| \sum_{\cS \in \bcE: i \in \cS_+} \frac{L_{\cS}}{L}\left\{\frac{e^{\tilde{\theta}_i}\left(\sum_{l \in \cS_+ \backslash \{i\}} e^{\tilde{\theta}_l}\right)}{\left(\sum_{j \in \cS_+} e^{\tilde{\theta}_j}\right)^2} \eb_i -  \sum_{l \in \cS_+ \backslash\{i\}} \frac{e^{\tilde{\theta}_i} e^{\tilde{\theta}_l} }{\left(\sum_{j \in \cS_+} e^{\tilde{\theta}_j}\right)^2}\eb_l \right\}  - [ \nabla^2 \cL(\thetab^* ; \bx)]_{i+1} \right\|_1 \|\hat{\thetab} - \thetab^*\|_{\infty}.
\end{align*}
Now consider any $\cS \in \bcE$ such that $i \in \cS_+$. Under the condition that $ \sqrt{{n \kappa_{\bcE} \log n}/{(\underline{\lambda}L)}}  \le c$ for some small enough constant $c > 0$,  by Theorem \ref{thm: inf norm MLE} there exists a small enough constant $\epsilon >0$ such that $\|\hat\thetab - \thetab^*\|_{\infty} \le \epsilon$ with probability at least $1 - O(n^{-10})$. Then with probability at least $1 -  O(n^{-10})$, for all $\tilde\thetab$ lies between $\hat\thetab$ and $\thetab^*$ and all $l \in \cS_+$,  we have
\begin{align*}
	\frac{e^{\tilde{\theta}_l}}{\sum_{j \in \cS_+}e^{\tilde\theta_j}} & =  \frac{e^{\theta_l^* + (\tilde{\theta}_l - \theta_l^*)}}{\sum_{j \in \cS_+}e^{\theta_j^* + (\tilde\theta_j - \theta_j^*)}} \le e^{2\|\hat\thetab - \thetab^*\|_{\infty}} \left(\frac{e^{\theta_l^*}}{\sum_{j \in \cS_+} e^{\theta_j^*}}\right) \le e^{2\epsilon}  \left(\frac{e^{\theta_l^*}}{\sum_{j \in \cS_+} e^{\theta_j^*}}\right) \lesssim \frac{e^{\theta_l^*}}{\sum_{j \in \cS_+} e^{\theta_j^*}}.
\end{align*}}
Then for any $l \in \cS_+$, by the mean value theorem,
\begin{align*}
	&\left| \frac{e^{\tilde{\theta}_l}}{\sum_{j \in \cS_+}e^{\tilde\theta_j}} \!-\! \frac{e^{{\theta}^*_l}}{\sum_{j \in \cS_+}e^{\theta^*_j}} \right| \!\!\lesssim \frac{e^{{\theta}^*_l}}{\sum_{j \in \cS_+}e^{\theta^*_j}}\left(\!\!1 -\! \frac{e^{{\theta}^*_l}}{\sum_{j \in \cS_+}e^{\theta^*_j}}\right)\!|\tilde\theta_l - \theta_l^*| \!+\! \frac{e^{{\theta}^*_l}}{\sum_{j \in \cS_+}e^{\theta^*_j}}\!\! \sum_{j \in \cS_+ \backslash \{l\}} \!\!\frac{e^{{\theta}^*_j}}{\sum_{j \in \cS_+}e^{\theta^*_j}} |\tilde\theta_j - \theta_j^*|\\
	& \quad \lesssim \frac{e^{{\theta}^*_l}}{\sum_{j \in \cS_+}e^{\theta^*_j}}\|\tilde\thetab - \thetab^*\|_{\infty} \le \frac{e^{{\theta}^*_l}}{\sum_{j \in \cS_+}e^{\theta^*_j}}\|\hat\thetab - \thetab^*\|_{\infty} .
\end{align*}
In turn for any $ l \in \cS_+ \backslash \{i\}$, by we have
\begin{align*}
	& \left| \frac{e^{\tilde{\theta}_i} e^{\tilde{\theta}_l} }{\left(\sum_{j \in \cS_+} e^{\tilde{\theta}_j}\right)^2} - \frac{e^{{\theta}^*_i} e^{{\theta}^*_l} }{\left(\sum_{j \in \cS_+} e^{{\theta}^*_j}\right)^2} \right| \le \left|\frac{e^{\tilde{\theta}_i} e^{\tilde{\theta}_l} }{\left(\sum_{j \in \cS_+} e^{\tilde{\theta}_j}\right)^2} -\left(\frac{e^{\tilde\theta_i}}{\sum_{j \in \cS_+} e^{\tilde\theta_j}}\right)\left(\frac{e^{\theta_i^*}}{\sum_{j \in \cS_+} e^{\theta_j^*}}\right)\right|\\ & \quad +\left|\left(\frac{e^{\tilde\theta_i}}{\sum_{j \in \cS_+} e^{\tilde\theta_j}}\right)\left(\frac{e^{\theta_i^*}}{\sum_{j \in \cS_+} e^{\theta_j^*}}\right) -  \frac{e^{{\theta}^*_i} e^{{\theta}^*_l} }{\left(\sum_{j \in \cS_+} e^{{\theta}^*_j}\right)^2}\right|\\
	& \lesssim \left(\frac{e^{{\theta}^*_i}}{\sum_{j \in \cS_+}e^{\theta^*_j}}\right)\left(\frac{e^{{\theta}^*_l}}{\sum_{j \in \cS_+}e^{\theta^*_j}}\right)\|\hat\thetab - \thetab^*\|_{\infty} .
\end{align*}
{ Thus with probability at least $1 - O(n^{-10})$, for all $i \in [n]_+$ we have that 
\begin{align*}
	&\left\|  \sum_{\cS \in \bcE, i \in \cS_+} \frac{L_{\cS}}{L}\left\{\frac{e^{\tilde{\theta}_i}\left(\sum_{l \in \cS_+ \backslash \{i\}} e^{\tilde{\theta}_l}\right)}{\left(\sum_{j \in \cS_+} e^{\tilde{\theta}_j}\right)^2} \eb_i -  \sum_{l \in \cS_+ \backslash\{i\}} \frac{e^{\tilde{\theta}_i} e^{\tilde{\theta}_l} }{\left(\sum_{j \in \cS_+} e^{\tilde{\theta}_j}\right)^2}\eb_l \right\}  - [ \nabla^2 \cL(\thetab^* ; \bx)]_{i+1} \right\|_1 \\
	& \lesssim  \sum_{\cS \in \bcE, i \in \cS_+} \frac{L_{\cS}}{L} \sum_{l \in \cS_+ \backslash \{i\}} \left| \frac{e^{\tilde{\theta}_i} e^{\tilde{\theta}_l} }{\left(\sum_{j \in \cS_+} e^{\tilde{\theta}_j}\right)^2} - \frac{e^{{\theta}^*_i} e^{{\theta}^*_l} }{\left(\sum_{j \in \cS_+} e^{{\theta}^*_j}\right)^2} \right|\\
	& \lesssim  \sum_{\cS \in \bcE, i \in \cS_+} \frac{L_{\cS}}{L}\sum_{l \in \cS_+ \backslash \{i\}} \left(\frac{e^{{\theta}^*_i}}{\sum_{j \in \cS_+}e^{\theta^*_j}}\right)\left(\frac{e^{{\theta}^*_l}}{\sum_{j \in \cS_+}e^{\theta^*_j}}\right)\|\hat\thetab - \thetab^*\|_{\infty}\\
	& \lesssim \|\hat\thetab - \thetab^*\|_{\infty} \sum_{\cS \in \bcE, i \in \cS_+} \frac{L_{\cS}}{L \cdot |\cS_+|} \lesssim  \|\hat\thetab - \thetab^*\|_{\infty} (\Lb_{\bcE})_{i+1,i+1} \lesssim  \bar\lambda \cdot\|\hat\thetab - \thetab^*\|_{\infty} .
\end{align*}}
{
Putting the above analysis together, with  probability at least $1 - O(n^{-10})$, we have that for all $i \in [n]_+$,
\begin{align*}
	& \left|\left[\nabla \cL(\hat\thetab; \bx) - \nabla \cL(\thetab^*; \bx) - \nabla^2 \cL(\thetab^* ; \bx)(\hat\thetab - \thetab^*) \right]_{i+1}\right| \lesssim  \bar\lambda\cdot \|\hat\thetab - \thetab^*\|_{\infty}^2 \lesssim \frac{\kappa_{\bcE}^2 n\log n}{L},
\end{align*}
and hence \eqref{eq: cck term 2} holds.}
\begin{remark}
	Recall that 
	{ $$
	\left[\nabla \cL(\hat\thetab; \bx) - \nabla \cL(\thetab^*; \bx) \right]_{i+1} = \! \sum_{\cS \in \bcE, i \in \cS_+}\!\! \frac{L_{\cS}}{L}\left\{\frac{e^{\tilde{\theta}_i}\left(\sum_{j \in \cS_+ \backslash \{i\}} e^{\tilde{\theta}_j}\right)}{\left(\sum_{j \in \cS_+} e^{\tilde{\theta}_j}\right)^2} \eb_i -  \sum_{l \in \cS_+ \backslash\{i\}} \frac{e^{\tilde{\theta}_i} e^{\tilde{\theta}_l} }{\left(\sum_{j \in \cS_+} e^{\tilde{\theta}_j}\right)^2}\eb_l \right\}^{\top}\!\! (\hat\thetab - \thetab^*),
	$$}
	where $\tilde\thetab$ lies on the line between $\hat\thetab - \thetab^*$. Thus with probability at least $1 - O(n^{-10})$, under the condition that $ \sqrt{{n \kappa_{\bcE} \log n}/{(\underline{\lambda}L)}}  \le c$ for some small enough constant $c > 0$, for all $ i \in [n]_+$ we have that 
	{
    \begin{align*}
		&\left| \left[\nabla \cL(\hat\thetab; \bx) - \nabla \cL(\thetab^*; \bx) \right]_{i+1} \right| \\
		& \le \left\| \sum_{\cS \in \bcE, i \in \cS_+}\frac{L_{\cS}}{L} \left\{\frac{e^{\tilde{\theta}_i}\left(\sum_{j \in \cS_+ \backslash \{i\}} e^{\tilde{\theta}_j}\right)}{\left(\sum_{j \in \cS_+} e^{\tilde{\theta}_j}\right)^2} \eb_i -  \sum_{l \in \cS_+ \backslash\{i\}} \frac{e^{\tilde{\theta}_i} e^{\tilde{\theta}_l} }{\left(\sum_{j \in \cS_+} e^{\tilde{\theta}_j}\right)^2}\eb_l \right\}\right\|_1 \|\hat\thetab - \thetab^*\|_{\infty} \\
		& \lesssim \left\{ \sum_{\cS \in \bcE, i \in \cS_+} \frac{L_{\cS}}{L} \cdot \frac{e^{\tilde{\theta}_i}\left(\sum_{j \in \cS_+ \backslash \{i\}} e^{\tilde{\theta}_j}\right)}{\left(\sum_{j \in \cS_+} e^{\tilde{\theta}_j}\right)^2} \right\} \|\hat\thetab - \thetab^*\|_{\infty} \lesssim \left( \sum_{\cS \in \bcE, i \in \cS_+} \frac{L_{\cS}}{L \cdot |\cS_+|} \right) \|\hat\thetab - \thetab^*\|_{\infty} \\
		& \lesssim  \|\hat\thetab - \thetab^*\|_{\infty} (\Lb_{\bcE})_{i+1,i+1} \lesssim  \bar\lambda \cdot\|\hat\thetab - \thetab^*\|_{\infty}.
	\end{align*}}
    Then with probability at least $1- O(n^{-10})$ we have 
	\begin{equation}\label{eq: gradient diff inf norm}
		\left\|\nabla \cL(\hat\thetab; \bx) - \nabla \cL(\thetab^*; \bx)\right\|_{\infty} \lesssim {  \bar\lambda \cdot \|\hat\thetab - \thetab^*\|_{\infty} \lesssim \sqrt{\frac{\kappa_{\bcE}^2 n \bar\lambda\log n}{ L}} .}
	\end{equation}
\end{remark}
\subsubsection{Proof of \ref{eq: cck term 3}}\label{sec: proof of cck term 3}
For any $i \in [n]_+$, very similar to the proof in Section \ref{sec: proof of cck term 2}, with probability at least $1 - O(n^{-10})$, for all $i \in [n]_+$ we have 
\begin{align*}
	& \|[\nabla^2 \cL(\hat\thetab; \bx)]_{i+1} - [\nabla^2 \cL(\thetab^*; \bx)]_{i+1}\|_1\\
	& = \left\|  \sum_{\cS \in \bcE, i \in \cS_+} \left\{\frac{e^{\hat{\theta}_i}\left(\sum_{l \in \cS_+ \backslash \{i\}} e^{\hat{\theta}_l}\right)}{\left(\sum_{j \in \cS_+} e^{\hat{\theta}_j}\right)^2} \eb_i -  \sum_{l \in \cS_+ \backslash\{i\}} \frac{e^{\hat{\theta}_i} e^{\hat{\theta}_l} }{\left(\sum_{j \in \cS_+} e^{\hat{\theta}_j}\right)^2}\eb_l \right\}  - [ \nabla^2 \cL(\thetab^* ; \bx)]_{i+1} \right\|_1 \\
	& \lesssim { \bar\lambda \cdot \|\hat\thetab - \thetab^*\|_{\infty},}
\end{align*}
and hence by Theorem \ref{thm: inf norm MLE} with probability at least $1- O(n^{-10})$ we have 
$$
\|\nabla^2 \cL(\hat\thetab; \bx) - \nabla^2 \cL(\thetab^*; \bx)\|_{\infty} = \max_{i \in [n+1]}\|[\nabla^2 \cL(\hat\thetab; \bx)]_{i+1} - [\nabla^2 \cL(\thetab^*; \bx)]_{i+1}\|_1 \lesssim { \sqrt{\frac{\kappa_{\bcE}^2 n \bar\lambda\log n}{ L}} .}
$$
\subsubsection{Proof of \ref{eq: cck term 5}}

For the simplicity of notation, we let $\tilde\lambda_1 \ge \ldots \ge \tilde\lambda_n \ge \tilde\lambda_{n+1} = 0$ be the eigenvalues of $\nabla^2 \cL(\thetab; \bx)$ in descending order, and let $\bv_1, \ldots, \bv_n$ and $\frac{1}{\sqrt{n+1}}\mathbf{1}$ be the normalized eigenvectors corresponding to $\tilde\lambda_1, \ldots, \tilde\lambda_{n+1}$. From the proof of Lemma \ref{lm: eigen hessian}, we know that for any $\thetab \in \RR^{n+1}$ such that $\|\thetab - \thetab^*\|_{\infty} \le C$ for some constant $C>0$, we have 
$\frac{1}{(\kappa_{\thetab} e^{2C})^2} \Lb_{\bcE} \preceq \nabla^2 \cL (\thetab; \bx) \preceq (\kappa_{\thetab} e^{2C})^2\Lb_{\bcE}$,
{ which in turn gives that $\tilde\lambda_1 \asymp \bar\lambda$ and $\tilde\lambda_n \asymp \underline{\lambda}$.} Also recall that $\nabla^2 \cL(\thetab; \bx) \mathbf{1} = \mathbf{0}$, and hence for any constant $C > 0$, we have the eigen-decomposition
\begin{align*}
	& \begin{pmatrix}
		\nabla^2 \cL(\thetab; \bx) & C\mathbf{1}\\
		C\mathbf{1}^{\top} & 0
	\end{pmatrix} = \begin{pmatrix}
		\nabla^2 \cL(\thetab; \bx) & \mathbf{0}\\
		\mathbf{0} & 0
	\end{pmatrix} + C\begin{pmatrix}
		\mathbf{0} & \mathbf{1}\\
		\mathbf{1}^{\top} & 0
	\end{pmatrix} = \sum_{i = 1}^{n} \tilde\lambda_i \begin{pmatrix}
		\bv_i\\ 0
	\end{pmatrix} \begin{pmatrix}
		\bv_i\\ 0
	\end{pmatrix}^{\top}\\
	& \quad + C \sqrt{n+1}\begin{pmatrix}
		\frac{1}{\sqrt{2(n+1)}}\mathbf{1} \\ \frac{1}{\sqrt{2}}
	\end{pmatrix} \!\!\!\begin{pmatrix}
		\frac{1}{\sqrt{2(n+1)}}\mathbf{1} \\ \frac{1}{\sqrt{2}}
	\end{pmatrix}^{\top}\!\!\!\!\! - C \sqrt{n+1}\begin{pmatrix}
		\frac{1}{\sqrt{2(n+1)}}\mathbf{1} \\ -\frac{1}{\sqrt{2}}
	\end{pmatrix} \!\!\!\begin{pmatrix}
		\frac{1}{\sqrt{2(n+1)}}\mathbf{1} \\ -\frac{1}{\sqrt{2}}
	\end{pmatrix}^{\top}.
\end{align*}
It is not hard to verify by the relationship between $\bv_i$'s and $\mathbf{1}$ that the above representation is a valid eigen-decomposition of $\begin{pmatrix}
	\nabla^2 \cL(\thetab; \bx) & C\mathbf{1}\\
	C\mathbf{1}^{\top} & 0
\end{pmatrix}$.
Thus we can see that the $n + 2$ eigenvalues of $\begin{pmatrix}
	\nabla^2 \cL(\thetab; \bx) & C\mathbf{1}\\
	C\mathbf{1}^{\top} & 0
\end{pmatrix}$ are $\tilde\lambda_1, \ldots, \tilde\lambda_n$ and $\pm C \sqrt{n+1}$,
and  $\begin{pmatrix}
	\nabla^2 \cL(\thetab; \bx) & C\mathbf{1}\\
	C\mathbf{1}^{\top} & 0
\end{pmatrix}^{-1}$ would take the following form 
\begin{align*}
	& \begin{pmatrix}
		\nabla^2 \cL(\thetab; \bx) & C\mathbf{1}\\
		C\mathbf{1}^{\top} & 0
	\end{pmatrix}^{-1} = \sum_{i = 1}^{n} \tilde\lambda_i^{-1} \begin{pmatrix}
		\bv_i\\ 0
	\end{pmatrix}\!\! \begin{pmatrix}
		\bv_i\\ 0
	\end{pmatrix}^{\top}\!\!\!\!+ C^{-1}(n+1)^{-1/2}\begin{pmatrix}
		\frac{1}{\sqrt{2(n+1)}}\mathbf{1} \\ \frac{1}{\sqrt{2}}
	\end{pmatrix} \!\!\!\begin{pmatrix}
		\frac{1}{\sqrt{2(n+1)}}\mathbf{1} \\ \frac{1}{\sqrt{2}}
	\end{pmatrix}^{\top}\!\!\!\!\! \\
	& \quad - C^{-1} (n+1)^{-1/2}\begin{pmatrix}
		\frac{1}{\sqrt{2(n+1)}}\mathbf{1} \\ -\frac{1}{\sqrt{2}}
	\end{pmatrix} \!\!\!\begin{pmatrix}
		\frac{1}{\sqrt{2(n+1)}}\mathbf{1} \\ -\frac{1}{\sqrt{2}}
	\end{pmatrix}^{\top} = \begin{pmatrix}
		\nabla^2 \cL(\thetab; \bx)^{\dagger}& \frac{1}{C(n+1)} \mathbf{1}\\
		\frac{1}{C(n+1)} \mathbf{1}^{\top} & 0
	\end{pmatrix},
\end{align*}
where $\nabla^2 \cL(\thetab; \bx)^{\dagger} = \sum_{i = 1}^{n} \tilde\lambda_i^{-1} \bv_i \bv_i^{\top}$. Then for any $\thetab \in \RR^{n+1}$ such that $\|\thetab - \thetab^*\|_{\infty} \le C$ for some constant $C >0$ and a constant $C_n > 0 $ (dependent on $n$) such that $\underline{\lambda} \le C_n \sqrt{n} $, we have 
$$
\left\|\begin{pmatrix}
	\nabla^2 \cL(\thetab; \bx) & C_n \mathbf{1} \\
	C_n \mathbf{1}^{\top} & 0
\end{pmatrix}^{-1}\right\|_2 = \max( \frac{1}{\lambda_{n}(\nabla^2 \cL(\thetab; \bx))},\frac{1}{C_n \sqrt{n+1}} ) \lesssim { \frac{1}{\underline{\lambda}}.}
$$

Thus by \ref{eq: cck term 3}, with probability at least $1 - O(n^{-10})$ we have
\begin{align*}
	&\left\|\begin{pmatrix}
		\nabla^2 \cL(\hat\thetab; \bx) & \mathbf{1} \\
		\mathbf{1}^{\top} & 0
	\end{pmatrix}^{-1} - \begin{pmatrix}
		\nabla^2 \cL(\thetab^*; \bx) & \mathbf{1} \\
		\mathbf{1}^{\top} & 0
	\end{pmatrix}^{-1}\right\|_2 = \left\| \nabla^2 \cL(\hat\thetab; \bx)^{\dagger} - \nabla^2 \cL(\thetab^*; \bx)^{\dagger} \right\|_2\\
	& = \left\|\begin{pmatrix}
		\nabla^2 \cL(\hat\thetab; \bx) & C_n\mathbf{1} \\
		C_n\mathbf{1}^{\top} & 0
	\end{pmatrix}^{-1} - \begin{pmatrix}
		\nabla^2 \cL(\thetab^*; \bx) & C_n\mathbf{1} \\
		C_n\mathbf{1}^{\top} & 0
	\end{pmatrix}^{-1}\right\|_2 \\
	& \le \!\left\|\!\begin{pmatrix}
		\nabla^2 \cL(\hat\thetab; \bx) & \!\!C_n\mathbf{1} \\
		C_n\mathbf{1}^{\top} & \!\!0
	\end{pmatrix}^{\!\!-1}\!\right\|_2 \!\left\|\!\begin{pmatrix}
		\nabla^2 \cL(\thetab^*; \bx) &\!\! C_n\mathbf{1} \\
		C_n\mathbf{1}^{\top} & \!\!0
	\end{pmatrix}^{\!\!-1}\!\right\|_2 \!\left\|\!\begin{pmatrix}
		\nabla^2 \cL(\hat\thetab; \bx) &\!\! C_n\mathbf{1} \\
		\!\!C_n\mathbf{1}^{\top} & 0
	\end{pmatrix} \!-\! \begin{pmatrix}
		\nabla^2 \cL(\thetab^*; \bx) & \!\!C_n\mathbf{1} \\
		\!\!C_n\mathbf{1}^{\top} & 0
	\end{pmatrix}\!\right\|_2\\
	& \lesssim { \frac{\sqrt{n}}{\underline{\lambda}^2} \|\nabla^2 \cL(\hat\thetab; \bx) - \nabla^2 \cL(\thetab^*; \bx)\|_{\infty} \lesssim \frac{n}{\underline{\lambda}} \sqrt{\frac{\kappa_{\bcE}^3\log n}{\underline{\lambda} L}}.}
\end{align*}

\subsection{Proof of Corollary \ref{col: bTheta entry bound}}\label{sec: proof col btheta entry bound}
From the proof of Lemma \ref{lm: cck term bounds} it can be seen that 
$$
\begin{pmatrix}
	\nabla^2 \cL(\thetab; \bx) & \mathbf{1}\\
	\mathbf{1}^{\top} & 0
\end{pmatrix}^{-1} = \begin{pmatrix}
	\nabla^2 \cL(\thetab; \bx)^{\dagger} & \frac{1}{n+1} \mathbf{1}\\
	\frac{1}{n+1} \mathbf{1}^{\top} & 0
\end{pmatrix}
$$
holds true for any $\thetab \in \RR^{n+1}$, where $\nabla^2 \cL(\thetab; \bx)^{\dagger}$ is the Moore-Penrose inverse of $\nabla^2 \cL(\thetab; \bx)$.
Now define $\mathbf{\Gamma} = (\bv_1, \bv_2, \ldots, \bv_n, \frac{1}{\sqrt{n+1}}\mathbf{1})$, and denote by $\boldsymbol{\gamma}_j = (\gamma_{j1}, \ldots, \gamma_{jn}, \frac{1}{\sqrt{n+1}})$ the $j$-th row of $\mathbf{\Gamma}$, $j \in [n+1]$. Then since $\mathbf{\Gamma}$ is orthonormal, it can be seen that $\|\boldsymbol\gamma_j\|_2^2 = \sum_{k = 1}^{n} \gamma_{jk}^2 + \frac{1}{n+1} = 1$ and $\sum_{k = 1}^{n} \gamma_{jk}^2 \asymp 1$. Thus for any $\thetab \in \RR^{n+1}$ such that $\|\thetab - \thetab^*\|_{\infty} < C$ for some constant $C>0$,  for $j, k \in [n+1], j \neq k$, 
we have
{
\begin{align*}
	\frac{1}{\bar\lambda} \lesssim [\nabla^2 \cL(\thetab; \bx)^{\dagger}]_{jj} & = \eb_j^{\top} \nabla^2 \cL(\thetab; \bx)^{\dagger}\eb_j = \boldsymbol{\gamma}_j^{\top} \begin{pmatrix}
		\tilde\lambda_1^{-1} & \ldots & & 0\\
		\vdots & \ddots & &\\
		& & \tilde\lambda_n^{-1} & \\
		0&&&0
	\end{pmatrix}\boldsymbol{\gamma}_j = \sum_{i = 1}^n \gamma_{ji}^2 \tilde\lambda_i^{-1} \lesssim \frac{1}{\underline{\lambda}},\\
	\big|[\nabla^2 \cL(\thetab; \bx)^{\dagger}]_{jk} \big|&= \big |\eb_j^{\top} \nabla^2 \cL(\thetab; \bx)^{\dagger}\eb_k \big| 
	\le \|\nabla^2 \cL(\thetab; \bx)^{\dagger}\|_2
	\lesssim \frac{1}{\underline{\lambda}}.
\end{align*}}

\subsection{Proof of Lemma \ref{thm: mle decompose}}\label{sec: proof thm mle decomp}
By Corollary~\ref{col: bTheta entry bound} and the definition of $\hat\thetab^d$, we have that
\begin{align*}
	\hat\thetab^d &=  \hat\thetab' - \nabla^2 \cL(\hat\thetab; \bx)^{\dagger}\nabla \cL(\hat\thetab; \bx) =  \hat\thetab' - \left(\Ib_{n+1}, \mathbf{0}\right)\begin{pmatrix}
		\nabla^2 \cL(\hat\thetab; \bx) & \mathbf{1}\\
		\mathbf{1}^{\top} & 0
	\end{pmatrix}^{-1} \begin{pmatrix}
		\Ib_{n+1}\\\mathbf{0}
	\end{pmatrix}\nabla \cL(\hat\thetab; \bx)\\
	& = \left(\Ib_{n+1}, \mathbf{0}\right) \begin{pmatrix}
		\hat\thetab'\\
		0
	\end{pmatrix} - \left(\Ib_{n+1}, \mathbf{0}\right) \begin{pmatrix}
		\nabla^2 \cL(\hat\thetab; \bx) & \mathbf{1}\\
		\mathbf{1}^{\top} & 0
	\end{pmatrix}^{-1}\begin{pmatrix}
		\nabla \cL(\hat\thetab; \bx) \\ 0
	\end{pmatrix},
\end{align*}
then left-multiply both sides by $\left(\Ib_{n+1}, \mathbf{0}\right)^{\top}$ we have 
\begin{align*}
	\begin{pmatrix}
		\hat\thetab^d \\
		0
	\end{pmatrix} &= \diag(\Ib_{n+1},0)\begin{pmatrix}
		\hat\thetab'\\
		0
	\end{pmatrix} - \diag(\Ib_{n+1},0)\begin{pmatrix}
		\nabla^2 \cL(\hat\thetab; \bx) & \mathbf{1}\\
		\mathbf{1}^{\top} & 0
	\end{pmatrix}^{-1}\begin{pmatrix}
		\nabla \cL(\hat\thetab; \bx) \\ 0
	\end{pmatrix}\\
	& \quad = \begin{pmatrix}
		\hat\thetab'\\
		0
	\end{pmatrix} - \begin{pmatrix}
		\nabla^2 \cL(\hat\thetab; \bx) & \mathbf{1}\\
		\mathbf{1}^{\top} & 0
	\end{pmatrix}^{-1}\begin{pmatrix}
		\nabla \cL(\hat\thetab; \bx) \\ 0
	\end{pmatrix},
\end{align*}
where the second equality is due to the fact that $\mathbf{1}_{n+1}^{\top} \nabla \cL(\hat\thetab; \bx)  = 0$.
Hence we can see that $\hat\thetab^d$ can be treated as the sub-vector for the first $n+1$ coordinates of an augmented Newton-debiased estimator.  
Recall that we state in the proof of Theorem \ref{thm: inf norm MLE}, we will abuse the notation and let $\hat\thetab$ denote $\hat\thetab' = \ponep\hat\thetab$ and let $\thetab^*$ denote $\thetab^{*\prime} = \ponep\thetab^* $. Then we have the following decomposition

\begin{align*}
	\left(\begin{array}{c}\!\!
		\hat\thetab^{d}-\thetab^{*}\!\! \\
		0
	\end{array}\right)\!\!=& \underbrace{\left(\!\!\!\left(\begin{array}{cc}\!\!
			\nabla^{2} \mathcal{L}(\hat{\thetab}; \bx) & \mathbf{1} \\
			\mathbf{1}^{\top} & 0
		\end{array}\right)^{\!\!-1}\!\!\!\!\!\!-\left(\begin{array}{cc}\!\!
			\nabla^{2} \mathcal{L}\left(\thetab^{*}; \bx\right) & \mathbf{1} \\
			\mathbf{1}^{\top} & 0
		\end{array}\right)^{\!\!-1}\!\right)\!\!\left(\begin{array}{c}
			\!\!\!-\nabla \mathcal{L}(\hat{\thetab}; \bx)\!+\!\nabla^{2} \mathcal{L}(\hat{\thetab}; \bx)\!\!\left(\hat{\thetab}\!-\!\thetab^{*}\right) \!\!\!\!\!\\
			0
		\end{array}\right)}_{I_{1}} \\
	&+\underbrace{\left(\begin{array}{cc}
			\nabla^{2} \mathcal{L}\left(\thetab^{*}; \bx\right) & \mathbf{1} \\
			\mathbf{1}^{\top} & 0
		\end{array}\right)^{-1}\left(\begin{array}{c}
			\nabla \mathcal{L}\left(\thetab^{*}; \bx\right)-\nabla \mathcal{L}(\hat{\thetab}; \bx)+\nabla^{2} \mathcal{L}(\hat{\thetab}; \bx)\left(\hat{\thetab}-\thetab^{*}\right) \\
			0
		\end{array}\right)}_{I_{2}} \\
	&+\left(\begin{array}{cc}
		\nabla^{2} \mathcal{L}\left(\thetab^{*}; \bx\right) & \mathbf{1} \\
		\mathbf{1}^{\top} & 0
	\end{array}\right)^{-1}\left(\begin{array}{c}
		-\nabla \mathcal{L}\left(\thetab^{*}; \bx \right) \\
		0
	\end{array}\right) .
\end{align*}
By Lemma \ref{lm: cck term bounds}, the following bounds for the terms $I_1$ and $I_2$ hold with probability at least $1-O(n^{-10})$,
\begin{align*}
	\| I_1 \|_{2} &\le \left\|\begin{pmatrix}
		\nabla^2 \cL(\hat\thetab; \bx) & \mathbf{1} \\
		\mathbf{1}^{\top} & 0
	\end{pmatrix}^{\!\!-1} \!\!-\! \begin{pmatrix}
		\nabla^2 \cL(\thetab^*; \bx) & \mathbf{1} \\
		\mathbf{1}^{\top} & 0
	\end{pmatrix}^{\!\!-1}\right\|_{2} \left\| - \nabla\cL(\hat\thetab; \bx) + \nabla^2 \cL(\thetab^*; \bx)(\hat\thetab - \thetab^*)\right\|_{2}\\
	& \lesssim \sqrt{n} \left\|\begin{pmatrix}
		\nabla^2 \cL(\hat\thetab; \bx) & \mathbf{1} \\
		\mathbf{1}^{\top} & 0
	\end{pmatrix}^{\!\!-1} \!\!-\! \begin{pmatrix}
		\nabla^2 \cL(\thetab^*; \bx) & \mathbf{1} \\
		\mathbf{1}^{\top} & 0
	\end{pmatrix}^{\!\!-1}\right\|_2 \left\| - \nabla\cL(\hat\thetab; \bx) + \nabla^2 \cL(\thetab^*; \bx)(\hat\thetab - \thetab^*)\right\|_{\infty}\\
	& { \lesssim \sqrt{n} \times \frac{n}{\underline{\lambda}} \sqrt{\frac{\kappa_{\bcE}^3\log n}{\underline{\lambda} L}} \left(\sqrt{\frac{ \bar\lambda\log n}{L}} + \frac{\kappa_{\bcE}^2 n \log n}{L}\right) \lesssim \frac{\kappa_{\bcE}^2 n \sqrt{n} \log n}{\underline{\lambda} L}\left(n\sqrt{\frac{\kappa_{\bcE}^3 \log n}{\underline{\lambda}L}}+1\right)},
\end{align*}
and 
\begin{align*}
	\|I_2\|_{2} &= \left\|\left(\begin{array}{cc}
		\nabla^{2} \mathcal{L}\left(\thetab^{*}; \bx\right)^{\dagger} & \frac{1}{n+1}\mathbf{1} \\
		\frac{1}{n+1}\mathbf{1}^{\top} & 0
	\end{array}\right)\!\!\left(\begin{array}{c}\!\!
		\nabla \mathcal{L}\left(\thetab^{*}; \bx\right)-\nabla \mathcal{L}(\hat{\thetab}; \bx)+\nabla^{2} \mathcal{L}(\hat{\thetab}; \bx)\left(\hat{\thetab}-\thetab^{*}\right) \\
		\!\!0
	\end{array}\!\!\!\right)\right\|_{2}\\
	&= \left\|\nabla^{2} \mathcal{L}\left(\thetab^{*}; \bx\right)^{\dagger}\left(\nabla \mathcal{L}\left(\thetab^{*}; \bx\right)-\nabla \mathcal{L}(\hat{\thetab}; \bx)+\nabla^{2} \mathcal{L}(\hat{\thetab}; \bx)\left(\hat{\thetab}-\thetab^{*}\right) \right)\right\|_{2}\\
	& \lesssim  \sqrt{n}\left\|\nabla^{2} \mathcal{L}\left(\thetab^{*}; \bx\right)^{\dagger}\right\|_2 \|\nabla \cL(\hat\thetab; \bx) - \nabla \cL(\thetab^*; \bx) - \nabla^2 \cL(\thetab^*; \bx)(\hat\thetab - \thetab^*)\|_{\infty}\\
	& { \lesssim \frac{\kappa_{\bcE}^2 n \log n}{L} \times \frac{\sqrt{n}}{\underline{\lambda} } =  \frac{\kappa_{\bcE}^2 n \sqrt{n} \log n}{ \underline{\lambda} L},}
\end{align*}
where the second equality is due to the fact that  $\mathbf{1}_{n+1}^{\top} \nabla \cL(\thetab; \bx) = 0$ and $\mathbf{1}_{n+1}^{\top}\nabla^2 \cL(\thetab; \bx) = \mathbf{0}$ for any $\thetab \in \RR^{n+1}$. Thus in turn we have that 
{ $$
\|\Rb_0\|_{2} \le \|I_1\|_{2} + \|I_2\|_{2} \lesssim  \frac{\kappa_{\bcE}^2 n \sqrt{n} \log n}{\underline{\lambda} L}\left(n\sqrt{\frac{\kappa_{\bcE}^3 \log n}{\underline{\lambda}L}}+1\right).
$$}
{
	\section{Complexity of Generating the Observed Offer Sets for Numerical Studies}\label{sec: sample complexity cE}
	\begin{proposition}\label{prop: sample complexity cE}
		Denote by $B_{\bcE}$ the complexity of generating $\bcE$ with the procedure in Section~\ref{sec: numerical results}. When $p \asymp n \log n / 2^n$, we have $\EE(B_{\bcE}) = O(n^2 \log n)$. Furthermore, with probability at least $1 - O(n^{-1})$, we have $B_{\bcE} \le C n^2 \log n$ for some constant $C > 0$.
	\end{proposition}
	\proof{Proof.} Note that Step~(1) only requires generating a scalar random variable, and its computational cost is dominated by Steps~(2) and (3). We first show that $\EE(B_{\bcE}) = O(n^2 \log n)$. Since $M \sim {\rm Binomial} (2^n - 1, p)$, we know that $\EE(M) = (2^n -1) p = O(n \log n)$. Conditional on $M$, for each $j =1, \ldots, M$, denote by $I_j$ the number of sampling attempts for the $j$-th iteration, and we have that $I_j$ follows a geometric distribution with success probability equal to $(2^n - j)/ 2^n $, i.e., $I_j - 1 \sim {\rm Geometric} (1, (2^n - j)/ 2^n)$ and $\mathbb{E} (I_j| M) = \mathbb{E}(I_j) = 1 + j/(2^n - j) $. Hence we have that
	\begin{align*}
		\EE(B_{\bcE}) &\lesssim n \mathbb{E} \Big(\sum_{j=1}^M I_j \Big) = n \mathbb{E}\Big(\EE \Big(\sum_{j=1}^M I_j | M\Big) \Big) = n\EE \Big( M + \sum_{j=1}^M \frac{j}{2^n - j}\Big)\\
		& \lesssim n^2 \log n + n \EE \Big(\frac{M^2}{2^n - M}\Big) = n^2 \log n + \frac{np}{1-p}\sum_{M=1}^{2^n - 1} (M-1+1) {2^n - 1 \choose M -1} p^{M-1} (1-p)^{2^n - M}\\
		& \lesssim  n^2 \log n + np\EE(M) \lesssim n^2 \log n .
	\end{align*}
	Next, we move on to show that there exists a constant $C >0$ such that  $B_{\bcE} \le C n^2 \log n$ with probability at least $1 - O(n^{-1})$. By Hoeffding's inequality \citep{hoeffding1994probability}, with probability at least $1-n^{-1}$, we have that
	\begin{equation}\label{eq: arg 1}
		M \le \EE(M) + \sqrt{2^{n+1}p(1-p) \log n} \le  C n \log n,
	\end{equation}
	where $C > 0$ is a constant independent of $n$. Besides, conditional on the event $M \le C n \log n$, the probability that $I_j = 1$ for all $j =1, \ldots, M$ is 
	\begin{equation}\label{eq: arg 2}
		\prod_{j = 2}^M \frac{2^n - j}{2^n} = 1 - O(n^2 \log^2 n / 2^n) = 1 - O(n^{-1}).
	\end{equation}
	Therefore, combining \eqref{eq: arg 1} and \eqref{eq: arg 2}, with probability at least $1 - O(n^{-1})$, the total complexity is $B_{\bcE} \le C n^2 \log n$. \Halmos
}
\end{document}